%% file: dissertacao.tex
\documentclass[msc]{ppgccufmg}

\pdfoutput=1

\usepackage[square]{natbib}
\usepackage[english]{babel}
\usepackage[latin1]{inputenc}
\usepackage[T1]{fontenc}
\usepackage{type1ec}
\usepackage{graphicx}
\usepackage{epigraph}
\usepackage{times}
\usepackage{parskip}
\usepackage[labelfont=bf,textfont=it]{caption}
\usepackage{subfig}
\usepackage{url}
\usepackage{caption}
\usepackage{pdflscape}
\usepackage{amssymb}
\usepackage{longtable}
\usepackage{boxedminipage}
\usepackage{multirow}
\usepackage{rotating}		

\usepackage[a4paper,
  portuguese,
  bookmarks=true,
  bookmarksnumbered=true,
  linktocpage,
  colorlinks,
  citecolor=black,
  	urlcolor=black,
  linkcolor=black,
  filecolor=black,
  ]{hyperref}

\usepackage{multirow}					 
\usepackage{algorithm, algorithmic}

\usepackage{yaacro}
   
\usepackage[usenames,dvipsnames]{xcolor}  
\newcounter{dicas}

\newcounter{portugues}

\newenvironment{Exemplo}
{
   \MakeFramed{\advance\hsize-\width\FrameRestore}%
   \noindent 
   \begin{adjustwidth}{}{7pt}%
	   \vspace{6pt}%
   \setstretch{1.2} 
}
{
   \vspace{6pt}
	   \end{adjustwidth}
   \endMakeFramed
}

\begin{document}


\ppgccufmg{
  title={A Methodology for Player Modeling based on Machine Learning},
	authorrev={Machado, Marlos Cholodovskis},  
	cutter={M149m}, 
  cdu={519.6*08 (043)},  
  university={Universidade Federal de Minas Gerais},
  course={Computer Science},
  portuguesetitle={Uma Metodologia para Modelagem de Jogadores Baseada em \\Aprendizado de M\'aquina},
  portugueseuniversity={Universidade Federal de Minas Gerais},
  portuguesecourse={Ci\^encia da Computa\c{c}\~ao},
  address={Belo Horizonte},
  date={2013-02}, 
  keywords={Computa\c{c}\~ao - Teses, Intelig\^encia artificial - Teses, Jogos eletr\^onicos}, 
  advisor={Luiz Chaimowicz},
  coadvisor={Gisele Lobo Pappa},
  approval={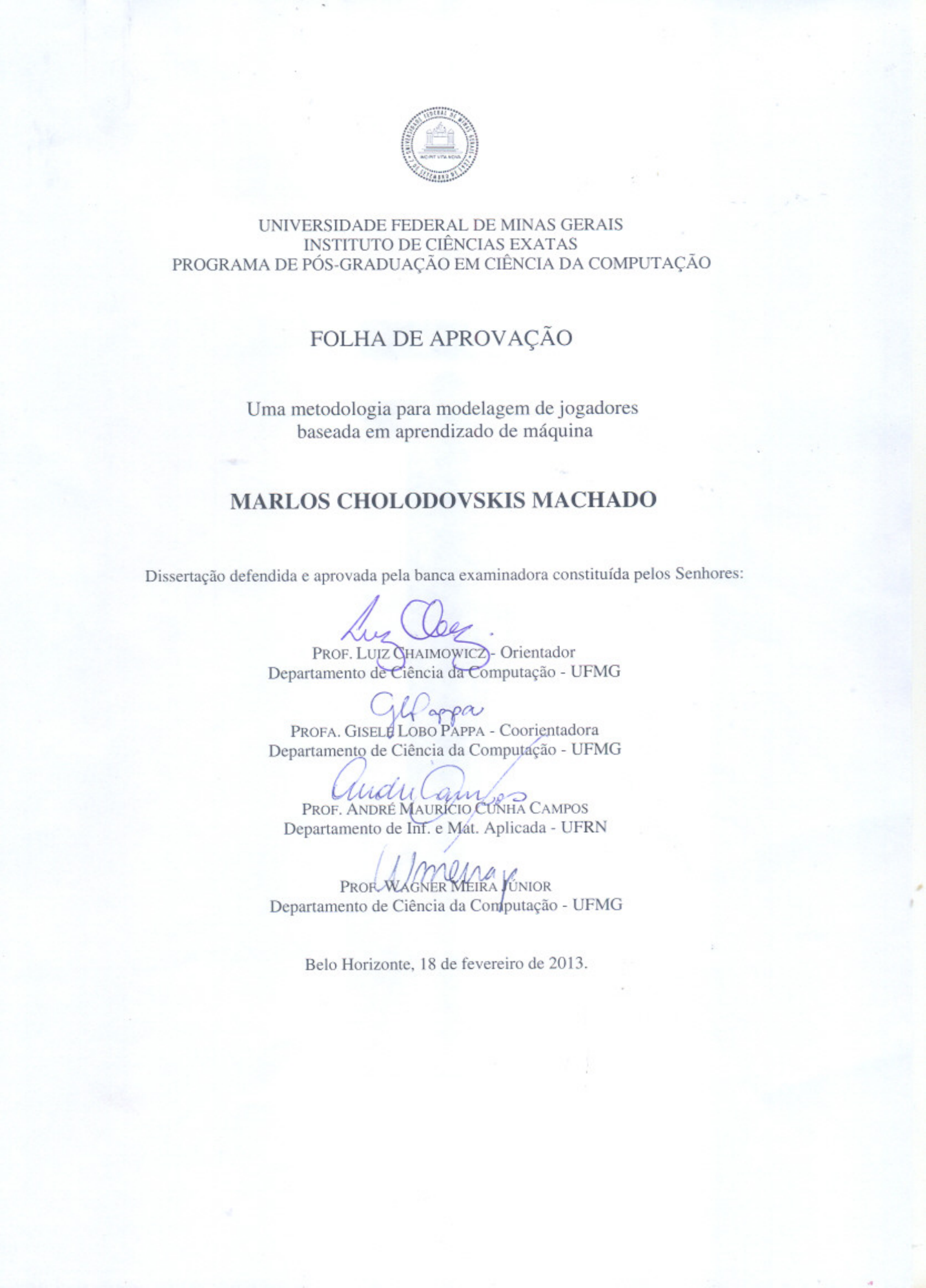},
  abstract=[brazil]{Resumo}{resumo},
  abstract={Abstract}{abstract},
	beforetoc={\input{acronimos}},
  dedication={dedicatoria},
  ack={agradecimentos},
  epigraphtext={(...) we can say that Muad'Dib learned rapidly because his first training was in how to learn. And the first 	lesson of all was the basic trust that he could learn. It is shocking to find how many people do not believe they can learn, and how many more believe learning to be difficult. Muad'Dib knew that every experience carries its lesson.}{Frank Herbert, \textit{Dune}.}
}

\input{introduction}
\input{background}
\input{related_work}
\input{generic_approach}
\input{player_modeling_civ4}
\input{experimental_results}
\input{conclusion}

\ppgccbibliography{dissertacao} 

\input{appendix} 
		
\end{document}

%% file: acronimos.tex
\chapter{List of Acronyms}

\begin{tabular}{r p{366.60538pt}}
\hline
\emph{Acronym}  & \emph{Description}
\tabularnewline
\hline
 \textbf{AI}    & \emph{Artificial Intelligence}
\tabularnewline
 \textbf{CT}    & \emph{Challenge Tailoring}
\tabularnewline
 \textbf{DDA}   & \emph{Dynamic Difficult Adjustment}
\tabularnewline
 \textbf{FPS}   & \emph{First-Person Shooter}
\tabularnewline
 \textbf{ML}    & \emph{Machine Learning}
\tabularnewline
\textbf{MMORPG} & \emph{Massively Multiplayer Online Role-Playing Game}
\tabularnewline
 \textbf{NPC}   & \emph{Non-Player Character}
\tabularnewline
 \textbf{$R^2$} & \emph{Coefficient of Determination}
\tabularnewline
 \textbf{RPG}   & \emph{Role-Playing Game}
\tabularnewline
 \textbf{RTS}   & \emph{Real-Time Strategy}
\tabularnewline
 \textbf{SMO}   & \emph{Sequential Minimal Optimization}
\tabularnewline
 \textbf{SVM}   & \emph{Support Vector Machines}
\tabularnewline
 \textbf{TBS}   & \emph{Turn-Based Strategy}
\tabularnewline
\hline

\end{tabular}

%% file: introduction.tex
\chapter{Introduction}\label{introduction}

\epigraphhead[250]{\scriptsize``Would you tell me, please, which way I ought to go from here?''\\
									 ``That dependes a good deal on where you want to get to'' said the Cat.\\
									 ``I don't much care where -'' said Alice.\\
									 ``Then it doesn't matter which way you go,'' said the Cat.\\
									 \vspace{-0.6cm}
									 \begin{flushright}
									 Lewis Carroll, Alice's Adventures in Wonderland
									 \end{flushright}
								   }
\vspace{2.5cm}

This chapter discusses the context of this thesis, as well as the motivation to research \textit{player modeling}. It also presents the thesis objectives, the main contributions of the work, the publications it generated, and an overview of the organization of the rest of the text.

\section{Context and Motivation}

The main goal of most games is entertainment \citep{Nareyek_Queue_04}. Entertainment is a subjective concept and, in order to know how much a game entertains a player, some general metrics are used. One of the most important metrics is immersion, which is generally related to how absorbing and engaging a game is~\citep{Manovich_Book_01, Taylor_Thesis_02, Bakkes_TCIAIG_09}. Two common approaches to achieve immersion are the use of stunning graphics and the development of a good Artificial Intelligence (AI) system. While graphics are responsible for initially ``seducing'' players, AI is responsible for keeping them interested in the game. 

For a long time, the game industry has put much of its efforts on the graphics of its AAA games\footnote{A game that has a high budget and is expected to sell a large number of copies.}. However, in recent years the focus has started to shift to AI, which has been commonly relegated to a less important role, and new techniques are now constantly being proposed. There are several reasons for this. Maybe the most important is the perception that the immersion achieved with amazing graphics can be spoiled by the behavior of \textit{dummy} non-player characters (NPCs). An example supporting this claim is predictable behaviors that may allow the player to discover a specific opponent weakness and repeatedly explore it during the game. \cite{Charles_CGAIDE_04} affirm that ``Often this means that the player finds it easier to succeed in the game but their enjoyment of the game is lessened because the challenge that they face is reduced and they are not encouraged to explore the full features of the game''.

Besides a greater interest from industry regarding AI techniques, the gap between the game industry and academic AI is being tightened due to the increasing performance of the new computer architectures, which has allowed the use of more sophisticated AI algorithms. At the same time, AI researchers have been considering digital games as an important platform for research. \cite{Fairclough_ICAICS_01} argue that ``computer games offer an accessible platform upon which serious cognitive research can be engaged'', while \cite{Laird_AAAI_00} suggest that computer games are the perfect platform to pursue research into human level AI. Moreover, the high level of realism achieved by some games has provided us an environment similar to the real world that can be used, for example, to evaluate robotics algorithms without the costs of sensors or real robots.

In this scenario, an AI approach that is gaining attention is \emph{player modeling}, the main topic of this thesis. According to \cite{Lucas_Report_12}:

\begin{Exemplo}
``Player modeling concerns the capturing of characteristic features of a game player in a model. Such features may encompass player actions, behaviors, preferences, goals, style, personality, attitudes, and motivations. Player models can be used to let the game adapt automatically to be better able to achieve its goals with respect to the player.''
\end{Exemplo}

We firmly believe that \emph{player modeling} is a very promising field and many works share this belief. To confirm our claim, \cite{Yannakakis_CF_12}, when discussing the current state of game AI in academy and industry, states that \emph{player experience modeling} is one of the ``four key game AI research areas that are currently reshaping the research roadmap in the game AI field''. Additionally, in 2012, a seminar was held in Dagstuhl, Germany, with the presence of several game AI experts, in order to identify their main research challenges. The report currently available \citep{Lucas_Report_12} states that \emph{player modeling} is one of these challenges. It also presents a relevant discussion about the main advantages of \emph{player modeling}:

\begin{Exemplo}
``(...) creating a player model as an intermediate step has at least two advantages: (1) it creates an understanding of who the player is, and therefore an argument for making specific adaptations; and (2) a player model allows generalization of adaptations to other games.''
\end{Exemplo}

Despite receiving much more attention recently, \emph{player modeling} has been considered a relevant topic for several years \citep{Carmel_AAAISymp_93, Herik_CIG_05, Laviers_AIIDE_09}. In this context, we present our research goals followed by our main contributions.

\section{Problem Definition and Objectives}

As discussed above, the term \textit{player modeling} can be used to model several player facets. In this work, we will use \textit{player modeling} referring to modeling player styles. This is called  \textit{preference modeling} in the nomenclature defined by \cite{Spronck_AIIDE_10, denTeuling_Thesis_10}. We intend to model player styles automatically, using data extracted from played games. In order to perform this task, it is important to ensure that the data extracted is relevant and allows us to distinguish different players.

Since \emph{player modeling} techniques can be generalized in order to be applied to a set of games, and not in a specific one, we present a generic approach for it and an evaluation of a generic representation for players in different games. To automatically identify styles in the game \textsc{Civilization IV} we instantiate our generic approach to the problem of \textit{preference modeling}. Finally, due to the huge attention \emph{player modeling} has been receiving, we also organize the field, creating a taxonomy that can be used to better understand current works and ease the discussion about their approaches.

\section{Contributions}

In summary, the main contributions of this work are:

\begin{itemize}
  \item A \emph{taxonomy} for the \emph{player modeling} field, more specifically:
    \begin{itemize}
      \item We extracted from the literature several different features and proposed a taxonomy that classifies each work according to six different aspects. These aspects are: \textit{Description, Categories, Goals, Applications, Methods} and \textit{Implementation};
      \item We categorized several important works in the literature using our taxonomy.
    \end{itemize}
    
  \item After presenting an organization for the field, we tackle the \textit{player modeling} problem in two phases:
		\begin{itemize}
			\item We propose a generic approach for \textit{player modeling} as a Machine Learning problem;
			\item We discuss a generic representation that can be used across different games, evaluating the possibility of its use in industry, showing that it attends most required features stated by \cite{Isla_GDC_05}.
			\end{itemize}

	\item We then instantiate the approach proposed in our problem i.e., modeling \textit{preferences} of \textsc{Civilization IV} players. In order to do it we performed several tasks:

\begin{itemize}
    \item Evaluated the possibility of using the \textit{generic representation} discussed, showing that we are able to infer an agent's representation observing its behavior, in the game \textsc{Civilization IV}.
    
    \item Evaluated the applicability of \textsc{Civilization IV} in-game indicators as features for an ML approach. In order to do this, we characterized \textsc{Civilization IV} agents behaviors with linear regressions: 
     \begin{itemize} 
     	\item Showing that different agents' preferences do cause an observable impact in several game indicators; and
     	\item Evaluating the impact of the match result in game indicators, verifying the importance of this information;
     \end{itemize}
    \item We used \textsc{Civilization IV} game indicators to classify, with a supervised learning approach, virtual agents' preferences. We also evaluated the use of the generated models to classify agents that were ``not known'' by the ML algorithm, since they were not in the training set;
    \item We used the models generated for virtual agents to classify self-declared human players preferences also in \textsc{Civilization IV}.
  \end{itemize}
\end{itemize}

In the following we enumerate the already published works as direct contributions of this dissertation:

\begin{itemize}
  \item \textbf{Machado, M. C.}, Fantini, E. P. C., and Chaimowicz, L. \textit{Player Modeling: Towards a Common Taxonomy}. In Proceedings of the 16th International Conference on Computer Games (CGames), pages 50-57, Louisville, United States of America, 2011.
  \item \textbf{Machado, M. C.}, Fantini, E. P. C., and Chaimowicz, L. \textit{Player Modeling: What is it? How to do it?}. In X Brazilian Symposium on Computer Games and Digital Entertainment (SBGames) - Tutorials, Salvador, Brazil, 2011.
  \item \textbf{Machado, M. C.}, Pappa, G. L., and Chaimowicz, L. \textit{Characterizing and Modeling Agents in Digital Games}. In Proceedings of the XI Brazilian Symposium on Computer Games and Digital Entertainment (SBGames), Brasilia, Brazil, 2012.
  \item \textbf{Machado, M. C.}, Rocha, B. S. L., and Chaimowicz, L. \textit{Agents Behavior and Preferences Characterization in Civilization IV}. In Proceedings of the X Brazilian Symposium on Computer Games and Digital Entertainment (SBGames), Salvador, Brazil, 2011.
  \item \textbf{Machado, M. C.}, Pappa, G. L., and Chaimowicz, L. \textit{A Binary Classification Approach for Automatic Preference Modeling of Virtual Agents in Civilization IV}. In Proceedings of the 8th International Conference on Computational Intelligence and Games (CIG), Granada, Spain, 2012.
  \item de Freitas Cunha, R., \textbf{Machado, M. C.}, and Chaimowicz, L. \textit{RTSmate: Towards an Advice System for RTS Games}. In ACM Computers in Entertainment (CiE), 2013 (\emph{in press}).
\end{itemize}

\section{Roadmap}

The remainder of this thesis is organized in seven chapters, as follows.\\

\textbf{Chapter~\ref{background}:} In the second chapter we discuss required background for this thesis. We first discuss the game platform used in this work and present its characteristics and programming interfaces. Secondly, we present the machine learning methods that were used to classify players (and virtual agents) and discuss their main differences.\\

\textbf{Chapter~\ref{related_work}:} In this chapter we present the main works related to \emph{player modeling}. We structure the chapter by \emph{player modeling} applications, namely: Game Design, Interactive Storytelling and Opponents Artificial Intelligence. When presenting the related work it became evident the huge amount of published papers in the field, and a lack of organization of these works. Due to that, we also present the taxonomy we proposed for \emph{player modeling}.\\

\textbf{Chapter~\ref{generic_approach}:} In this chapter we propose a generic approach for \textit{player modeling} as a machine learning problem and present a generic representation for players. This representation can be used for several different goals.\\

\textbf{Chapter~\ref{player_modeling_civ4}:} Once we defined a generic approach for obtaining player models with machine learning techniques, we instantiate it in our problem: preference modeling in the game \textsc{Civilization IV}. In order to do this, in this chapter we discuss the application of the generic representation proposed in Chapter~\ref{generic_approach} in the game \textsc{Civilization IV}, and perform a characterization of virtual agents behaviors. This characterization was useful to define the features to be used by the ML technique.\\

\textbf{Chapter~\ref{experimental_results}:} After instantiating the approach proposed in Chapter~\ref{generic_approach}, we classify different agent's preferences with four different ML techniques. All methods use supervised learning and they are: Naive Bayes, JRip, AdaBoost and SVM. In this chapter we evaluate the performance of these techniques to model virtual agents and (human) players' preferences in the game \textsc{Civilization IV}.\\

\textbf{Chapter~\ref{conclusion}:} Finally, in this last chapter we present our conclusions, summarize our results and discuss some future directions.

%% file: background.tex
\chapter{Background}\label{background}

\epigraphhead[230]{\scriptsize If knowledge can create problems, it is not through ignorance that we can solve them.
									 \vspace{-0.6cm}
									 \begin{flushright}
									 Isaac Asimov
									 \end{flushright}
								   }
\vspace{1cm}

This chapter presents background knowledge for the rest of this thesis. We discuss two main topics used in this work: (1) the game platform, its characteristics and programming interfaces; and (2) the main concepts of machine learning, the classifiers used in the experiments and the experimental methodology of our tests.

\section{Civilization IV}

In this thesis, we used \textsc{Civilization IV} as a game platform to perform our experiments. The platform selection is a very important step when researching digital games because implementation is generally restrained by the game interface. Additionally, it is also important to ensure that the selected game presents the basic characteristics required for the proposed research. We discuss all these topics in sequence. A deeper discussion about several different game platforms and its possibilities is presented in \citep{Machado_CGames_11}.

\textsc{Civilization IV} is a turn-based strategy game (TBS)\footnote{A turn-based game is a game where each player plays his/her turn while the others wait for his/her move. Its dynamic is very similar to board games, and it differs from real-time games because no actions are taken in parallel.} released in 2005, developed by the studio \textit{Firaxis Games}. In this game, each player is represented by a leader who controls an empire. Players/Empires compete with each other to reach one of the many game victory conditions.

\begin{figure}[t]
  \centering              
  \includegraphics[width=0.9\columnwidth]{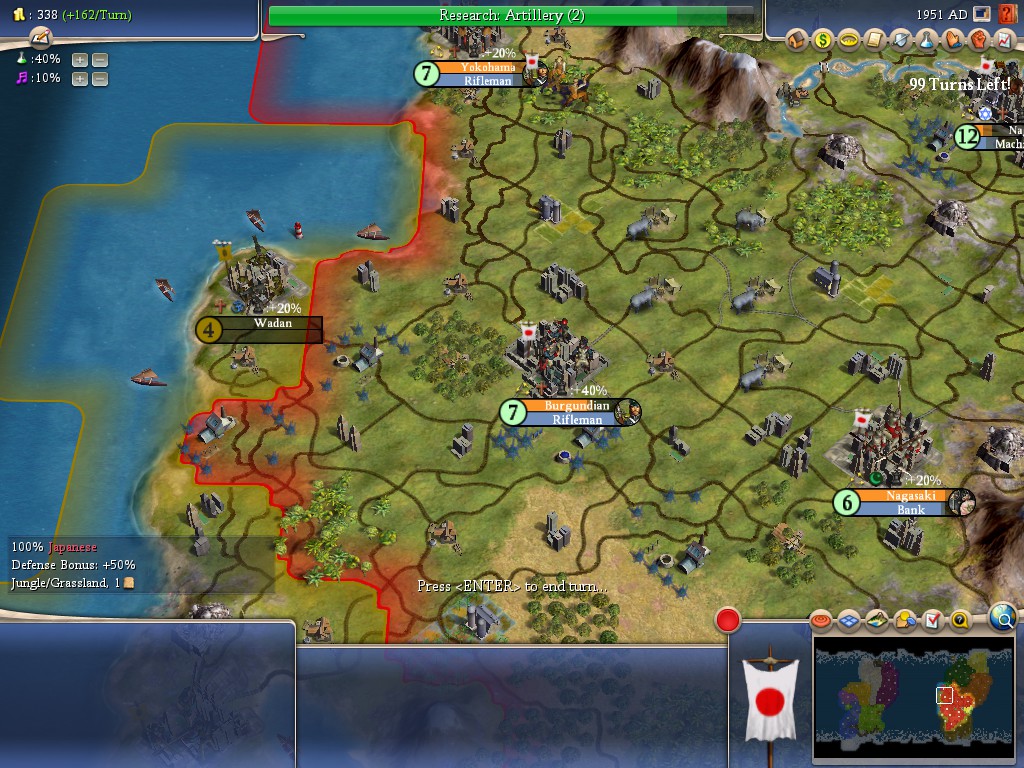}
  \caption{Screenshot of \textsc{Civilization IV}.}
  \label{fig:civ4}
\end{figure}

A high-level description of this game is nicely presented by \cite{denTeuling_Thesis_10}: ``In \textsc{Civilization IV} a player begins with selecting an empire and an appropriate leader. There are eighteen different empires available and a total of 26 leaders. Once the empire and leader have been selected, the game starts in the year 4000~BC. From here on, the player has to compete with rival leaders, manage cities, develop infrastructure, encourage scientific and cultural progress, found religions, etcetera. An original characteristic of \textsc{Civilization IV}, is that defeating the opponent is not the only way to be victorious. There are six conditions to be victorious as mentioned in \citep{Manual}: (1) \textit{Time Victory}, (2) \textit{Conquest Victory}, (3) \textit{Domination Victory}, (4) \textit{Cultural Victory}, (5) \textit{Space Race} and (6) \textit{Diplomatic Victory}. Because of these six different victory conditions the relation between the player and the opponent is different from most strategy games. The main part of the game the player is at peace with his opponents. Therefore it is possible to interact, to negotiate, to trade, to threaten and to make deals with opponents. Only after declaring war or being declared war upon, a player is at war. Any player can declare war any time, unless that player is in an agreement with an opponent which specifically forbids war declaration.'' An in-game screenshot is presented in Figure~\ref{fig:civ4}.

These six different victory possibilities make this game very interesting to this research, being one of the main reasons for selecting this platform. The game allows completely different behaviors to succeed, unlike other games in which the unique way to win is to defeat your opponents by attacking them.

In order to encompass different behaviors in its AI, each agent is characterized by a set of weighted preferences. The preferences are represented by attributes that define the way an agent plays, and are: (1)~\textit{Culture}, (2)~\textit{Gold}, (3)~\textit{Growth}, (4)~\textit{Military}, (5)~\textit{Religion} and (6)~\textit{Science}. The assigned weights represent a ``weak'' (value 2) or ``strong'' preference (value 5), besides no preference at all (value 0). Each behavior allows the agent to seek one of the six victory conditions. The main focus of this thesis is to be able to automatically identify suitable weights that represent an observed behavior, both for virtual agents and human players.

\subsubsection{Programming Interface}

In order to access game data and edit agents behaviors, \textsc{Civilization IV} offers two different possibilities: (1) to edit game resources, such as XMLs; or (2) to edit the game source code (or attach scripts to it).

The XML interface offers the possibility of configuring several game parameters, such as the agents ``flavors'' (the name they gave to the agents preferences listed above). This XML files set each agent's preferences, allowing us to edit them.
This explicit representation is another reason we selected \textit{Civilization~IV} as a testbed platform, allowing us to check each agent preference.

Editing game source code is another possibility. Its interface is an SDK that allows people to change the source code of the game and compile it, generating a DLL that replaces the traditional one.

To model agents preferences we need to use indirect observations to infer them. We attached an script to the source code to retrieve game score indicators that we use as evidences for different behaviors generated from different preferences. These indicators are constantly available to all players, and we decided to use them instead of directly evaluating actions because they represent a generalization of actions. 

The script we used is called \textit{AiAutoPlay} and it is easily found on the Web. We used a modification of it that was used to generate the dataset presented in \citep{denTeuling_Thesis_10, Spronck_AIIDE_10}. We discuss the generated dataset in Chapter~\ref{player_modeling_civ4}.

\section{Machine Learning}

Machine Learning~(ML) is a common approach for preference modeling, since we want to ``learn'' a model from a set of available players (examples), and then use this model to classify new players. There are three main approaches of learning: supervised, semi-supervised and unsupervised learning \citep{Alpaydin_Book_10}.

While in supervised learning a complete set of labeled data is available (we know beforehand the users preferences) in unsupervised learning no classes are known. Taking as an example the task of preference learning, in both supervised and unsupervised learning the data we learn from is a set of matches already played, together with the players preferences. However, in the case of supervised learning, these preferences were previously labeled by an expert, while in unsupervised learning the algorithm learns using distance measures  between data examples. Semi-supervised learning, in turn, uses both labeled and unlabeled data during the training process.

Here we model virtual agents' preferences in the game \textit{Civilization~IV} with different supervised learning techniques. Each applied technique uses a different paradigm. We discuss the main characteristics of each algorithm used in this thesis in the next sections. For a deeper explanation see \citep{Alpaydin_Book_10}.

Particularly, we discuss the four classifiers applied to solve our problem: \textit{SVM}, \textit{Naive Bayes}, \textit{JRip} and \textit{AdaBoost}. Each of them produces a different type of model, which might explore different characteristics of the data.

\subsection{Support Vector Machine}

Support Vector Machines (SVM) model classification as an optimization problem, and are considered the state of the art in classification for many different domains. Each training/test instance is modeled as a vector, where each feature represents a different dimension. SVM tries to separate the space into two different subspaces with an hyperplane. It assumes that the data can be linearly separable or that there is a kernel function able to transform the space to achieve this goal.

This separation is done with an hyperplane using a margin\footnote{``the distance from the hyperplane to the instances closest to it on either side'' \citep{Alpaydin_Book_10}.}. This margin allows noise shifts to exist without changing the class classification, since it has ``breathing space''. Each class is in one side of the margins.

The SVM problem is an optimization problem that seeks for an \textit{optimal separating hyperplane}, \emph{i.e.} maximizing the margin. This problem can be solved using quadratic optimization methods \citep{Alpaydin_Book_10}. We used the \textsc{libSVM} \citep{Chang_ACMTrans_11} implementation in our experiments.

\begin{figure}[t]
  \centering              
  \includegraphics[width=0.7\columnwidth]{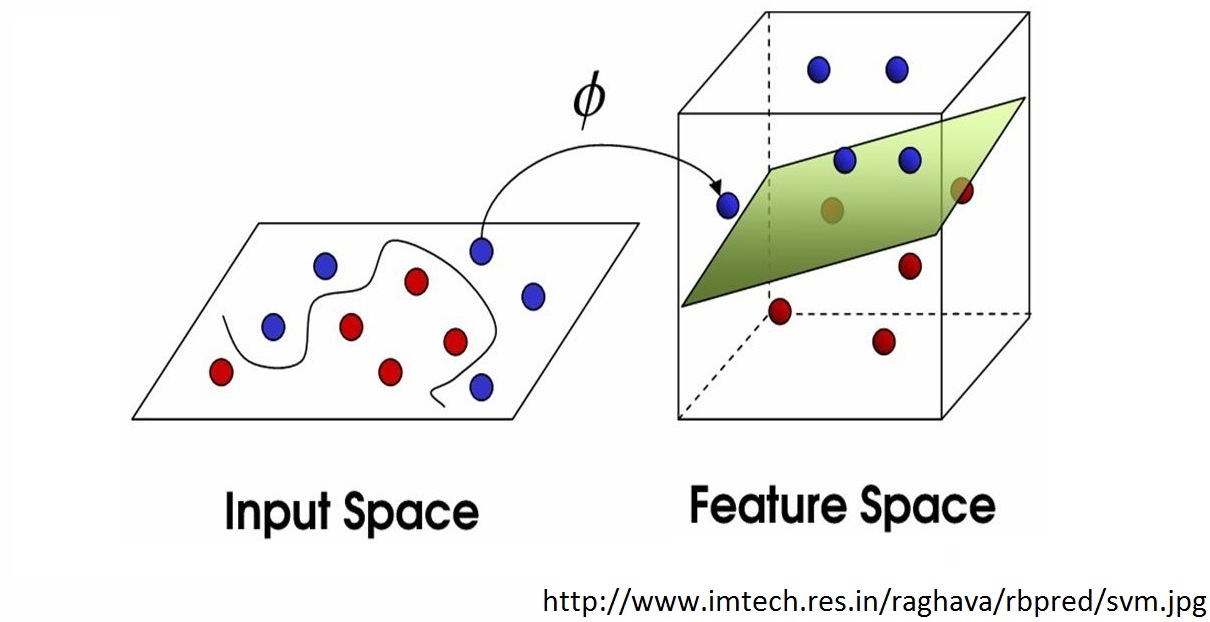}
  \caption{SVM main concepts.}
  \label{fig:svm}
\end{figure}

SVM has two parameters that are very important and directly influence its performance: \textit{cost~(c)} and \textit{gamma~(g)}. The cost parameter is responsible for evaluating the cost of a misclassification
in the training examples. A low cost may imply in a simpler surface, which may misclassify some training examples but avoid overfitting, \emph{i.e.}  a situation where the algorithm is adjusted to very specific features of the training data and does not generalize for additional data. On the other hand, a high value for this parameter may generate very specific surfaces, able to correctly classify all training examples but with limited generalization capability. The gamma parameter defines the influence of a support vector upon its surroundings. A low gamma means a higher influence, leading to a small number of support vectors, while a high gamma transforms each training vector in a support vector, since each vector has a small influence in the whole space.

Figure~\ref{fig:svm} shows the basic \textit{SVM} concepts. Once we have an input space, we apply a kernel function $\phi$ to map the input space into a feature space, where classification will be held. Classification is then performed finding an hyperplane that maximizes the distance between the instances (blue and red spheres) and the support vectors, that are derived from the hyperplane.

\subsection{Naive Bayes}

\textit{Naive Bayes} is a probabilistic classifier that assumes that all features~(inputs) are independent, generating a classifier that makes its prediction evaluating the probability of each class given an input. Although this assumption is unrealistic, \textit{Naive Bayes} performs well in a wide range of domains, apart from being fast. It can be represented by the following equation:

$$
P(\mathbf\chi|C) = \prod_{j=1}^d p(\chi_j|C),
$$

\noindent \noindent where $\mathbf\chi$ is the input and $C$ a multinomial variable taking a class code, as defined by \cite{Alpaydin_Book_10}.
Once the algorithm assumes a conditional independence, we can calculate the conditional distribution over the class $\mathbf C$ as the product of $p(\chi_j|C)$ for all $j$, then easily discovering the probability of being a specific class given the features input ($\chi_j$).

\subsection{JRip}

JRip is a Java implementation of RIPPER \citep{Cohen_ICML_95}, and follows a divide-and-conquer strategy that divides the input space into different regions, and finds rules for these regions. 

In this approach, rules with IF-THEN statements are learned, one at a time. The algorithm successively executes two different phases: (1) grow and (2) prune. \cite{Alpaydin_Book_10} states that ``we start with the case of two classes where we talk of positive and negative examples (...) Rules are added to explain positive examples such that if an instance is not covered by any rule, then it is classified as negative. So a rule when it matches is either correct (true positive), or it causes a false positive. (...) Once a rule is grown, it is pruned back by deleting conditions in reverse order, to find the rule that maximizes'' a metric called rule value metric, calculated using the number of true and false positives.

We have used the algorithm implemented in the Weka framework called JRip\footnote{\url{http://wiki.pentaho.com/display/DATAMINING/JRip}} \citep{Cohen_ICML_95}. It is important to stress that this algorithm is very interesting because it generates comprehensible knowledge. While the other algorithms generate mathematical models that can be hard to be understood, here the generated rules are quite easy to be understood.

\subsection{AdaBoost}

This algorithm is based on the idea of combining multiple learners\footnote{In fact, these classifiers are \textit{weak learners}, \emph{i.e.} simple classifiers with an accuracy higher than $\frac{1}{2}$. This means that, for a binary classification problem, they are still better than a random algorithm.} that complement each other in order to generate a classifier with higher accuracy.

Using $\chi_i$ to represent an arbitrary dimensional input and $d_i$ the prediction of a base learner, \cite{Alpaydin_Book_10} defines a Boosting algorithm, presenting an example, as follows: ``Given a large training set, we randomly divide it into three. We use $\chi_1$ and train $d_1$. We then take $\chi_2$ and feed it to $d_1$. We take all instances misclassified by $d_1$ and also as many instances on which $d_1$ is correct from $\chi_2$, and these together form the training set of $d_2$. We then take $\chi_3$ and feed it to $d_1$ and $d_2$. The instances on which $d_1$ and $d_2$ disagree form the training set of $d_3$. During testing, given an instance, we give it to $d_1$ and $d_2$; if they agree, that is the response, otherwise the response of $d_3$ is taken as the output.''

We use the \textit{AdaBoost} algorithm \citep{Freund_ICML_96}, an abbreviation for adaptive boosting. It differs from basic Boosting algorithms because it does not require a large training set to work properly, as it does not divide the dataset in disjoint sets. It uses the same dataset successively, giving different weights to each instance as it is misclassified or not.

\subsection{Summary}

This section discussed the main techniques used in this thesis. Its main goal was not to deeply discuss each technique, but to show that each algorithm is based in a different paradigm expecting a different characteristic from the dataset.

Our main concern was to highlight these differences between algorithms and show that our choice was not arbitrary. In summary, \textit{Naive Bayes} assumes that the different features that represent a player are independent, with no feature depending on the other. Another simple approach is \textit{JRip}, which requires each class to be distinguished by a set of simple rules. On the other hand, more complex approaches are \textit{SVM} and \textit{AdaBoost}. \textit{SVM} has the premise that different classes can be separated in the space using a kernel function (or linearly), while \textit{AdaBoost} combine different weak classifiers, focused on subparts of the input space, in order to generate a classifier with higher accuracy.

Since each algorithm has a premise that directly impact its performance, we evaluate all of them in our problem. Our results are presented, and further discussed, in Chapter~\ref{experimental_results}.

%% file: related_work.tex
\chapter{Related Work and Player Modeling Taxonomy}\label{related_work}

\epigraphhead[300]{\scriptsize You raise up your head\\
									 And you ask, ``Is this where it is?''\\
									 And somebody points to you and says ``It's his''\\
									 And you say, ``What's mine?''\\
									 And somebody else says, ``Where what is?''\\
									 And you say, ``Oh my God\\
									 Am I here all alone?''\\
									 \vspace{-0.6cm}
									 \begin{flushright}
									 Bob Dylan, \textit{Ballad of a Thin Man}
									 \end{flushright}
								   }
\vspace{3.8cm}

As previously stated in Chapter~\ref{introduction}, \textit{player modeling} is currently a very relevant topic in game AI research. Due to the attention it is receiving, several papers related to this topic are constantly being published. However, \textit{player modeling} is a loose concept and, until recently, the field was not completely structured. Papers used different terms for the same things, generating a lack of precise terminology. In this chapter we present some of the most relevant works in the field and propose a taxonomy to organize them, classifying the works discussed here in several facets.

\section{Related Work}

\cite{Slagle_CommACM_70} were the first to present an attempt to model players, but research specifically focused on \emph{player modeling} started in 1993, first aiming at improving search in game trees \citep{Carmel_AAAISymp_93}. At that time, the processing power available in computers was much smaller than today and, due to this limitation, the authors proposed modeling players as an alternative to better prune game trees in \textsc{Chess}. Another work, in that same year, which also studied the potential of player models in tree search is \citep{Iida_ICCA_93}. 

From this point, for almost one decade, the few studies in this area were applied to classical games such as \textsc{Chess} \citep{Carmel_AAAISymp_93}, \textsc{Go} \citep{Ramon_CGAT_02}, \textsc{RoShamBo}\footnote{The game \textsc{RoShamBo} is also known as Rock-Paper-Scissors} \citep{Billings_ICGA_00, Egnor_ICGA_00}, the iterated prisoner's dilemma~\citep{Kendall_Online_05} and \textsc{Poker} \citep{Billings_AAAI_98, Davidson_ICAI_00}. This scenario started to change with~\cite{Houlette_AIWisdom2_03}, who discussed the applicability of player modeling in more complex games, such as those in the FPS genre, and suggested a model to represent these players. \cite{Houlette_AIWisdom2_03} described this representation as ``a collection of numeric attributes, or \textit{traits}, that describe the playing style of an individual player''.

After Houlette's work, several other researchers started focusing on non-traditional games, such as FPS, RPG and Strategy games. With a broader scope, \emph{player modeling} started being used for different goals. We now discuss some recent works dealing with different applications where \emph{player modeling} can be used, including the generation of artificial intelligent opponents, game design and interactive storytelling.

\subsection{Artificial Intelligent Opponents}

The great majority of research related to \emph{player modeling} is in this topic, and the first works of the field, discussed at the beginning of this section \citep{Carmel_AAAISymp_93, Iida_ICCA_93, Ramon_CGAT_02}, can also be classified here. It is also sometimes called \textit{Opponent Modeling}.

A first research branch in \emph{player modeling} is related to games where AI is generally implemented using tree search algorithms such as \textit{MiniMax}. Games in this category are board and card games. Some of the most recent works focus on the game of \textsc{Poker}. \cite{Billings_Thesis_06, Aiolli_FrontEntComp_08} used a set of weights for each \textsc{Texas Hold'em Poker} possible player type and predicted one's choice of action as a weighted voting by all player types. On the other hand, \cite{Ponsen_AAAI_10} created a poker player with \textit{Monte-Carlo Tree Search} algorithms and used \emph{player modeling} to focus on relevant parts of the game tree. Another work, applied to \textsc{Heads Up Poker No Limit}, is \citep{Southey_UAI_05}, in which the authors ``present a Bayesian probabilistic model for a broad class of poker games, separating the uncertainty in the game dynamics from the uncertainty of the opponent's strategy''.

Besides \textit{player modeling} in tree search algorithms, a second research branch is related to games where tree search algorithms are not able to generate a satisfactory AI. This may be due to several reasons, such as the difficulty of modeling, in a good granularity, different game states and its transitions; or the size of the generated tree. For example, \cite{Aha_ICCBR_05} states that modern strategy games have a branching factor of approximately $1.5 \times 10^{3}$, while \textsc{Chess} has a branching factor close to $30$. Besides strategy games, other game genres where AI is not generated by search in the state space are action, adventure, RPG and sports. 

This second research branch is very generic, encompassing several different works and approaches. In order to better organize this section we have clustered works focusing on what characteristics of the opponent they aim at modeling. We discuss four possibilities here: (1) actions, (2) tactics/preferences, (3) movement/position and, (4) knowledge.

Most of the research in \emph{player modeling} is done with \textit{action models}, which are an attempt to model players' activities in a way that makes it possible to predict the next player's action \citep{Spronck_AIIDE_10} . Indeed, many works follow this line and \citep{Rohs_Thesis_07} is a classical example, as they try to predict if an agent will declare war against other in the game \textsc{Civilization IV}. Another important work is \citep{Laird_AGENTS_01} that anticipated actions in \textsc{Quake~II}.

Player's tactics/preferences are similar to the modeling of player's actions but, while the first models directly observable actions, this second modeling focuses on a goal, that will be reached by a set of small actions. This thesis is mainly focused on this task. 

Three works that modeled player's tactics/strategies are \citep{Heijden_CIG_08, Laviers_AIIDE_09, Weber_CIG_09}. In these works, the authors automatically inferred characters goals by observing their actions. \cite{Laviers_AIIDE_09} used \textit{Support Vector Machines (SVM)} to recognize a defensive play ``as quickly as possible in order to maximize (...) team's ability to intelligently respond with the best offense''. \cite{Heijden_CIG_08}, on the other hand, presented a method that obtains dynamic formations capable of adapting to the formation of the opponent player. This adaptation is performed by a learning algorithm, while the classification of the opponent is done with a set of steps that determine the likelihood of each opponent exhibiting the observed situation. Finally, \cite{Weber_CIG_09} applied multi-class classification techniques in \textsc{Starcraft} game replays to predict players' strategies. Their approach showed to be less susceptible to noise and imperfect information.

A research very related to ours is \citep{denTeuling_Thesis_10} and the paper generated from his master's thesis \citep{Spronck_AIIDE_10}. In this research, the authors modeled strategy preferences of virtual agents in the game \textsc{Civilization IV}, through a supervised learning technique called \textit{Sequential Minimal Optimization (SMO)} as a multi-class classification problem. After learning virtual agents preferences, they tried to identify virtual agents not present in the training and also human preferences. They did not have much success on this last task. The modeled preferences were: \textit{Culture, Gold, Growth, Military, Religion} and \textit{Science}. This is a very important work and we will further discuss it along this thesis. We use their dataset in our research and we use their work as baseline in some parts of this thesis. 

Modeling players position/movement is also a common approach in the literature. This approach may be seen as an attempt to present smart NPCs who do not break game rules (ignoring \textit{fog of war}\footnote{Parts of the game world where the player has no units visible to him/her, generating an environment with imperfect information. It is said that these invisible parts are hidden by a fog of war.}, for example), a common used resource, as \cite{Laird_AAAI_00} already discussed. Two recent works in this topic are \citep{Weber_AIIDE_11, Sukthankar_CIG_12}. \cite{Weber_AIIDE_11} modeled players movement/position, in the game \textsc{Starcraft}, using a particle based approach, while \citep{Sukthankar_CIG_12}
used ``inverse reinforcement learning to learn a player-specific motion model from sets of example traces''.
\cite{Valkenberg_Thesis_07} also worked in this problem trying to foresee players position in the game \textsc{World of Warcraft}. Despite the fact that he did not have much success, the problem he worked is an excellent example of the discussed topics. A more successful approach was presented in \citep{Hladky_CIG_08} for the game \textsc{Counter Strike: Source}. Other works that also predict the position of opponent players in FPS games are \citep{Laird_AGENTS_01, Darken_AIWisdom4_08}.

To conclude, it is also possible to model players' knowledge. A knowledge modeling was proposed by \cite{Cunha_CIE_13}, although the authors did not use this terminology. In this work, we developed an aid system to RTS players and one of its activities is to predict the technological level of an agent based on its units in the game \textsc{Wargus}. We implemented a reverse path in the dependency tree of the game, deriving what are the technologies known by the enemy once the player has seen a building or unit of his/her adversary.

Several other works were also successful in modeling players in order to improve their AI. We have presented some of the most recent and important papers in the field, and discussed common applications of this technique. Nevertheless, there is a large number of works that we did not cover, such as \citep{Lockett_GECCO_07, Schadd_GAMEON_07, Aiolli_IJCGT_09} among others.

\subsection{Game Design}

The main idea of \emph{player modeling} related to {\em Game Design} is to generate environments that are best suited to each player. This is one of the possibilities of game customization. Once one obtains a model of a player it may generate levels that maximize player's entertainment.

There are several possibilities for maximizing one's entertainment, such as identifying player's gameplay preferences in order to adapt the scenario for his/her style. For instance, once the game notes that a player likes to be a sniper in FPS games, it may generate spots for him/her to stay. The importance of including player models to procedural content generation has been discussed in \citep{Togelius_TCIAIG_11}.

An important work in the field of game design is \citep{Drachen_CIG_09}. In this paper, the authors use tools to extract gameplay information from the game \textsc{Tomb Raider: Underworld} and feed neural networks (\textit{emergent self-organizing maps}) with these data in order to obtain playing styles (\textit{Pacifist, Runner, Veteran} and \textit{Solver}). During this process, the authors extract several information to perform classification, e.g., \textit{cause of death, completion time} and \textit{number of deaths}. Using similar features, \cite{Mahlmann_CIG_10} predicted when players would stop playing the game or how long they would take to complete it. 

Recently, two other methods that cluster gameplay data in order to find players stereotypes in different games were proposed by \citep{Gow_TCIAIG_12, Drachen_CIG_12}. \cite{Gow_TCIAIG_12} apply Linear Discriminant Analysis (LDA) in data from two different games: \textsc{Snaketron} and \textsc{Rogue Trooper}, detecting different gameplay patterns and abilities. They also cluster data from the game \textsc{Rogue Trooper} finding four player's stereotypes: \textit{hiperactive, normal, timid} and \textit{naive}. \cite{Drachen_CIG_12}, on the other hand, uses k-Means and Simplex Volume Maximization to cluster players both from the MMORPG \textsc{Tera} and the FPS game \textsc{Battlefield 2: Bad Company 2}.

All these papers discussed above, despite of not directly changing the game scenario, are very useful for game designers since they give cues about playing patterns and preferences, which could change game design approaches.

Another work that studies game design improvement based on adaptive analysis is \citep{Pedersen_TCIAIG_10}. Using the game \textsc{Infinite Mario Bros} as a test platform, the authors look for correlations between players emotions (\textit{Fun, Challenge, Frustration, Predictability, Anxiety} and \textit{Boredom}) and level characteristics, e.g. \textit{presence of gaps, blocks} and \textit{enemies}. These correlations were evaluated with questionnaires.

There are several other papers in this game design context, such as \citep{Dormans_TCIAIG_11}, which discusses the use of generative grammars to create levels, and \citep{Yannakakis_TranAffectComp_11}, which presents a framework for procedural content generation driven by computational models of user experience.

Another goal of obtaining an opponent model may be challenge tailoring (CT) or dynamic difficult adjustment (DDA). \cite{Zook_AIIDE_12} distinguish these two problems: ``CT is similar to \textit{Dynamic Difficulty Adjustment (DDA)}, which only applies to online, real-time changes to game mechanics to balance difficulty. In contrast, CT generalizes DDA to both online and offline optimization of game content and is not limited to adapting game difficulty''. In their work, \cite{Zook_AIIDE_12} employed tensor factorization techniques for modeling player performance in an action RPG developed by them, also discussing some approaches to tackle the challenge tailoring problem. A work related to dynamic difficult adjustment is \citep{Missura_DS_09}. In this paper the authors automatically adjust the difficulty of a game implemented by them by clustering players into different types.

\subsection{Interactive Storytelling}

Besides customizing the game scenario or difficulty, another possibility is to customize the dramatic storyline of a game. This is called interactive storytelling. Another short definition of interactive storytelling was given by \cite{Thue_AIIDE_07}:

\begin{Exemplo}
``(...) a story-based experience in which the sequence of events that unfolds is determined while the player plays.'' \citep*{Thue_AIIDE_07}.
\end{Exemplo}

This application explicitly demands adaptive behavior since it is interactive. In spite of that it does not necessarily implements a customization for player preferences, although \cite{Sharma_AIIDE_07} showed that ``\emph{player modeling} is a key factor for the success of the Drama Management based approaches in interactive games.''

One of the first works to use information of specific players to generate a customized story is \citep{Thue_AIIDE_07}. In this work, the authors propose PaSSAGE, implemented in the game \textsc{Neverwinter Nights}. The method models interactive storytelling as a decision process that is influenced by different weights that characterize each player. The authors define their own method as ``an interactive storytelling system that uses player modeling to automatically learn a model of the player's preferred style of play, and then uses that model to dynamically select the content of an interactive story''.

Another important work regarding interactive storytelling and \emph{player modeling} is \citep{Sharma_AIIDE_07}. In this work the authors validate the important assumption that ``if the current player's actions follow a pattern that closely resembles the playing patterns of previous players, then their interestingness rating for stories would also closely match''. Additionally, they present features to differ player types, e.g. ``the average time taken by the player to perform an action in the game''. They also investigate important features that ``can be extracted from the player trace to improve the performance of the player preference modeling''.

In the same year, \cite{Roberts_AAAI_07} also developed a work related to interactive storytelling. The authors used player models to obtain a feature distribution regarding story characteristics. They calculated the importance of each feature for a player and learned a policy to select branches that custom storytelling for each player model.

A recent work regarding storytelling and some kind of player modeling was done by \cite{Cardona-Rivera_FDG_12}. The authors discussed the importance of differing between important and unimportant events when modeling a player's story comprehension.

Despite the applicability of \emph{player modeling} in interactive storytelling, there are not many works that have done that. This claim was done by \cite{Thue_AIIDE_07} and it seems to be still valid.

\section{Player Modeling Taxonomy}

As we already stated at the beginning of this chapter, \textit{player modeling} is currently a hot research topic, with several works being published in this field. Despite that, the field is not completely structured, since no organization is used to name the possible different approaches for this problem. Hence, our first contribution in this thesis is the proposal of a taxonomy, where we discuss research approaches and goals when modeling players. We first presented this taxonomy in \citep{Machado_CGames_11}. 

In this section, we first discuss some works related to our taxonomy, i.e. other proposals aiming at organizing the field. We then present our taxonomy, its classes and possible values. To help the reader to distinguish between each class, before its description we present a question that can be seen as a guideline to classify each work. Finally, Table~\ref{table:classification_taxonomy} shows some of the most relevant works in the field classified in our taxonomy.

\subsection{Taxonomy's Related Work}\label{sec:other_taxonomies}

A first more general work that presented a rough division organizing the field was \citep{Sharma_AIIDE_07}. In this paper the authors divided \emph{player modeling} simply by its measurements approach (direct or indirect). They state that direct measurements may use, for example, biometric data, while indirect measurements use in-game data (or \textit{game metrics}) gathered from observation. We focus this thesis only on this second category, \textit{i.e.} our taxonomy classifies works that use indirect measurements approaches.

Recently, at the same time we developed our \emph{player modeling} taxonomy, another research group, at University of California, Santa Cruz, independently developed theirs \citep{Smith_TechRep_11, Smith_FDG_11}. The authors divided the area in four main categories, which they called \textit{facets}, namely \textit{Domain, Purpose, Scope} and \textit{Source}: ``The \textit{Domain} facet of a model answers the question of what it is that the model generates or describes'' while ``The \textit{Purpose} of a model describes the function of a model in its intended application''. ``The \textit{Scope} of a model describes to whom the model is intended to be relevant or who is being distinguished in the model'' and ``Finally, the \textit{Source} facet describes how a player model is motivated or derived''~\citep{Smith_TechRep_11}.

Finally, \citep{Bakkes_EntComp_12} also reviews and organizes the works in the literature. The authors survey the field, classifying each work in one of four categories: \textit{Player Action Modeling}, \textit{Player Tactics Modeling}, \textit{Player Strategies Modeling} and \textit{Player Profiling}. 

The unique concept present in \citep{Bakkes_EntComp_12} that was not already discussed here is \textit{Player Profiling}. \cite{Bakkes_EntComp_12} defines it as the attempt ``to establish automatically psychologically or sociologically verified player profiles. Such models provide motives or explanations for observed behaviour, regardless whether it concerns strategic behaviour, tactival behaviour, or actions.'' To clarify this topic, we present a contrast between \emph{player modeling} and \emph{player profiling} presented by \cite{vanLankveld_CG_10}:

\begin{Exemplo}
``\textit{Player modeling} is a technique used to learn a player's tendencies through automatic observation in games \citep{Thue_AIIDE_07} (...) \textit{Player profiling} is the automated approach to personality profiling (...) In player profiling we look for correlations between the player's in game behavior and his scores on a personality test.'' \citep{vanLankveld_CG_10}
\end{Exemplo}

Since \emph{player profiling} is not the research subject in this work, we are not going to further discuss it. Some of the main researches in this field are \citep{Bohil_TechRep_07, Yannakakis_TCIAIG_09, Yannakakis_TransManCyber_09, Yannakakis_TranAffectComp_11, Lankveld_CIG_11, Spronck_AIIDE_12}.

\subsection{Proposed Taxonomy}\label{proposed_taxonomy}

Our taxonomy is composed of six different classes that encompass different aspects of a \textit{player modeling} research, ranging from high level concepts, such as what must be modeled, to implementation details. We discuss each of the six classes in the next sections, and we summarize its main components and possible values in Table~\ref{table:taxonomy}.

Note that we present several facets of a \emph{player modeling} work but, some of them have been already discussed by other researchers in different scenarios. Our main contribution was to select and organize different classifications in order to obtain a common taxonomy.

\begin{center}
\begin{table*}[t]
\caption{Player Modeling Taxonomy Summary.}
{\tiny
\hfill{}
\begin{tabular}{l l l l l l}
\textbf{Description} & \textbf{Time Frame} & \textbf{Goals} & \textbf{Applications} & \textbf{Methods} & \textbf{Implementation} \\
\hline
Knowledge     &Online Tracking              &Collaboration &Speculation in Search  &Action Modeling     &Explicit\\
Position      &Online Strategy Recognition  &Adversarial   &Tutoring               &Preference Modeling &Implicit\\
Strategy      &Off-line Review              &Storytelling  &Training               &Position Modeling   &\\
Satisfaction 	&                             &              &Substitution           &Knowledge Modeling  &\\
							&                             &              &Game Design 					 &									  &\\
\end{tabular}
}
\hfill{}
\label{table:taxonomy}
\end{table*}
\end{center}

\vspace{-1.5cm}

\subsubsection{Description}

\textit{What do you want to describe with your model?}\\

\vspace{-0.1cm}

Player modeling can be defined as an abstract description of the current state of a player at a moment. This description can be done in several ways like \textit{satisfaction}, \textit{knowledge}, \textit{position} and \textit{strategy} \citep{Herik_CIG_05}. 

The main goal of a game is to entertain its players, which are different from each other and may not enjoy the same challenges or possibilities of the game. When their \textit{satisfaction} is modeled, we may be able to adapt the gameplay to each player. This is called \textit{satisfaction modeling}.

More related to agents' artificial intelligence, we may want to model the player \textit{knowledge}, since this can be useful in several environments with imperfect information. A concrete example is found in games that have \textit{fog of war}, in which answers to the following questions can be very useful: which part of the map the player knows? He/She knows our position? In a game with constant evolution, which evolution level has already been achieved? All these questions may be answered with \textit{knowledge modeling}.

Similarly, we can model the player's movement. Once we are in a partially observable environment, the \textit{position} of other players is generally an important information since it can guide your strategy or actions. This is generally called \textit{position modeling}. 

Finally, a higher-level modeling is the \textit{strategy modeling}, which intends to interpret the player actions and relate them with game goals, {\em i.e.}, we abstract low-level actions seeking a high-level goal. For example, if we want to know our adversary aggressiveness, we will not be concerned with particular actions but with its strategy along the game.

\subsubsection{Time Frame}

\textit{When are you going to process the data? Do you have enough time for doing this online?}\\

\vspace{-0.1cm}

The different descriptions presented in the previous section lead us to different levels or categories of \textit{player modeling} use, and different moments to process the data. The lowest level of abstraction, with constant processing, is the \textit{Online Tracking}, which is concerned in predicting immediate future actions. 

A higher level (\textit{Online Strategy Recognition}) is related to strategy recognition as it involves the identification of a set of actions as a higher level objective or strategy. Finally, the \textit{Off-line Review} is the evaluation of a game log, after its finish. This last level is what many professional players do when they are ``studying'' their adversaries for a game. \cite{Laviers_AIIDE_09} discuss these three topics and argues that \textit{Online Tracking} is used for single players while \textit{Online Strategy Recognition} can be applied to entire teams. This is true since there is not a definition for a unique team's action. Nevertheless, it is important to note that we can also recognize strategies for individual players.

\subsubsection{Goals}

\textit{What are you intending to generate with Player Modeling? What is your main goal?}\\

\vspace{-0.1cm}

As previously mentioned, we firmly believe \textit{Player Modeling} is a suitable approach to improve the AI of NPCs in games. This improvement can be seen in different types of goals for the NPCs, that can be divided into three main sets: (1) to collaborate with the human players, (2) to be their opponents, or (3) to be neutral to them, but part of the story. 

The first set, related to the collaborative agents, is very difficult because human players have expectations when being aided by NPCs. These expectations are related to their actions: frequently, human players are not able to act properly because the NPCs will not behave accordingly as an unique team. Most of the games implement agents coordination and collaboration through basic orders such as ``Attack'', ``Patrol'' and ``Hide''. The main challenge would be to make these agents act autonomously according to the player behavior, without the need of specific orders.
 
The second set, the modeling of opponents, has motivation even on literature from centuries ago, as the famous Sun Tzu's quote.

\begin{Exemplo}
``Know your enemy and know yourself and you can fight a thousand battles without disaster.''
 \textit{Sun Tzu, The Art of War.}
\end{Exemplo}

In addition to the advices of an ancient general, it is very common for human players to study their adversaries before a match. Kasparov, the great chess player, is an example \citep{Carmel_AAAISymp_93}. Modeling opponents is a fundamental aspect in making games more immersive and challenging, so this is where most of the works in player modeling focuses.

One last possible goal developers can be concerned with is storytelling. In complex games, not all agents are allies or enemies. They can be neutral to players, being part of the scenario and interacting with them to help advancing the plot. Many times this interaction is offered to improve the storytelling of a game (in a medieval world not all people are warriors or mages, there are common people that should make the story more immersive) since once we model the player we may adjust all neutral agents to act properly to him.

\subsubsection{Applications}

\textit{What gameplay activity are you going to improve with your model?}\\

\vspace{-0.1cm}

In a lower abstraction level we can list, as \cite{Herik_CIG_05} discussed, four main applications to player modeling: \textit{speculation in heuristic search}, \textit{tutoring and training}, \textit{non-player characters} and \textit{multi-person games}. We renamed and redistributed them in terms of importance and generality. We renamed \textit{speculation in heuristic search} to \textit{speculation in search} and we split \textit{tutoring and training} as two different applications. Finally, we grouped \textit{non-player characters} and \textit{multi-person games} as \textit{substitution} application. We added a fifth application that is \textit{game design}. Several other applications can be listed but we believe that these five cover a satisfactory spectrum of them.

Speculation in search is generally applied to games in which Artificial Intelligence is more related to search in game trees, generally for adversarial goals. Depending on the game complexity, it may be infeasible to check every possibility and even pruning techniques such $\alpha-\beta$ are not sufficient. In these scenarios, we may use player modeling to create a bias that helps the search heuristics.

The collaborative goal can be expressed as the use of \emph{player modeling} to assist players. This can be be done with tutoring, when a non-human agent teaches the player (the \emph{player modeling} is important because this tutoring process can focus on the player preferences) or training, with the presentation of challenges suited to the player characteristics (weakness, style or strategy, for example). Its behavior is important because, if the NPCs do not act properly, the game will no longer be interesting to the player \citep{Scott_AIWisdom_02}.

Another main application to player modeling is in multi-player games. The objective is to allow NPCs to substitute human-players in multi-player games, even mimicking their behavior. It does not matter if the NPCs will be allies or enemies, they must be able to replace the player to keep the previous game balance. Most of the games does not have this approach and its gameplay may be impaired by players that are not able to play the whole game.

Finally, game design is applicable when the game does not want to generate NPCs behaviors, but to change its level or plot, for example.

\subsubsection{Methods}

\textit{What models are you intending to use to generate an understanding about the agent?}\\

\vspace{-0.1cm}

We can also divide the player modeling field in more specific methods, which are closely related to its description purpose. \cite{Spronck_AIIDE_10} mentioned that most of the research in player modeling is done with \textit{action models}, that are an attempt to model players' activities in a way that makes possible to predict the next player's action (\textit{Online Tracking}). Works that use a set of actions (not concerned in predicting the next atomic action) are also included here.

\cite{Spronck_AIIDE_10} define \textit{preference modeling} as the modeling of the ``player desires to accomplish or experience in the game, and to what extent he is able to do that''. This is a very precise definition and, in fact, is concerned to the player's satisfaction.

The last two listed methods are \textit{positioning} and \textit{knowledge modeling}. The first one attempts to obtain relevant information about players location while the second tries to model the player knowledge itself, it means, what he already knows.

\textit{Positioning modeling} can be better defined as the attempt to predict NPCs positions on games with imperfect information (\textit{fog of war}, for instance). This is a valuable information because, in general, the knowledge of a player position gives a tactical advantage in a game, as previously discussed here. This approach may be seen as an attempt to present smart NPCs who do not break game rules (ignoring \textit{fog of war} for example), a common used resource as \cite{Laird_AAAI_00} already discussed.

\textit{Knowledge modeling}, on the other hand, tries to ``predict''  the other players knowledge, humans or not.

It is important to stress the difference between the \textit{Description} and the \textit{Methods} classes, since they have similar possible values. The main difference is that description will use one of the methods to describe a player, e.g. one may model player's actions in order to describe its knowledge; or one may model players' knowledge through questionnaires, for example, aiming at describing players' satisfaction.

\subsubsection{Implementation}

\textit{What is the interface between your algorithms and the game in which your model is going to be used?}\\

\vspace{-0.1cm}

Once we defined some of the \textit{Player Modeling} subsets related to goals, applications, research areas, among others, we may finish this section with the lower abstraction level of discussion: the implementation. Two approaches can be highlighted: \textit{explicit} and \textit{implicit}. 

\cite{Spronck_IJCAI_05} says that ``An opponent \textit{[player]} model is explicit in game AI when a specification of the opponent's \textit{[player's]} attributes exists separately from the decision-making process''. Thus, an explicit player model is separated from the main source code and it is generally implemented through scripts or XML files. On the other hand, in \textit{implicit} approaches, the attributes are generally embedded and diluted in different parts of the code, which makes the task of identifying and describing these attributes more difficult.

\begin{landscape}

\begin{center}
\begin{table*}[ht]
\caption{Classification, in our taxonomy, of some works in the field.}
\hfill{}
\begin{tabular}{l | c c c c c c}
\textbf{Work} & \textbf{Description} & \textbf{Time Frame} & \textbf{Goals} & \textbf{Application} & \textbf{Methods} & \textbf{Implem.} \\[0.15cm]
\hline
\citep{Bard_AAAI_07}                &Strategy     &Online Strat. Rec. &Adversarial  &Spec. in Search &Action Model.     &Implicit \\[0.15cm]
\citep{Drachen_CIG_09}              &Strategy     &Off-line Review 	  &Storyt./Adv. &Game Design     &Action Model.     &Implicit \\[0.15cm]
\citep{Hladky_CIG_08}               &Position     &Online Strat. Rec. &Collab./Adv. &Substitution    &Position Model. &Implicit \\[0.15cm]
\citep{Laviers_AIIDE_09}            &Strategy 		&Online Strat. Rec. &Adversarial  &Substitution    &Action Model. &Implicit \\[0.15cm]
\citep{Machado_CIG_12}              &Strategy 		&Off-line Review    &Collab./Adv. &Substitution    &Preference Model. &Explicit\\[0.1cm]
\citep{Martinez_CIG_10}             &Satisfaction &Off-line Review    &Adversarial  &Game Design     &Action/Pref. Model. &Implicit \\[0.15cm]
\citep{Pedersen_TCIAIG_10}          &Satisfaction &Off-line Review    &Storytelling &Game Design     &Preference Model. &Implicit \\[0.15cm]
\citep{Spronck_AIIDE_10}            &Strategy 	  &Off-line Review    &Collab./Adv. &Substitution    &Preference Model. &Explicit\\[0.1cm]
\citep{Thue_AIIDE_07}               &Strategy     &Online Tracking    &Storytelling &Game Design     &Preference Model. &Explicit \\[0.15cm]
\citep{Valkenberg_Thesis_07}        &Position     &Online Strat. Rec. &Adversarial  &Spec. in Search &Position Model    &Implicit \\[0.15cm]
\citep{Yannakakis_TransManCyber_09} &Satisfaction &Off-line Review    &All 					&Game Design    &Action Model. &Implicit \\[0.15cm]
\end{tabular}
\vspace{0.2cm}
\hfill{}
\label{table:classification_taxonomy}
\end{table*}
\end{center}

\end{landscape}

%% file: generic_approach.tex
\chapter{A Generic Approach for Player Modeling as an ML Problem}\label{generic_approach}

\epigraphhead[280]{\scriptsize There are two ways of constructing a software design: one way is to
                               make it so simple that there are {\em obviously} no deficiencies
								               and the other way is to make it so complicated that there are no
								               {\em obvious} deficiencies.\\
									 \vspace{-0.6cm}
									 \begin{flushright}
									 C.A.R. Hoare ~The Emperor's Old Clothes'' CACM Feb 1981
									 \end{flushright}
								   }

\vspace{2.2cm}

This chapter discusses how \textit{player modeling} can be tackled as a Machine Learning (ML) problem, presenting a general methodology to apply it. In this methodology, a first step is to define how to represent different players. Hence this chapter also advocates for a generic representation of players proposed by \cite{Houlette_AIWisdom2_03}. The author introduced it almost as a theoretical model while this chapter discusses its applicability. The next chapter, focused on the instantiation of the methodology proposed here, evaluates the feasibility of the discussed representation.

This methodology was first presented in \citep{Machado_CIG_12}, while most of the discussions regarding the generic player representation are in \citep{Machado_SBGames_12}. 

\section{A Methodology for Preference Modeling as a Machine Learning Problem}

This section proposes a methodology able to model players following an ML approach. Several player's aspects can be modeled, such as actions, preferences and position. The methodology has six phases:

\begin{itemize}
  \item Define a representation for the player;
	\item Define relevant features according to the game;
	\item Select which relevant examples should be used;
	\item Model the \textit{player modeling} problem as an ML task, which can be supervised (as done in this thesis) or unsupervised;
	\item Select the appropriate algorithm(s); and
	\item Find the best parameters configuration for the selected algorithm(s).
\end{itemize}

All these topics are generically discussed here, since this approach may be applied to any game in order to define player models.

\subsection{Representation Definition}

As previously stated, a first general concern related to \emph{player modeling} is the representation of different players. It is important to be capable of representing different aspects of a player in the game, being able to obtain models for players with different preferences or knowledge, for instance. This representation is what will be accessed by the game AI in order to generate different behaviors, scenarios or plots, for example.

This step is determinant when defining an ML approach to \textit{player modeling} because the selected algorithm must be capable of generating the player's model, e.g. if one decides to represent a player as a tree, he may struggle to use a neural network to generate it. Hence, this task is required regardless of the approach, using ML or not. 

Section~\ref{sec:generic_representation} presents a generic representation that may be used to model players/agents.

\subsection{Features and Examples Definition}

Machine learning algorithms require a set of features as input. For the preference modeling problem, for instance, these features should be able to represent different behaviors of players with different preferences. This is based on the assumption that different behaviors are generated by different preferences. This approach is completely dependent of the game being used as testbed. Once the game platform is defined, a study of the selected data may be useful to assure that the assumptions of the data relevance are correct. We have performed all these steps in Chapter~\ref{player_modeling_civ4} when instantiating this methodology to the game \textsc{Civilization~IV}.

Despite being dependent of the game used as testbed, some general approaches may be common among games of the same genre. For example, strategy games generally present several game indicators along a match. These indicators are related to resources, military and technological characteristics of the game, and they are strong candidate features for an ML technique.

In a First-Person Shooter (FPS) game, indicators derived from players behavior observations, such as life time, the selected weapons and spots, number of kills, among others, may define a player preference. In the case of Action-adventure games, such as the \textsc{Tomb Raider} series, player preferences may be defined by other indicators. Among them we can list causes of death, total number of deaths, completion time and demand for help. In fact, all these indicators were used by \cite{Drachen_CIG_09} to define models of players in the game \textsc{Tomb Raider: Underworld}.

Although feature definition is essential for any game genre, it is important to stress that the data availability to researchers may vary among different game types. This is mainly because different game developers have different approaches to data extraction. While some allow scripts and even source code modifications, others are extremely reluctant to permit any interaction with the game, other than playing it.

Apart from defining sets of attributes, decisions on how much data to use may also be crucial. For instance, in turn-based games such as \textsc{Civilization IV}, \cite{Spronck_AIIDE_10} suggested that the first turns should be ignored, because the indicators of different players evolve similarly at the beginning and are not useful to distinguish preferences.

\subsection{Problem Modeling and Appropriate Algorithms}\label{sec:generic_problem_modeling}

Once we define relevant features and examples for modeling a player, we need to decide how to represent different players. This representation depends on two different aspects: if we are able to previously identify the players' models and what algorithm we will select for the task.

All three ML approaches discussed in Chapter~\ref{background} (supervised, semi-supervised and unsupervised) may be applied to \textit{player modeling}. If the player information is not previously known, unsupervised learning may fit better. In this case, players are clustered according to similar attributes using a distance metric, and the researcher is expected to identify each group by its characteristics. An example of this approach can be found in \citep{Drachen_CIG_09}.

On the other hand, if players' models are previously known, we can classify new players based on them. This is the approach presented in this thesis, which uses the preferences of \textsc{Civilization IV}'s virtual agents to identify players' preferences. In general, classifiers are appropriate algorithms for this task.

There are many ways to model this problem using classifiers. Perhaps the most intuitive approach is to model the obtainment of each possible characteristic as a classification problem. Using preference modeling to simplify this discussion, working with concrete concepts, we can say that modeling $n$ preferences requires $n$ different classifiers. For each classifier, we can predict different types of things. One approach is to predict different levels of preference. For example, the player can be considered to have no preference, weak preference, or strong preference for a specific topic. This is the approach followed in \citep{denTeuling_Thesis_10, Spronck_AIIDE_10}, called a multi-class problem, in which three classes were considered. We believe that the main problem with this approach is that human players have difficulties in defining their preferences in terms of levels, it is much easier to say simply if they have a preference or not. Furthermore, obtaining datasets detailing different levels can be more complicated than knowing if a player has a preference or not.

Given these drawbacks, an alternative approach that is used in this work, is to model the problem using a binary classification strategy, in which we want to find if the player simply has or has not a preference for a given characteristic. Hence, a binary classifier will predict if a player (example) belongs to the class or not. 

Another possible way of modeling this problem is to consider it as a multi-label classification problem. In this case, only one classifier is used, regardless of the number of preferences being modeled. For each example, we could have many different classes (that is the reason for the name multi-label), and a single classifier would be able to predict all of them. For the best of our knowledge, this approach has not been tested for player modeling so far.

Regarding the algorithms used, choosing the best classifier is an open question \citep{Bradzil_Book_09} and, as discussed in the next chapter, we do not tackle this problem here. This problem is generally studied by the sub-area of meta-learning. 

\subsection{Parameters Configuration}

ML algorithms are very sensitive to parameters. Hence, it is very important for researchers to spend some time tuning the algorithm parameters. A common and simple approach for this task, when applied to classifiers, is to perform a grid search on the relevant parameters. We further discuss this topic in the next chapter.

If the researcher does not know the most relevant parameters to be studied, it may be useful to apply some technique that may assist him/her to select them. A common approach is called $2^k$ Factorial Design, which ``is used to determine the effect of $k$ factors, each of which have two alternatives or levels'' \citep{Jain_Book_91}.

An important point when talking about parameter tuning is computational time. The higher the number of examples used for training, the worse the computational time. A simple solution to this problem is to sample the dataset preserving its original features, such as class distribution. 

\section{A Generic's Player Representation}\label{sec:generic_representation}

Just as a generic methodology for \textit{player modeling} is a valuable discussion, a generic representation that is able to model different aspects of a player is a useful resource. We advocate here for a representation proposed by \cite{Houlette_AIWisdom2_03}, who did not present any real implementation or evaluation of its feasibility. Thus, our main contribution here is to present a discussion regarding its applicability, theoretically and in the industry, based on several industry requirements listed by \cite{Isla_GDC_05}. 

Besides \citep{Houlette_AIWisdom2_03}, as far as we know, one of the few papers that discusses a general methodology for obtaining models through different environments is \citep{Charles_CGAIDE_04}. In spite of that, it does not validate its assumptions in a real game. A sequence of this work is presented in \citep{Charles_Digra_05}, in which the authors also discuss a high-level framework for adaptive game AI. They briefly present an approach for \textit{player modeling} with factorial models but do not investigate further. A more recent work that somehow also discussed a generic representation for agents intentions is \citep{Doirado_AAMAS_10}. This work only deals with actions, proposing a framework called \textit{DogMate}. The authors left more general concepts such as preferences out, limiting their model.

\subsection{Houlette's Representation}

The representation discussed here to model players is based on two main components: a set of variables representing specific features of the game and a set of weights that multiply these variables. The set of variables is determined by the game designer and the AI programmer, and it is based on the {\em domain knowledge} of each game, an inevitable requirement for adaptive AI as \cite{Spronck_IJCAI_05} and \cite{Bakkes_TCIAIG_09} previously stated. The weights represent the importance that is given to the feature represented by that variable. They can be manually set by the designer / programmer or learned from experience. Completely different behaviors can be obtained varying the weights of each player, which allows players to adapt to different game conditions.

More formally, the model is represented as a set of $n$ weights, in which $n$ is the number of characteristics that are going to be modeled:
$$
Pm = \langle w_0, w_1, ..., w_i \rangle
$$
where $w_i$ is a weight for characteristic $c_i$ of the model $Pm$.

This representation is generic and can be applied to various games such as \textsc{Chess}, \textsc{Poker}, FPS and Strategy games. It is very simple, yet powerful enough to represent and model different player behaviors. The main characteristic of this representation is that it is defined as a vector of weights for different game features. This modeling is generic but the model's features/variables are particular to each game and must be defined by an expert. The main advantage of this approach is that, once techniques are created to infer weights, they can be applied to different games that use this approach.

As an example we may use \textsc{Poker} to illustrate our discussion. \textsc{Poker} is a very hard game for computers to play and, so far, they are not able to defeat the best human players. A promising approach is player modeling and there are many works in this area such as \citep{Billings_AAAI_98, Davidson_ICAI_00, Southey_UAI_05, Bard_AAAI_07}.

To apply the proposed representation to model \textsc{Poker} players, it is necessary to represent different player behaviors. A very simplistic model, created just for this example, can be extracted analyzing relevant game features such as agents temerity ($T_e$), probabilistic capacity ($P_c$) and bluff tendency ($B_t$). We could define each of these variables as: 

\begin{itemize}
\item $T_e$ representing the agents tendency to take risks;
\item $P_c$ representing agent's awareness about the probability of winning given a card distribution in a match;
\item $B_t$ describing how much does an specific agent bluffs during a game.
\end{itemize}

Once we created this simple set of variables, we generate the following representation, capable of modeling completely different players:
$$
Pm = \langle T_e, P_c, B_t \rangle.
$$

To exemplify, defining weights between $0$ and $1$ and a higher value representing a stronger preference, we could easily generate an expert conservative player, represented as $ Pm_1 = \langle 0.1, 0.9, 0.2\rangle$ or an aggressive player, $ Pm_2 = \langle0.8, 0.6, 0.5\rangle$. 

Thus, this representation implies in two different tasks to generate a player model:

\begin{enumerate}
\item To define the model variables: with the \textit{domain knowledge} of the specific genre or game, define what are the relevant features to be modeled and in what level of abstraction it will be done.
\item To define the variables weights: this  is what distinguishes behaviors, {\em i.e.}, different behaviors are set in this second step. It can be done to model non-player characters as well as human players. 
\end{enumerate}

The definition of variables is an easier task since we just need to define what we want to model, but the weights setup is a very hard task for several reasons. Maybe the biggest one is the fact that there is no rule of thumb for doing it. Generating these weights (or other type of modeling) from repeated play, observing players and other virtual agents and tracking game results, often involves the use of machine learning techniques and the most used are those inspired in nature as neural networks or genetic algorithms.

After discussing this representation applicability, our focus will be in instantiating the generic methodology proposed here to model players' preferences in the game \textsc{Civilization IV}. We will use this representation, then our focus on this thesis will be on setting the weights to represent both virtual agents and human players preferences.

\subsection{Applicability}\label{sec:applicability}

A useful model must be capable of satisfactorily representing different agents with different characteristics. This topic includes the capacity to allow {\em any} behavior just deriving it from the model, what could be seen as a coverage requirement. As we already discussed, once appropriate features are selected, this is completely feasible.

Additionally, it is desirable that it is applicable in industry, what would certify its validity. We discuss both topics below.

\subsubsection{Representativeness}\label{sec:represent}

Regarding the representativeness of a model,  we consider that it would be effective if capable of performing two different tasks:

\begin{enumerate} 
\item The generation of different behaviors by varying the model. In this approach, different weights generate different behaviors; and
\item A model that generates a specific behavior must be ``inferable'', i.e., one must be able to infer a model that generates an observed behavior. This task, in this modeling, consists in inferring variables weights from observation.
\end{enumerate}

The requirements listed above create a cycle: once we observe a specific behavior we must be able to infer the model which generated it and we must be capable of generating different behaviors from the model.

This evaluation must be performed in the specific game that is being used for modeling players. We validate it for the game \textsc{Civilization IV} in the next chapter, corroborating our assumption about this representation usefulness, since we were able to perform both tasks.

\subsubsection{Applicability in Commercial Games}

An important topic to discuss before presenting the instantiation of the discussed methodology and representation, is related to the applicability of the representation in commercial games. In general, the game industry is somewhat reluctant about the game AI solutions proposed by academy. As discussed by \cite{Fairclough_ICAICS_01, Laird_AIMagazine_01, Nareyek_Queue_04, Bakkes_TCIAIG_09}, there are several reasons for this: the concern about unpredictable behavior, the necessity of heavy modification and specialization for each game, and the difficulty in understanding the reasons for some observed behaviors and in modifying any configuration. 

We argue that due to its simplicity and expressiveness, the representation discussed here avoids most of these problems. We use as a base for this discussion the work of \cite{Isla_GDC_05}, which presents the AI architecture of a very well-succeeded game, \textsc{Halo~2}\footnote{Bungie Studios, Microsoft Games Studios, Halo 2 (2004): \url{http://www.microsoft.com/games/halo2/}}, and discusses several design principles that should be targeted when developing the AI of a complex game.

First of all, Isla defines some basic AI requirements: {\em coherence, transparency, run-time, mental-bandwidth, usability, variety} and {\em variability}. Coherence is related to actions' selection at appropriate times, {\em i.e.}, how we select actions once we have defined an agent model. Thus, coherence is much more related to the mapping between agent models and actions than to the model itself. On the other hand, transparency is one of the major points of this representation, since we are able to understand agents and even try to predict their behaviors only by observing the weights of each modeled feature. This is also related to mental-bandwidth, since we need to ``reason about what's going on in the system''~\citep{Isla_GDC_05}.

The representation is also important and useful to level designers since it allows usability (the characters are configurable since we only need to change their weights) and variety (the AI works differently for each different weight combination). Variability, as many of the discussed requirements, is related to the use of the model. It is important to observe that, once this model is defined, it is necessary that game programmers use the weights paradigm to develop their games. It is also important to note that this representation does not limit or hinder any desirable feature as neural network or kernel machine models generally do. Moreover, this approach, due to its simplicity, does not impact game performance.

Once we discussed the basic AI requirements, we are able to show that this representation goes further and also meets the design principles proposed by~\cite{Isla_GDC_05}. The first one is {\em ``Everything is customizable''}, {\em i.e.}, the model should be general enough to allow modifications in behaviors. This is exactly the main advantage of this representation since it can be seen as an organized set of parameters, a clear definition of customization.

The second design principle, {\em ``Value explicitness above all other things''} is obviously met by the explicit use of weights. Also, the use of weights simplifies the generation and test of specific behaviors and makes easier the process of generating relatively similar behaviors with some small variations in each weight. This is exactly Isla's fourth principle: {\em ``Take something that works and then vary from it''}. (The third principle, {\em ``Hackability is key''}, is not applicable directly to the representation but to other programming levels).

It is interesting to note that, despite the independent development of~\citep{Houlette_AIWisdom2_03} and~\citep{Isla_GDC_05}, it seems they were developed together because of the similarity in most of the discussed requirements. A final sentence of~\cite{Isla_GDC_05} evidences its usefulness: ``...~we are not interested in a scripting system in which the designer specifies EVERYTHING the AI does and where it goes - that would be too complex. We do need, however, the AI to be able to handle high-level direction: direct them to behave generally aggressively, or generally cowardly.''.

Hence it is clear that this representation is applicable in the industry. After defining a generic methodology and representation to \textit{player modeling}, we must instantiate it to our problem, the preference modeling in \textsc{Civilization IV}. This is the subject of the next chapter.

%% file: player_modeling_civ4.tex
\chapter{Player Modeling in Civilization IV}\label{player_modeling_civ4}

\epigraphhead[230]{\scriptsize``It's a job that's never started that takes the longest to finish.''\\
									 \vspace{-0.6cm}
									 \begin{flushright}
									 J. R. R. Tolkien, The Fellowship of the Ring
									 \end{flushright}
								   }
\vspace{1.2cm}

This chapter instantiates the generic ML approach proposed in previous chapter to the game \textsc{Civilization IV}. This instantiation takes into account our final goal, which is to model/predict the \textsc{Civilization IV} player's preferences, namely \textit{Culture, Gold, Growth, Military, Religion} and \textit{Science}.

Here we present the steps required prior to the execution of the ML algorithm, which final results are presented in Chapter~\ref{experimental_results}. Recall that the proposed methodology had six phases: (1) defining a representation for the player; (2) defining relevant features according to the game; (3) selecting relevant examples that will be used; (4) modeling the problem as a ML task; (5) selecting the appropriate algorithms; and (6) selecting best parameters configuration. 

Considering these phases, we first discuss the dataset that is used in our experiments, generated from \textsc{Civilization~IV}'s matches played by virtual agents. We then perform an evaluation of the available data, validating its relevance and using it to select the appropriate algorithms to model players' preferences. Finally, we evaluate the representativeness of the generic player representation, presented in Chapter~\ref{generic_approach}, using the game \textsc{Civilization IV}.

\section{Dataset}\label{sec:dataset}

Chapter~\ref{background} presented the game \textsc{Civilization~IV}, its mechanics and programming interface, discussing the script \textit{AIAutoPlay} that allows us to log matches between two virtual agents. Based on these topics, we are able to sniffer the game and capture data while virtual agents (or human players) play. Using this feature, \cite{denTeuling_Thesis_10} was able to generate a dataset to classify players preferences (we will further discuss this classification approach in the next sections). We present the dataset at this point because, from now on, most of our results are obtained from it.

Three different datasets were used in this thesis, two composed of virtual agents gameplay data, which we discuss here, and other created with human players data. The first two were created by \cite{denTeuling_Thesis_10} and edited here to our purpose. The third will be discussed in Chapter~\ref{experimental_results}.

The first decision to be made in order to sniffer gameplay is when to capture data. In the case of \textsc{Civilization IV}, the game structure eases this decision because its turn-based pace clearly defines the data collection moment, at the end of each turn.

As we previously said, \cite{denTeuling_Thesis_10} used the script \textit{AiAutoPlay} to generate the dataset. This script allows the game to be played by two virtual agents, removing the requirement of human players. This is an important feature because it allows the dataset to have hundreds of matches, what would be impossible if human players had to play these matches. \cite{denTeuling_Thesis_10} modified this script to collect, for each turn, a set of game indicators. We use this same dataset, and its main features are described next.

The first dataset, called \textit{Traditional Dataset}, was generated by randomly selecting six leaders in the game, and making them play against each other eight times, in a total of 40 games per leader. For each turn, information of each agent was collected. At the end, the shorter game had 240 turns while the longer took 460 turns (the maximum allowed).

From all the data collected, we kept the same 21 features used in~\citep{denTeuling_Thesis_10, Spronck_AIIDE_10}, which are game indicators available to every player during the game. These indicators are scores and counters, modified by players actions, and here we refer to them as features. They all describe some aspect of the game, and a subset is presented in Table~\ref{tb:features}. A complete list is presented in Appendix~\ref{appendix_features}.

\begin{table}[th]
{
\caption{Subset of features and their meanings. The features are related to one player, e.g., the number of units indicator is the number of units that a specific player has.}
\label{tb:features}
\hfill{}
\begin{tabular}{l l}
    \noalign{\hrule height 1.2pt}
    Feature &Meaning\\
    \hline
    Turn              & Turn number \\
    War               & 0 = not in war; 1 = in war \\
    Cities            & Number of cities \\
    Units             & Number of units \\
    Economy           & Overall economic score \\
    Industry          & Overall industrial score \\
    Culture           & Overall cultural score \\
    Maintenance       & Gold needed for maintenance per turn \\
    ResearchRate      & Amount of research gained per turn\\
    CultureRate       & Amount of culture gained per turn\\
    \hline
\end{tabular}}
\hfill{}
\end{table}

Notice that the decision to use the same features used by \cite{denTeuling_Thesis_10, Spronck_AIIDE_10} was not made without consideration. In Section~\ref{sec:characterization} we show that some indicators are able to distinguish different behaviors, which supports our decision. In fact, these features were used in all three mentioned datasets.

\cite{denTeuling_Thesis_10} modeled each turn as a vector (example), and since evolution along the match is an important factor, they extended the set of basic features to add this notion of time. The authors name these new features, which are are presented in Table~\ref{tb:modifications}, \textit{composite features}.

A second dataset, also related to virtual agents, was created to evaluate generalization. We call it \textit{Alternative Dataset}. It was created using a different set of six agents and we will further discuss its use in Chapter~\ref{experimental_results}, where it is used.

\begin{table*}[th]
{
\small
\caption{List of composite features, as in \citep{denTeuling_Thesis_10}.}
\label{tb:modifications}
\hfill{}
\begin{tabular}{l c l}
    \noalign{\hrule height 1.2pt}
    Modification &Calculation &Meaning\\
    \hline
    Derivate          & $v_t - v_{t-1}$                                 & Increase or decrease in the base feature per turn \\
    Trend             & $\frac{(\sum_{i=0}^4 v_{t-i})}{5}$              & Average of base feature over multiple turns \\
    TrendDerivate     & $v_t - v_{t-5}$                     & Derivate of the trend \\
    Diff              & $v_t - w_t$                                     & Difference of the base feature with the opponent's \\
    DiffDerivate      & $(v_t - w_t) - (v_{t-1} - w_{t-1})$             & Derivate of the difference \\
    DiffTrend         & $\frac{(\sum_{i=0}^4 v_{t-i} - w_{t-i})}{5}$    & Trend of the difference \\
    DiffTrendDerivate & $(v_t - w_t) - (v_{t-5} - w_{t-5})$ & Derivate of the trend of the difference \\
    \hline
\end{tabular}}
\hfill{}
\end{table*}

Each different virtual agent has a specific set of preferences, which are defined by different values representing levels (0 -- no preference; 2 -- medium preference; 5 -- high preference). For our classification experiments, as we will mention when discussing our ML approach, we have removed the first 100 turns of each match to make our results comparable to our baseline.

\section{Features Definition}\label{sec:characterization}

A first important step when discussing the available data is to better understand virtual agents behavior and how they are expressed in game data. The evaluation of the data generated from gameplay of different players is what we use to define our features in the proposed ML approach.

In order to perform this task, we model the game \textsc{Civilization IV} as a set of states, where each state is defined by the data gathered at the end of each player turn. These data consist of several game \textit{information} like, for example, the amount of gold a civilization has or the number of cities. Moreover, each virtual agent in the game may have different \textit{preferences}, which are descriptions of the way the agents play, {\em i.e.} their main priorities during the game such as gold, culture or religion, as discussed in Chapter~\ref{background}. 

To evaluate the usefulness of the available gameplay data, and consequently to define our features, some questions must be answered:

\begin{itemize} 
 \item The information of intermediate states of the game do characterize distinct preferences of different agents?
 \item What available information distinguish the agents preferences? What is the relation between their predefined attributes and this information?
\end{itemize}

In this section we answer these questions with a characterization of the behavior of AI controlled agents, looking for relations between the agents predefined preferences and their behavior. The discussions presented here are derived from the results in \citep{Machado_SBGames_11}.

\subsection{Methodology}

Our objective is to characterize the behavior of different agents by its gameplay data, and to correlate them with their preferences. This is done by generating linear regressions based on game state indicators, gathered in several matches played between different AI agents. Our intuition was that we would be able to find different functions describing game data for different agents, since they have different preferences, what would justify the selection of this data as a feature.

To perform this evaluation we used a subset of the \textit{Traditional Dataset}, discussed in Section~\ref{sec:dataset}. We studied three agents preferences: \textit{Culture}, \textit{Gold} and \textit{Growth}. The characterization was performed by observing games between two different agents and analyzing the data generated by these observations. We have carefully chosen these agents in a way that one of them has no interest in a certain preference and the other has high interest in this same preference (values~0 and~5 in the game, respectively). This was done to simplify the comparison between indicators that were supposed to indirectly represent preferences, {\em i.e.}, we expected a higher value for an agent indicator that has a high interest in the preference related to that indicator.

For example, the agent called \textit{Mansa Musa} has a high interest in \textit{Gold} while the agent \textit{Louis XIV} has no interest in it. Based on this fact, we compared some of their indicators to model their \textit{Gold} preference, looking for different functions to each agent. In fact, we expect \textit{Louis XIV}'s indicators related to gold to be lower than \textit{Mansa Musa}'s indicators since \textit{Mansa Musa} has a higher preference.

We start from the premise that the adversarial agent actions do not impact the state of the player we are analyzing, and we characterize several of its behaviors and preferences. After this first phase, we relax this premise and we observe that it is correct, as we will show in Section~\ref{sec:victory_defeat}. This result give us confidence to assume this independence when applying our ML approach.

We have used the leaders in the example above to analyze \textit{Gold} preference (\textit{Louis XIV} and \textit{Mansa Musa}). To analyze the \textit{Growth} and \textit{Culture} preferences, we have used the agents \textit{Alexander} and \textit{Hatshepsut}. The \textit{Growth} preference has a peculiarity: in our dataset there was no agent with a high interest on this preference (value 5), just an average interest (value 2). We used the agent \textit{Alexander} as the one having interest on it while \textit{Hatshepsut} was the one who has no interest.

In all analyses, we have characterized each preference comparing the agents states in each turn, seeking for a function capable of representing this evolution. We did linear regressions in the data and, when the data did not fit in this model, we applied transformations on it to be able to use a linear regression, since the mathematical analysis is simpler and it does not imply in a loss of generality. A liner regression generates functions in the form $y = b_0 + b_1 x$. Our main concern is $b_1$, since it represents the evolution of the indicators in the game.

We have summarized the agents states calculating, for each turn and for each indicator, the average of 40 matches (the amount of matches each agent played in the used dataset). At the end, we had 460 points where each point $p_i$ represented the mean of turn $i$ for all agent matches. We have selected some indicators collected during gameplay under the premise that they would be relevant features to distinguish the studied preferences. These indicators were selected intuitively based on our knowledge about the game. All the regression algorithms and evaluation metrics used here are discussed in~\citep{Jain_Book_91}.

After the characterization of the three listed preferences, we separated matches by their results: victory or defeat. We revisited every analysis, using two different sets to understand the game result impact in our characterization, allowing us to answer the question whether the result influences the analyzed functions.

Next we will present the characterization of each modeled preference. After this first analysis, in Section~\ref{sec:victory_defeat}, we analyze the impact of separating games by their result.

\subsection{Agents Characterization}

This section presents the characterization of three different preferences, evaluating the use of seven different features. The preferences characterized using the mentioned features are: \textit{Culture, Growth} and \textit{Gold}.

We summarize regressions data in Appendix~\ref{app:regression}, in which we present each obtained coefficient, their confidence interval, and the regression's coefficient of determination\footnote{``The fraction of the variation that is explained determines the goodness of the regression and is called the coefficient of determination, $R^2$'' \citep{Jain_Book_91}.}.

\subsubsection{\textit{Culture} Preference}

We have selected two indicators (features) to distinguish this preference: \textit{Culture} and \textit{CultureRate}. As previously discussed, the characterization was done using the agents \textit{Alexander} and \textit{Hatshepsut}. The indicators are defined in \citep{denTeuling_Thesis_10} as being the ``Overall cultural score'' and the ``Amount of culture gained per turn'', respectively.

We were able to characterize almost perfectly this preference with the two selected indicators. To do it we have modeled the \textit{Culture} indicator as a polynomial of degree five, and \textit{CultureRate} as a polynomial of degree four. This was very satisfying since it is the order of the derivative of the polynomial that represents the \textit{Culture} (as expected, we have tested  regressions of these indicators to other functions, we decided to present only the best result). As we discussed in the previous section, a linear regression simplifies this analysis without loss of generality, so we applied the fifth root to all values of \textit{Culture} and the fourth root to all values of \textit{CultureRate}.

\begin{figure}[t]
  \centering
  \subfloat[Alexander]{\label{fig:culture_geral_Alexander}\includegraphics[width=0.5\columnwidth]{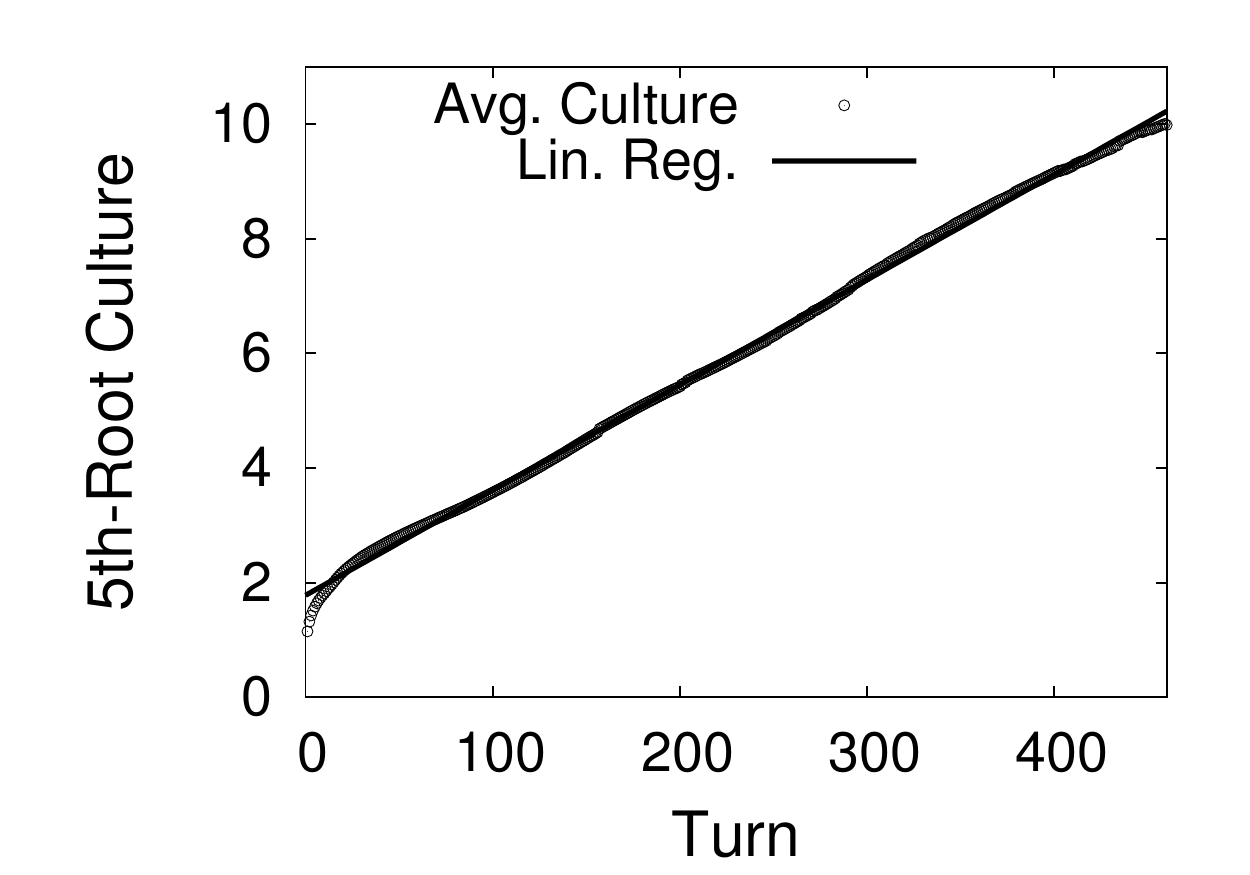}}               
  \subfloat[Hatshepsut]{\label{fig:culture_geral_Hatshepsut}\includegraphics[width=0.5\columnwidth]{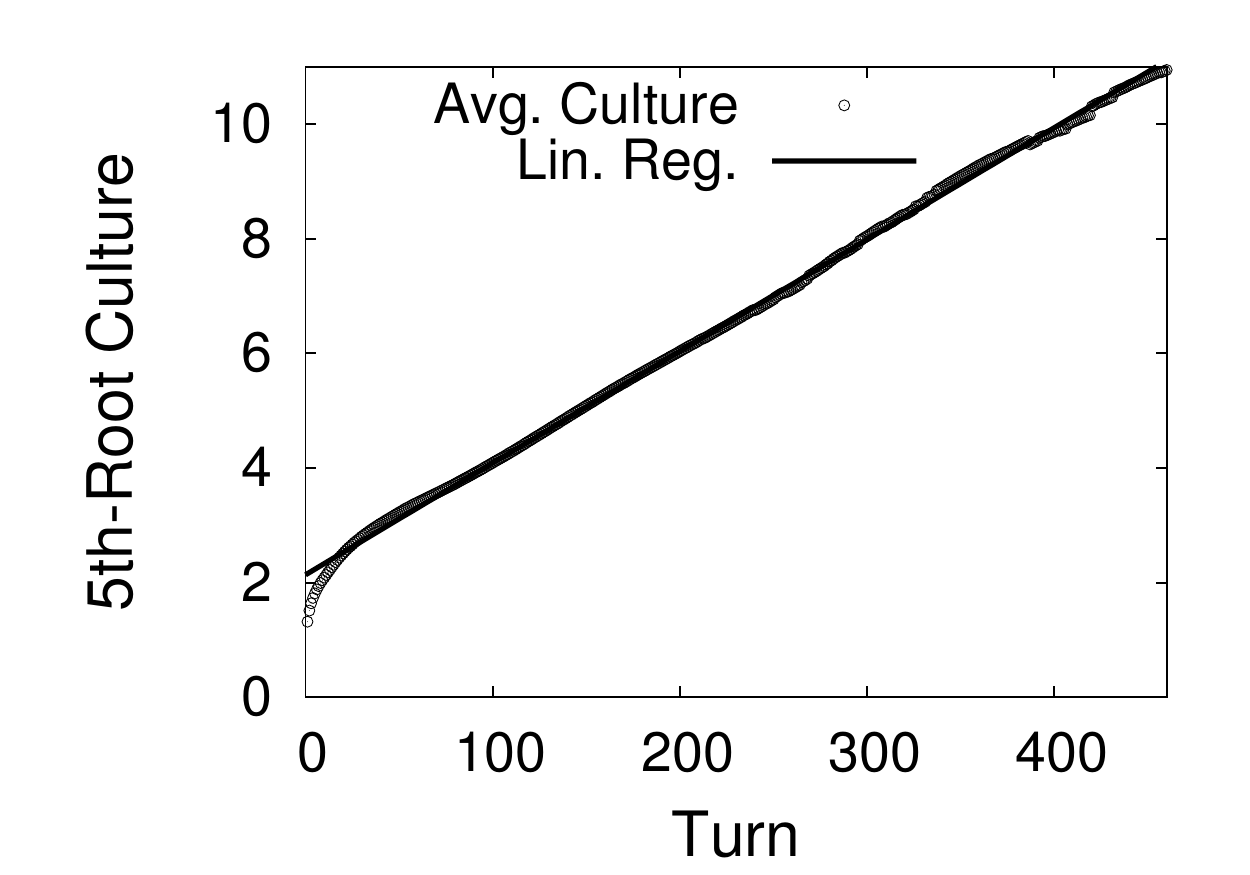}}
  \caption{Linear Regression of the $\sqrt[5]{Culture}$ indicator.}
  \label{fig:culture_geral}
\end{figure}

\begin{figure}[t]
  \centering
  \subfloat[Alexander]{\label{fig:cultureRate_geral_Alexander}\includegraphics[width=0.5\columnwidth]{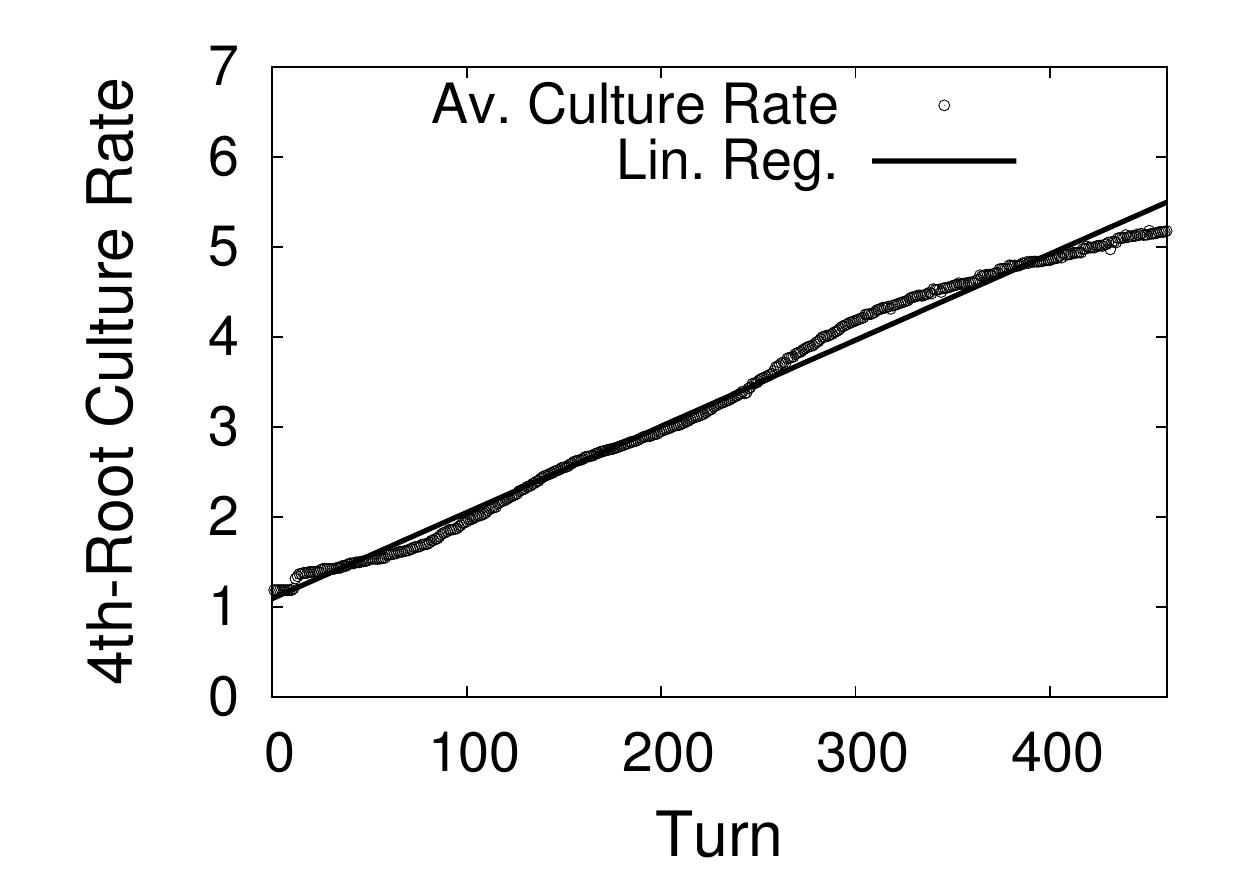}} 
\subfloat[Hatshepsut]{\label{fig:cultureRate_geral_Hatshepsut}\includegraphics[width=0.5\columnwidth]{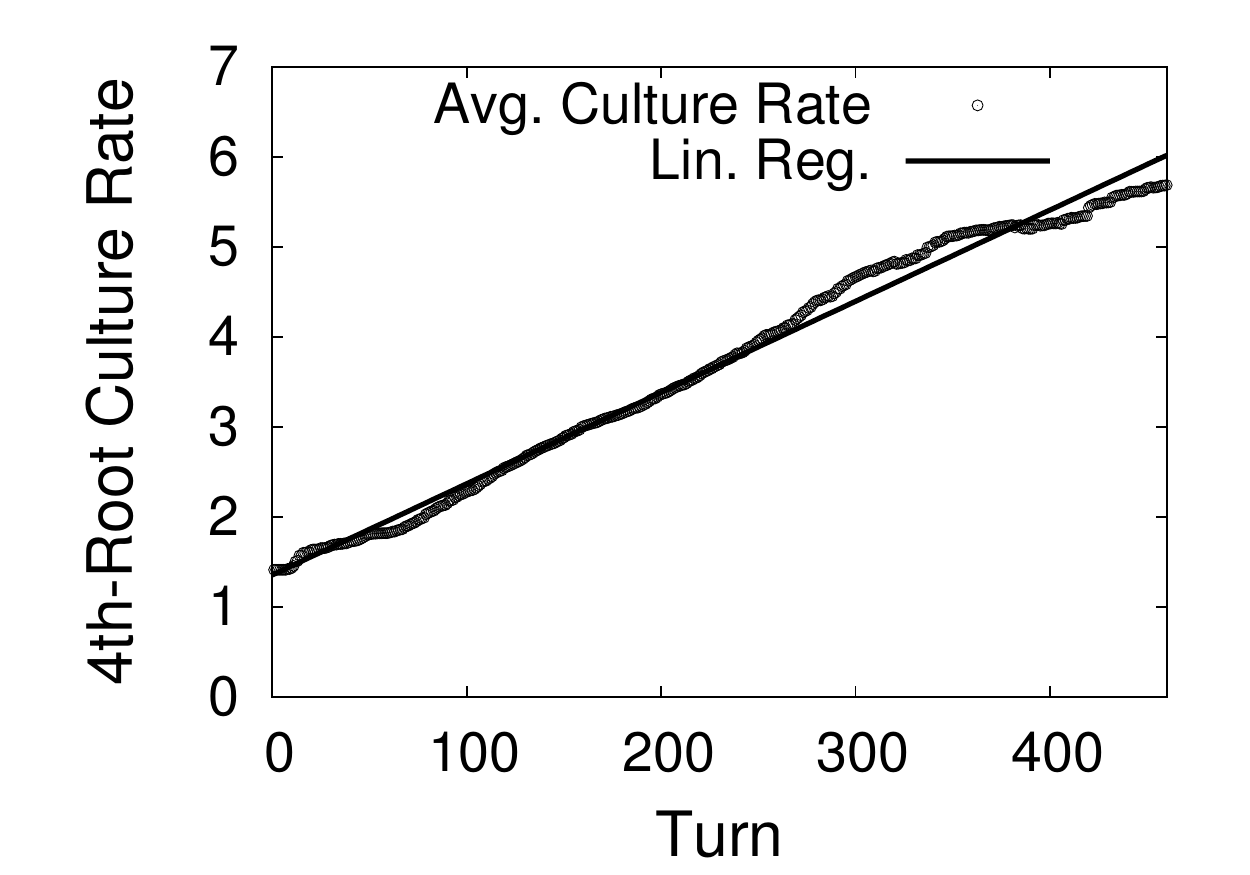}}
  \caption{Linear Regression of the $\sqrt[4]{CultureRate}$ indicator.}
  \label{fig:cultureRate_geral}
\end{figure}

We obtained very high coefficients of determination to the \textit{Culture} (99.86\% to \textit{Alexander} and  99.85\% to \textit{Hatshepsut}) and \textit{CultureRate} indicators (99.11\% and 98.93\% to \textit{Alexander} and \textit{Hatshepsut}, respectively). Besides this, all obtained coefficients are significant with a confidence of 99\%. The graphs with the regressions are presented in Figures~\ref{fig:culture_geral} and~\ref{fig:cultureRate_geral}.

Beyond the regression quality, it is important to note that the coefficients $b_0$ e $b_1$, of \textit{Hatshepsut}, are greater than those of \textit{Alexander}, with a confidence of 99\%. This is what we expected in this situation since \textit{Hatshepsut} has a higher \textit{Culture} preference. This result confirms our hypothesis that some gameplay data are able to distinguish preferences of two different agents, suggesting that they can be used as features when using ML to model players preference.

It is interesting to note that this preference has very few interactions with other game indicators, and maybe \textit{Culture} is the easiest preference to be isolated, since only specific constructions in the game generate culture score. Among the buildings that generate culture are: palaces, educational and religious buildings and wonders. Most of the buildings that generate culture score do not exist at the beginning of the game and they are constructed along the game, explaining why we obtained a polynomial of fifth degree. In fact, we believe that it should be represented by an exponential function, but the limited number of turns does not allow enough growing. 

\subsubsection{\textit{Growth} Preference}

We have analyzed the \textit{Growth} preference observing three different indicators (features): \textit{Cities, Land} and \textit{Plots}. The first one is defined as the ``Number of cities'', the second as ``Amount of land tiles'' and the third as ``Amount of land and water tiles'' \citep{denTeuling_Thesis_10}. Recall that \textit{Alexander}, the chosen agent to represent a high preference, did not have this preference on its higher level (value 5), but on average level (value 2).

The analysis of these three indicators presented a recurrent and expected situation: the existence of two distinct intervals in the dataset. Initially, there is a period in which the growth rate (of \textit{Cities, Land} or \textit{Plots}) is high. This {\em expansionist} period occurs when there are unoccupied lands that are easily dominated. After this initial period, we can observe a {\em maintenance} phase where there is almost a stabilization of these indicators, since all the world has already been ``colonized'' by some agent. The turn number we have chosen as turning point is also presented in Appendix~\ref{app:regression}.

We were able to model mainly the expansionist period, i.e. to obtain functions that fit well in the available data.
All these functions were modeled as line segments, one for each period. 

\begin{figure}[t]
  \centering
  \subfloat[Hatshepsut]{\label{fig:cities_geral_Hatshepsut}\includegraphics[width=0.5\columnwidth]{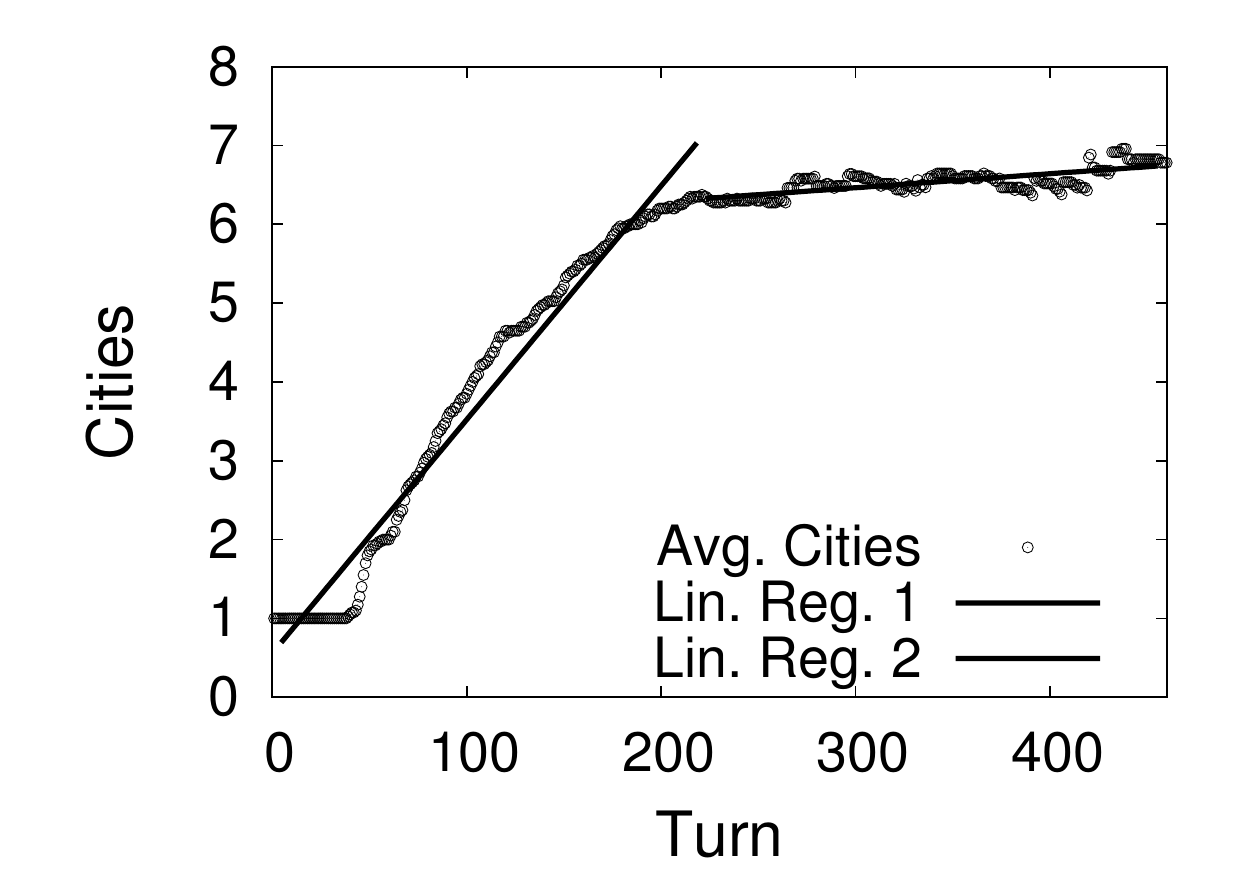}}
  \subfloat[Alexander]{\label{fig:cities_geral_Alexander}\includegraphics[width=0.5\columnwidth]{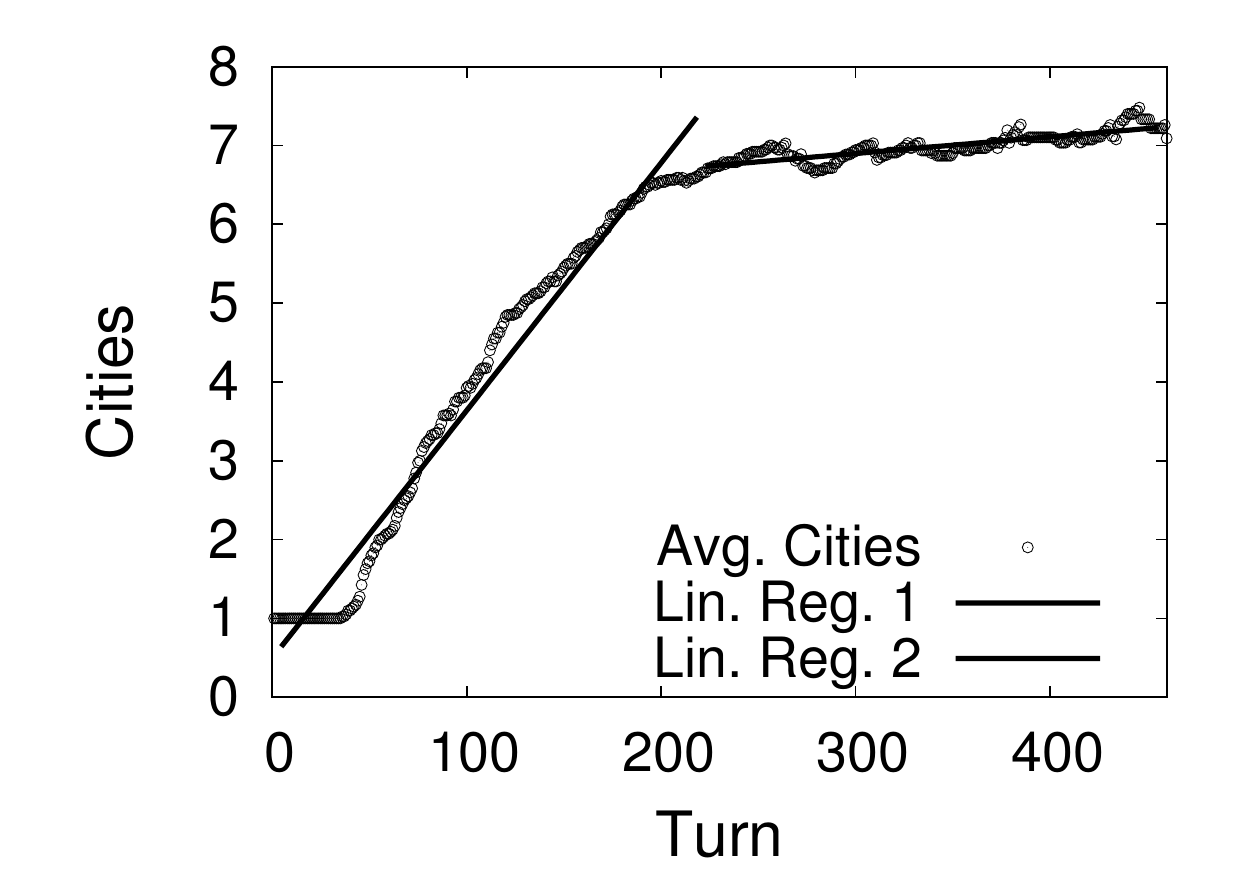}}                
  \caption{Linear Regression of the \textit{Cities} indicator.}
  \label{fig:cities_geral}
  \vspace{-0.4cm}
\end{figure}

The best indicator for \textit{Growth} characterization was \textit{Cities}. For this indicator, we were able to obtain a linear function representing the expansionist period with coefficients different from zero with a confidence of 99\%, and a coefficient of determination equals to 97.17\% to \textit{Alexander} and 96.80\% to
\textit{Hatshepsut}. The second line segment, of the maintenance period, was not so successful in modeling the agents behavior. As in the first line segment, all coefficients are different from zero with a confidence of 99\%, but we were only able to achieve a coefficient of determination equals to 71.39\% to \textit{Alexander} and 56.02\% to \textit{Hatshepsut}. The regressions of these indicators are in Figure~\ref{fig:cities_geral}.

In the expansionist period, we were able to show that the model coefficients are different between agents. As we said, using the model $y = b_0 + b_1x$, the coefficient $b_1$ is larger for \textit{Alexander} with a confidence of 99\%. The equality of $b_0$ is also expected since all agents start with the same number of cities.

We were also able to show that the coefficients in the second line segment of \textit{Alexander} are greater than those of \textit{Hatshepsut}. This result is not so important due to the coefficient of determination of these regressions, but it is still interesting to note that these data corroborate the hypothesis that agents with a higher preference by \textit{Growth} have larger coefficients.

The \textit{Land} indicator allowed us to characterize the expansionist period (coefficient of determination equals to 97.90\% to \textit{Alexander} and 93.76\% to \textit{Hatshepsut}), with a confidence of 95\% that the coefficients are different from zero -- we were able to achieve $b_1 \neq 0$ with a confidence of 99\%, but we were not able to distinguish the coefficients between agents. In the maintenance interval,  the regression did not explain the data nicely, since \textit{Alexander} and \textit{Hatshepsut} coefficients of determination were, in this interval, 23.90\% and 50.04\%, respectively. Even being able to show that the coefficients are different from zero with a confidence of 99\%, there is no sense in evaluating the intersecction between these two agents. Figure~\ref{fig:land_geral} presents this regression and, as the others regressions, its coefficients are presented in Appendix~\ref{app:regression}, in which the confidence intervals are presented with a confidence of 90\% to show that relaxing the confidence interval still does not allow the coefficient separation.

\begin{figure}[t]
  \centering
  \subfloat[Hatshepsut]{\label{fig:land_geral_Hatshepsut}\includegraphics[width=0.5\columnwidth]{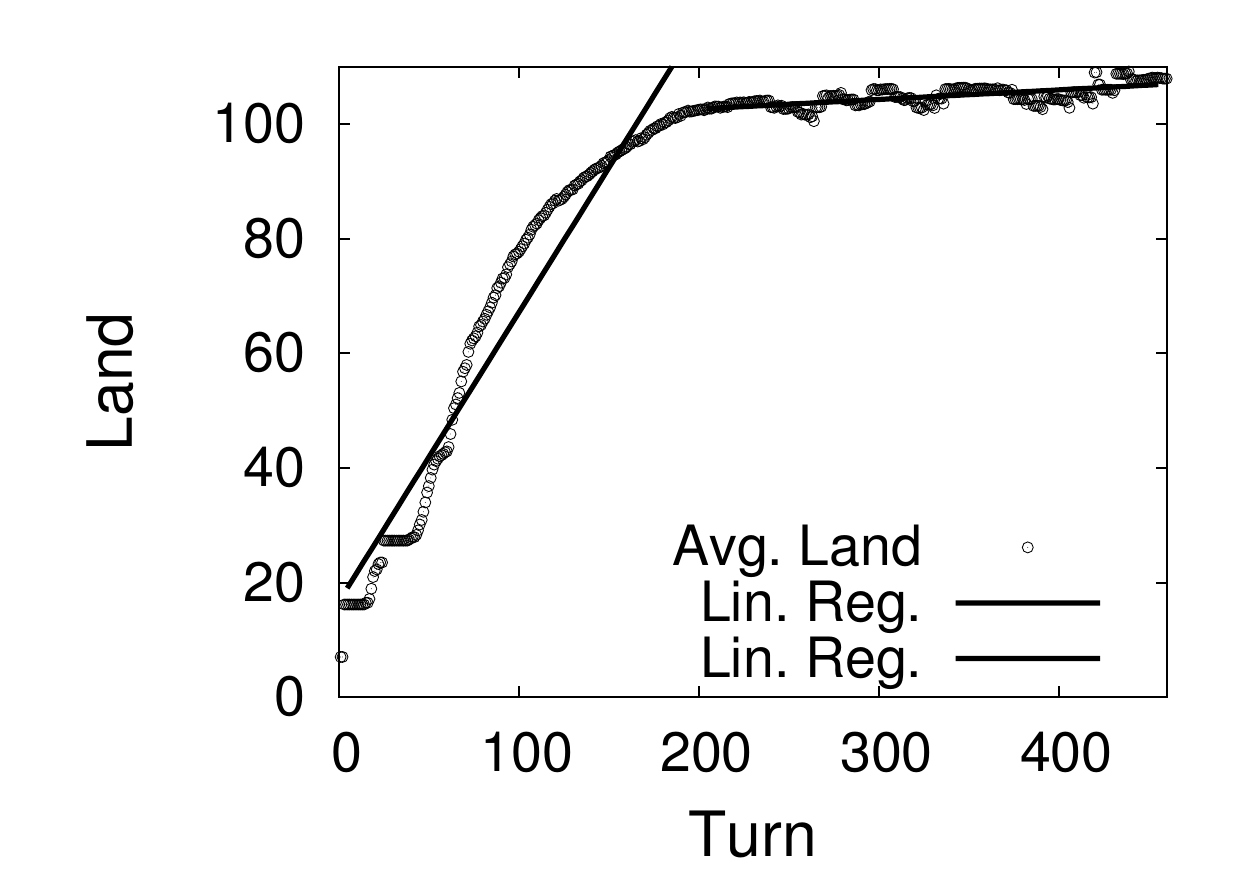}}
  \subfloat[Alexander]{\label{fig:land_geral_Alexander}\includegraphics[width=0.5\columnwidth]{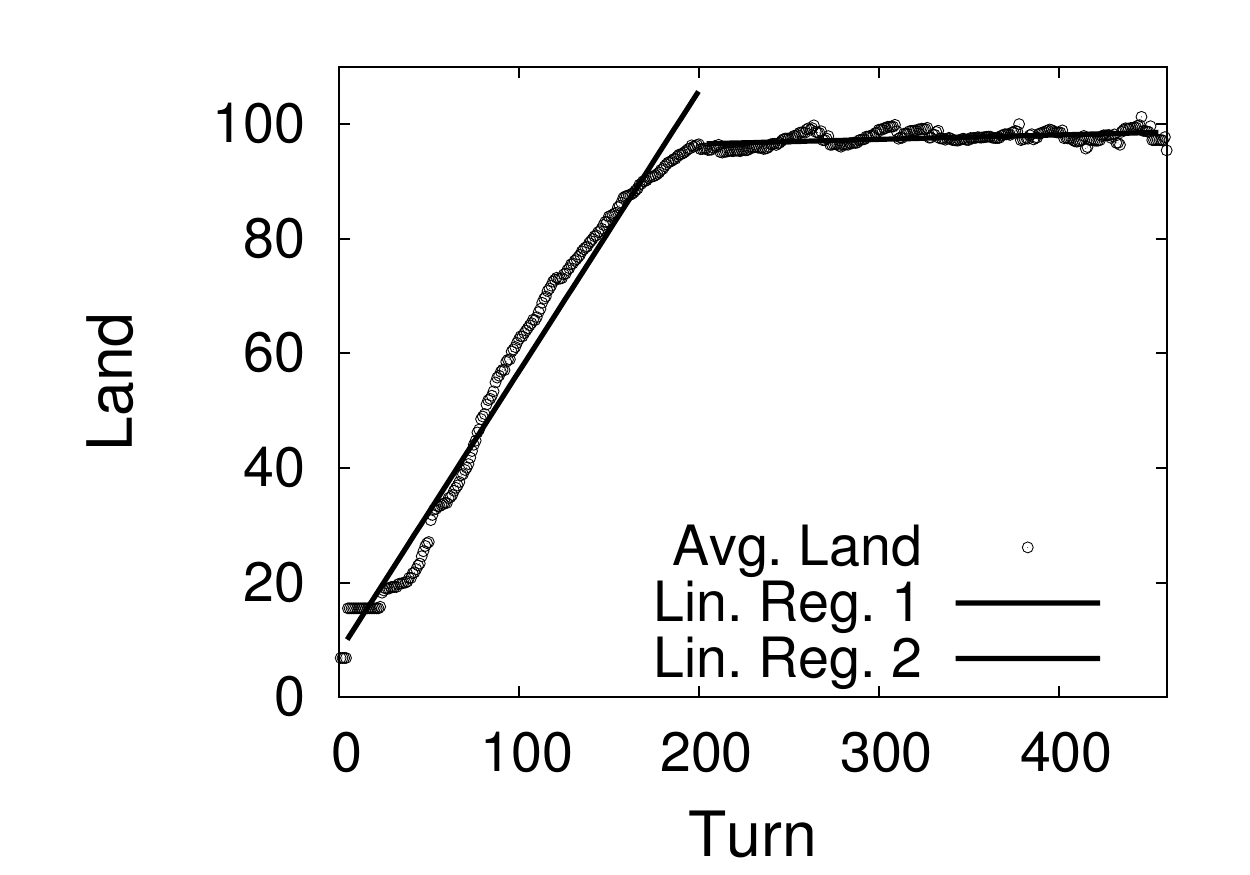}}                
  \caption{Linear Regressions of the \textit{Land} indicator.}
  \label{fig:land_geral}
\end{figure}

There is a reason to the \textit{Land} indicator be descriptive but not discriminative. As the \textit{Cities} indicator, initially there are too much to be conquered and this allow us to precisely describe the expansionist period, but not the maintenance period, since there is a natural unpredictability in the game after its stabilization. This unpredictability is smaller to the \textit{Cities} indicator because it is harder to have a decrease in its value, since it is easier to lose territory's tiles than to lose cities. Due to this, the \textit{Land} indicator has a higher variability. 

This alteration also depends on other preferences like \textit{Culture}, that also makes it harder to be modeled. We believe the inability to discriminate the generated model coefficients are due to the fact that \textit{Land} can also grow with investment in culture, not necessarily just building cities. A player who privileges cities may evolve its territory just like a player that does not, but invest in culture, what raises its cities borders and maybe imply in a high \textit{Land} value. The non-independent coefficient ($b_1$) is the growth rate of the agents borders, this implies that the cities creation generates peaks in some curve points but, in general, this is amortized since we generally have a maximum of 10 cities and 460 turns.

\begin{figure}[t]
  \centering
  \subfloat[Hatshepsut]{\label{fig:plots_geral_Hatshepsut}\includegraphics[width=0.5\columnwidth]{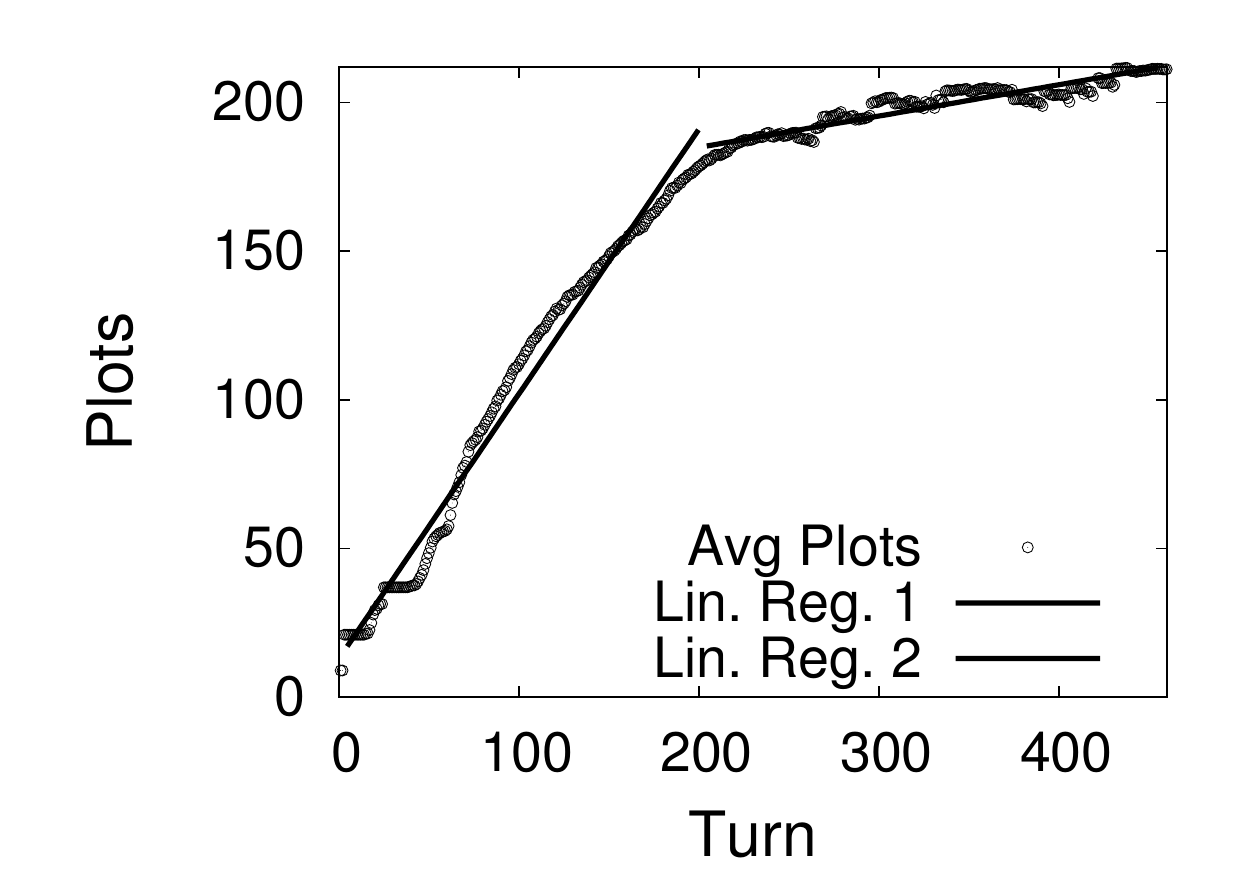}}
  \subfloat[Alexander]{\label{fig:plots_geral_Alexander}\includegraphics[width=0.5\columnwidth]{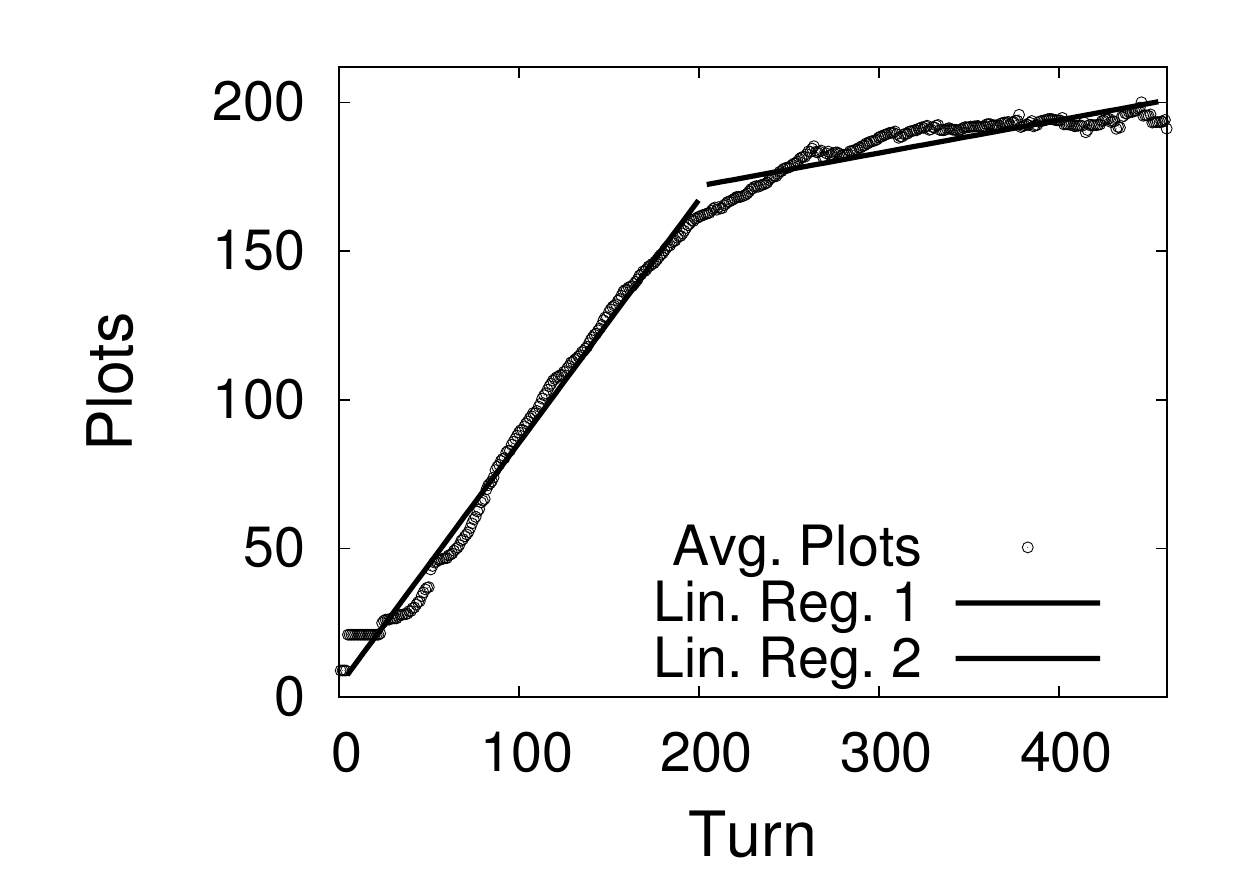}}               
  \caption{Linear Regressions of the \textit{Plots} indicator.}
  \label{fig:plots_geral}
\end{figure}

Finally, the last indicator we evaluated was \textit{Plots}. This indicator is the sum of the number of land and water tiles in the game. As \textit{Land}, we were able to nicely describe the expansionist period but we were not able to discern the two agents (\textit{Alexander} coefficient of determination is 99.15\% and \textit{Hathshepsut} is 95.05\%, with $b_1$ different from zero with a confidence of 99\%). All the discussions previously done are also applicable here. The biggest difference between these two indicators is related to the second phase, the maintenance. We were able to achieve better models than those obtained using the \textit{Land} indicator ($R^2$ equals to 78.73\% to \textit{Alexander} and 88.22\% to \textit{Hatshepsut}, with coefficients different from zero with a confidence of 99\%). The regression of these indicators is in Figure~\ref{fig:plots_geral}. We believe this higher ``stability'' is explained by the water \textit{tiles}, that are harder to be lost by reasons like \textit{Culture}.

In conclusion, as \textit{Plots}, \textit{Land} coefficients overlap. Based on this, the number of cities in each turn is the unique indicator that allows us to discern agents with different preferences, while all indicators successfully describe the agents behaviors. It is interesting to highlight that even not using an agent with a high preference for \textit{Growth}, we were able to distinguish agents with different preferences, showing that even intermediate levels may be useful.

We were able to observe that the characterization/differentiation sometimes is impaired by the interaction of different preferences in the indicator. But these results also gave us hints that the data representing player's state during the game can be used as features by an ML algorithm.

\subsubsection{\textit{Gold} Preference}

The selected indicators for this preference were \textit{Gold} and \textit{GoldRate}. They are respectively: ``Amount of gold'' and ``Amount of gold gained per turn'' \citep{denTeuling_Thesis_10} .

We were able to model the \textit{GoldRate} indicator for the two agents as a straight line. The coefficient of determination of the linear regressions to \textit{Louis XIV} and \textit{Mansa Musa} was, respectively, 98.72\% and 96.14\%. Besides this, the models coefficient $b_1$ of each agent are different from zero with a confidence of 99\% (we were not able to show $b_0$ different from zero, what is not a problem since it represents the initial value). 

\begin{figure}[t]
  \centering
  \subfloat[Louis XIV]{\label{fig:goldRate_geral_LouisXIV}\includegraphics[width=0.5\columnwidth]{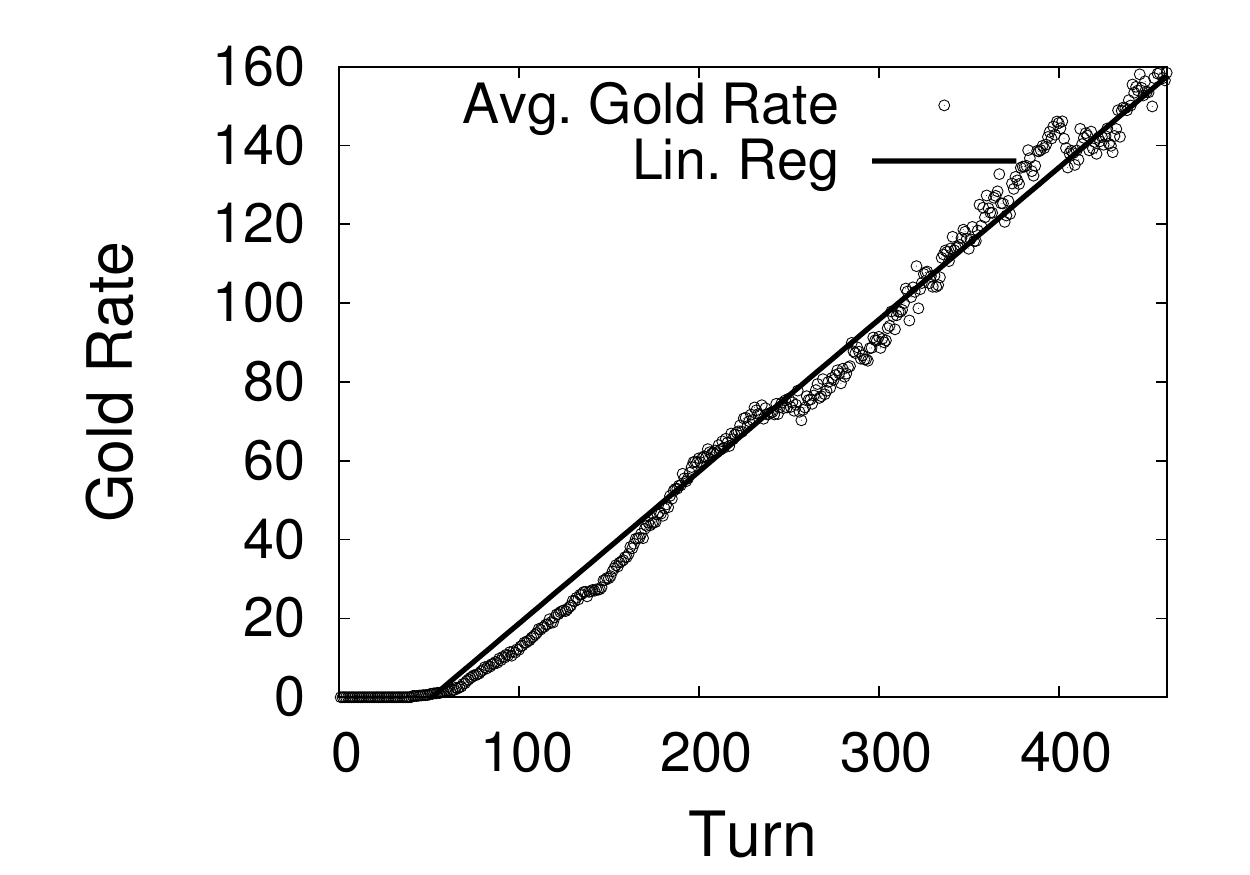}}                
  \subfloat[Mansa Musa]{\label{fig:goldRate_geral_MansaMusa}\includegraphics[width=0.5\columnwidth]{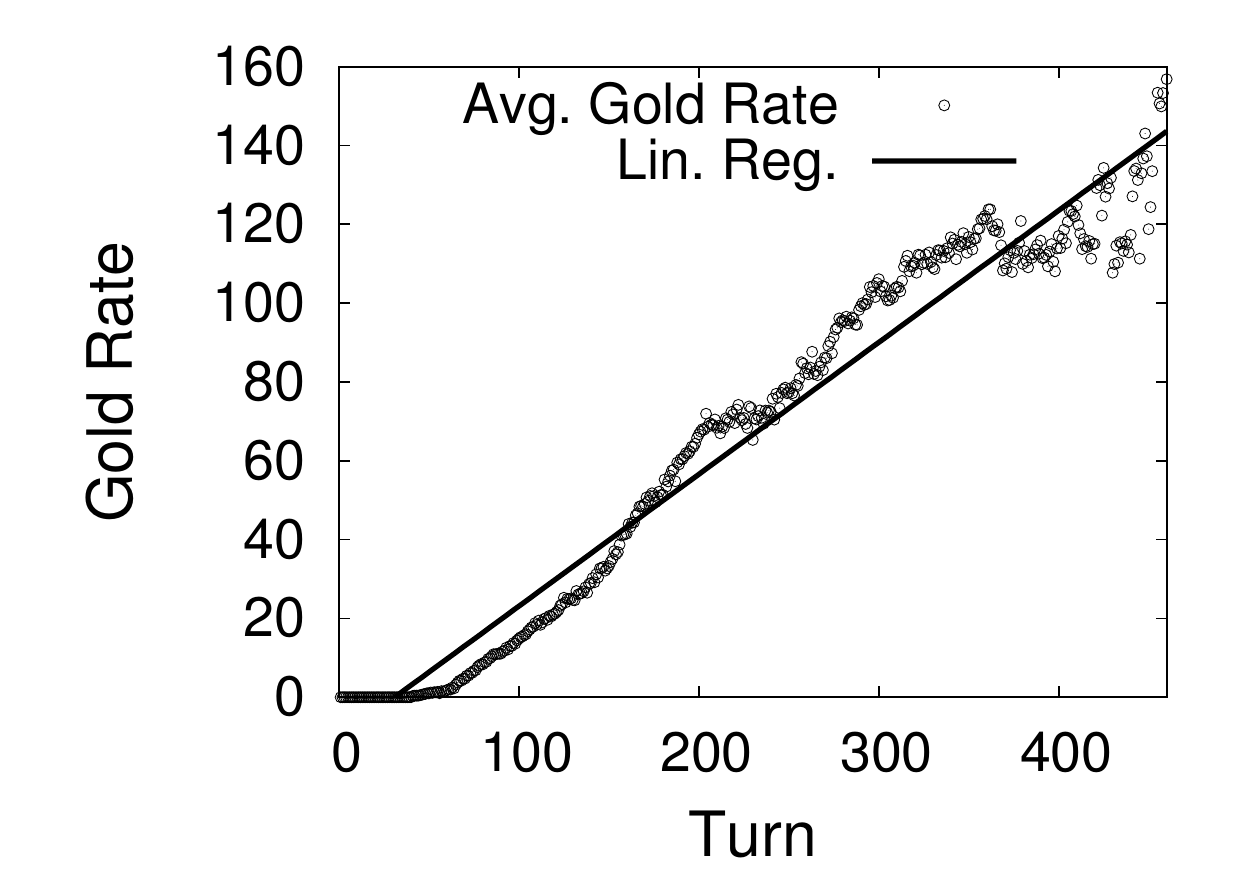}}
  \caption{Linear Regression of the \textit{GoldRate} indicator.}
  \label{fig:goldRate_geral}
\end{figure}

These regressions indicate us that the amount of gold received each turn grows following a linear function but, apparently, the agent preference does not impact in the way it receives gold. This affirmative is valid because even relaxing the confidence of our evaluations we were not able to find intervals that do not overlap. Figure~\ref{fig:goldRate_geral} presents a visual evaluation of the regressions.
 
The gold indicator is not modeled as a polynomial of degree two (integral of the gold rate) because it is not the total gold income, but the income decreased by agent's expenses. Still, apparently, the amount of gold received each turn is similar, independently of the player preference. We then raised a second hypothesis that the amount of gold stored by each agent would be different. 

\begin{figure}[t]
  \centering
  \subfloat[Louis XIV]{\label{fig:gold_geral_LouisXIV}\includegraphics[width=0.5\columnwidth]{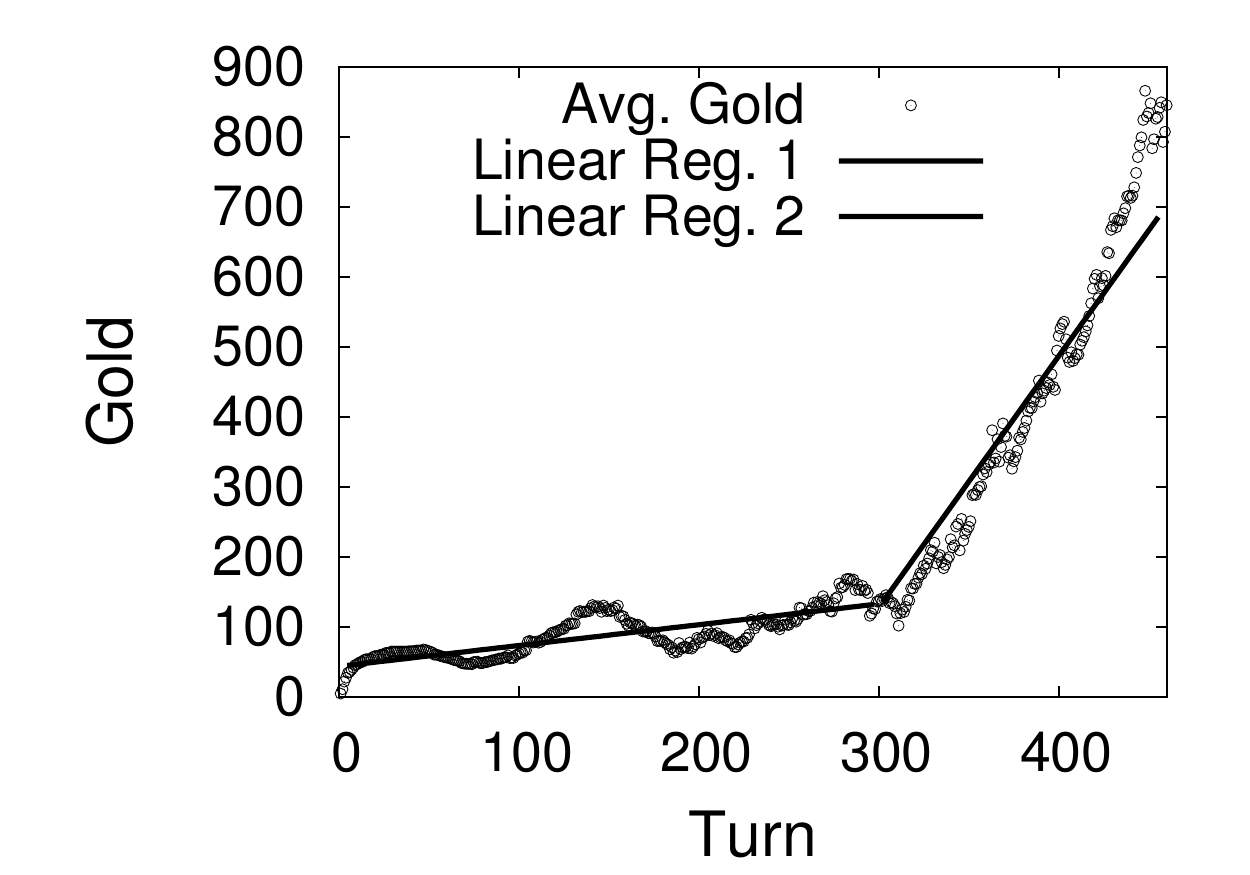}}                
  \subfloat[Mansa Musa]{\label{fig:gold_geral_MansaMusa}\includegraphics[width=0.5\columnwidth]{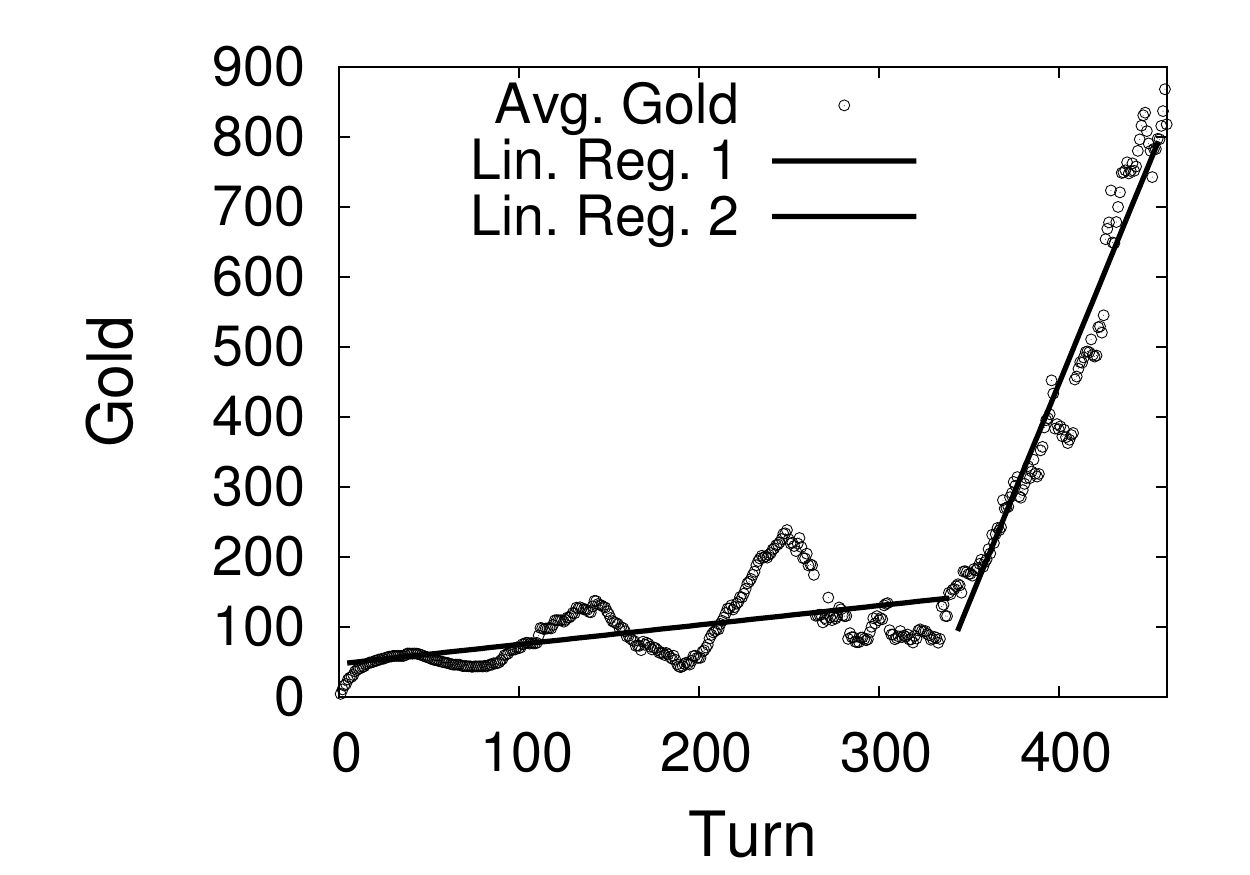}}
  \caption{Linear Regression of the \textit{Gold} indicator.}
  \label{fig:gold_geral}
\end{figure}

The best characterization we achieved for this indicator was using two different line segments. We believe this is due to the gold importance in the game and the several activities that can be done spending it, like donations to other agents, conversion in units upgrades and even receiving it due to the incapacity to continue constructing some buildings, for example. Figure~\ref{fig:gold_geral} exemplifies very well this variable behavior.

Thus, each graph is divided in two line segments. For \textit{Louis XIV}, the first segment goes from turn 1 to turn 300, while for \textit{Mansa Musa}, this first interval goes from 1 to 340. As we can observe in the graph, there is a large variability in this first segment while the second segment is more stable. Despite this analysis, we were not able to assure that the regression coefficients are different from zero (the lowest evaluated level of confidence was 90\%).

Our premise that the indicators \textit{Gold} and \textit{GoldRate} would describe well the agents behavior was partially satisfied since we were able to characterize the \textit{GoldRate} growth but we were not able to do the same for the \textit{Gold} indicator. We believe this difficulty to describe this preference is due to fact that \textit{Gold} is one of the ``most common'' and important resources in the game, permeating several possibilities, what ``degenerates'' the \textit{Gold} evolution during the time.

This last preference being analyzed was harder to be characterized and we were just partially able to model it, since the \textit{GoldRate} describe it but not the \textit{Gold}. The reasons for this are easy to be comprehended after the analysis of the results: creating a city (which we just related to \textit{Growth} preference) implies in a great loss of gold since the cost of new cities is greater than its income. Thus, the variation observed is explained by this. The great ``jump'' after turn 300 can be explained by the ``discovery'' of mercantilism, besides the fact that most of the cities became profitable.

In summary, we were unable to distinguish different agents preferences related to \textit{Gold} and we believe this is because gold is an essential resource in the whole game and the preferences are ``weaker'' when compared to other characteristics not so essential like \textit{Culture}. The reason for this claim is that agents can obtain this resource from different ways and, for a better player experience, a better balancing of this distribution is expected. 

Due to our results and discussions during this section, the impact of preferences interaction to distinguish and characterize agents became very evident with this preference. This may be another explanation to our failure in distinguishing different agents. 

This last result shows that, for some game variables, the evaluation of a unique feature may be insufficient to obtain an agent preference. These results also gave us confidence that a set of features containing gameplay data can be used in order to distinguish player preferences using ML algorithms.

\subsubsection{Victory and Defeat}~\label{sec:victory_defeat}

Prior to continue modeling our problem as an ML approach, it is important to evaluate the influence of the game result in the features' discriminative power. To do this, we separated the data in two disjoint subsets: those originated from matches that were won and those from lost matches. This decision was motivated by the following question: the analysis of all matches as being in the same group, independently of their result, does not generate noises that distort the real agents behavior? Their separation does not make these data more ``stable''?

To answer this question, we revisited every generated model recreating it for the two different subsets, {\em i.e.} each previous model generates two others, using data of victories and defeats. The intuition is that the game result would have impact in the indicators. We present this analysis below.

Regarding culture, another analysis over this preference is useful just to validate the results previously obtained, since they were extremely satisfactorily. Again, we modeled \textit{Culture} as a polynomial of degree five and \textit{CultureRate} as a polynomial of degree four, for both won and lost matches. As in the previous modeling, our regressions were very good for both sets ($R^2$ greater than 98\% to all indicators, for both agents) and all obtained coefficients are not zero with a confidence of 99\%.

As obtained in the general analysis, the \textit{Hatshepsut} coefficients were greater than those from \textit{Alexander}, who has no interest for culture, while \textit{Hatshepsut} has.

To model \textit{Growth}, we followed the previous methodology, dividing all indicators in two periods: expansionist and maintenance. In this preference, we were able to observe benefits of the separation between matches that were won or lost. The benefit was a decrease in the data variability and a better understanding of the problem. We were not able to obtain a more distinguishable model for the different agents.

The first evaluated indicator was \textit{Cities}. As in the general evaluation, the expansionist period was easily characterized to \textit{Hatshepsut} and \textit{Alexander} for victory ($R^2$ equals to 98.01\% and 97.91\%, respectively) and for defeat ($R^2$ equals to 96.98\% and 96.58\%, respectively), with a confidence of 99\% that the coefficients are not zero. It is interesting to note that, in the subset of matches won, the coefficients overlap, while they do not in the matches lost. The non-independent coefficient of \textit{Alexander} was greater than the one of \textit{Hatshepsut} with a confidence of 95\%.

Besides this, when \textit{Hatshepsut} presents a ``more'' expansionist behavior, it increases its victory chances. We believe the statistical difference is achieved only in lost matches because even under adverse situations, \textit{Alexander} still aims to expand his borders while \textit{Hatshepsut} does not.

In the stabilization period, we were able to better characterize the models of won matches due to the lower data variability ($R^2$ equals to 95.11\% to \textit{Alexander} and 81.79\% to \textit{Hatshepsut}). This result corroborates our hypothesis that the game result may influence some indicators, since we observed a higher variability in the lost matches, probably due to the different types of victory: an agent can lose a game by score (without a single military conflict), or may have its lands devastated by the enemy. The coefficients of determination to \textit{Alexander} and \textit{Hatshepsut} in this situation were, respectively, 79.84\% and 39.59\%. This difference between them is probably explained by the military preference of the first, not evaluated here. He is probably better able to defend his lands, even when he loses the game.

\begin{figure}[t]
  \centering
  \subfloat[Alexander (Defeat)] {\label{fig:cities_derrota_Alexander}\includegraphics[width=0.5\columnwidth]{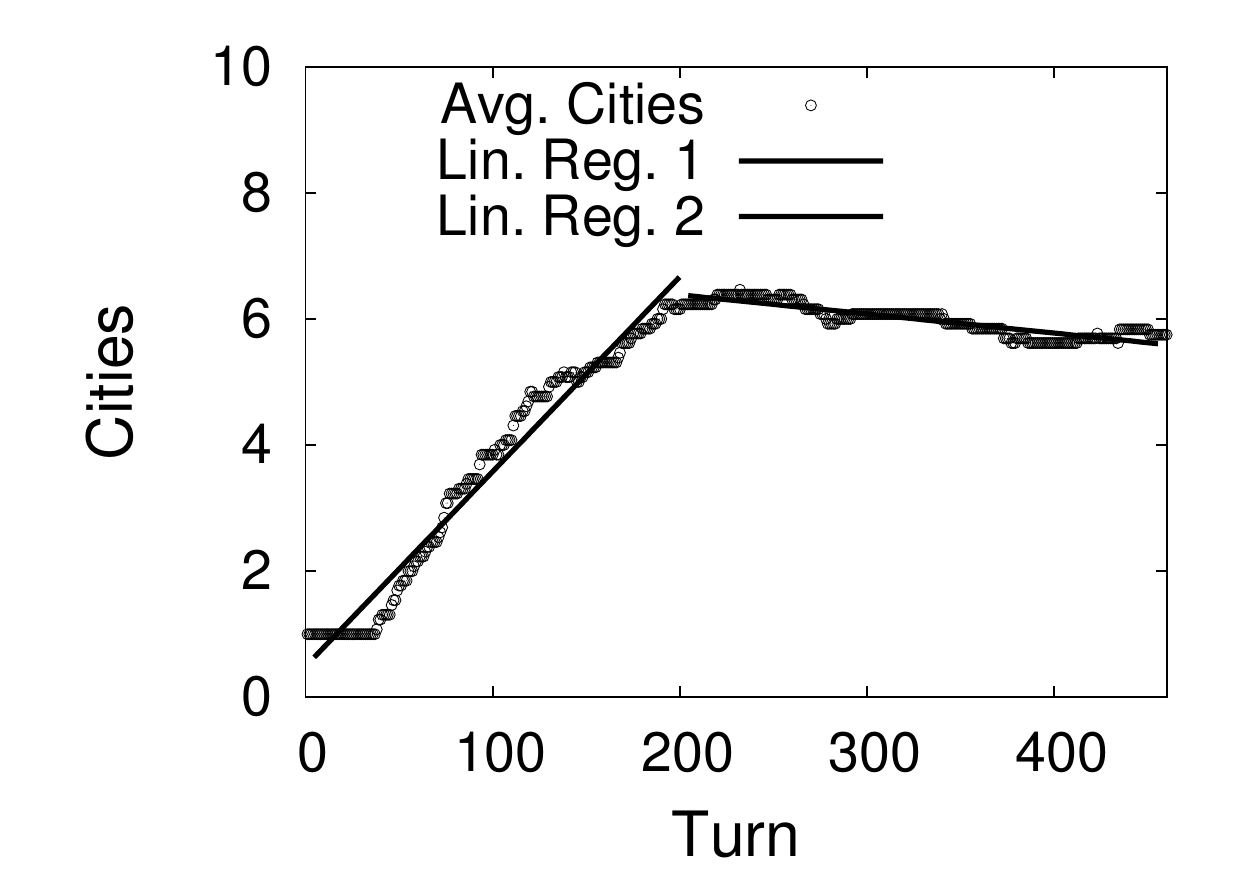}}
  \subfloat[Hatshepsut (Defeat)] {\label{fig:cities_derrota_Hatshepsut}\includegraphics[width=0.5\columnwidth]{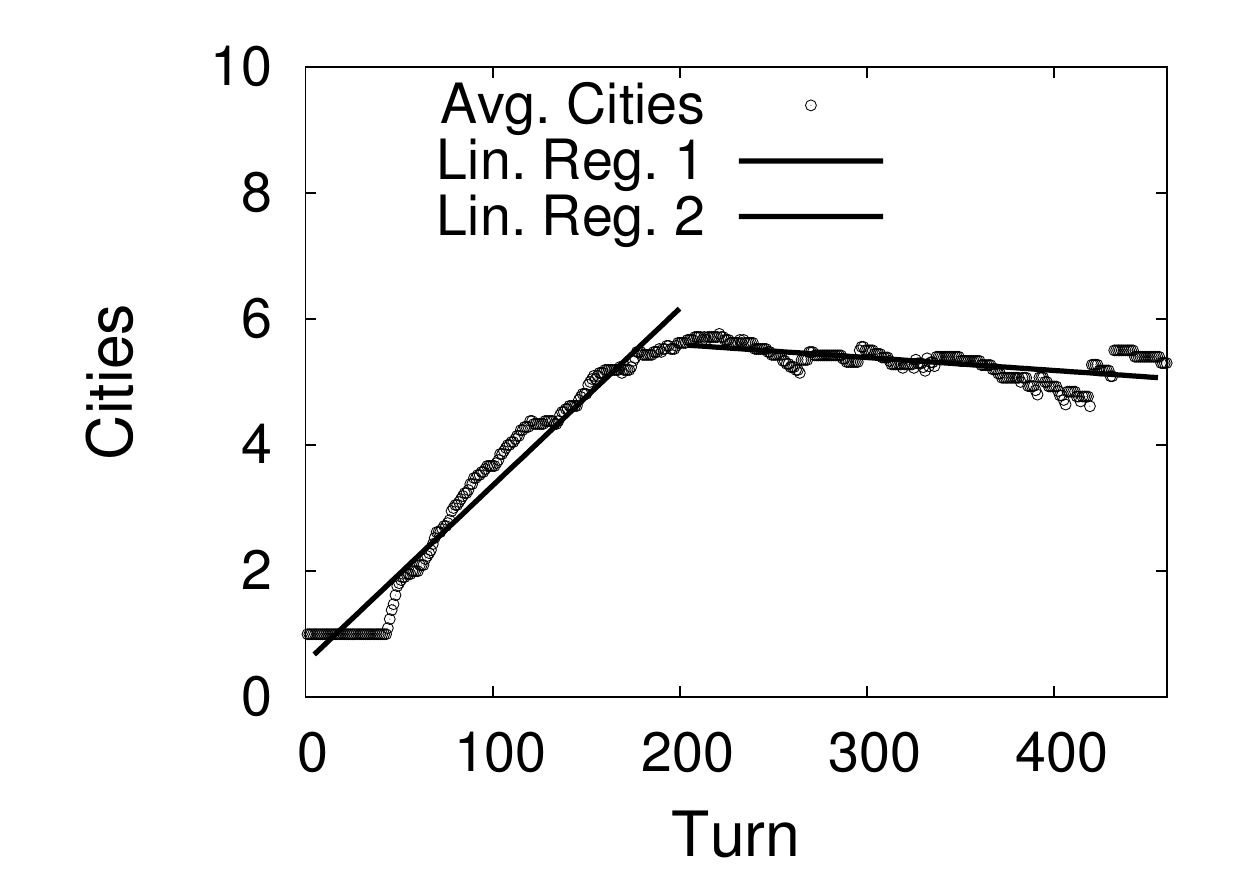}}                
  \caption{Linear Regressions for the \textit{Growth} preference in the lost matches subset.}
  \label{fig:derrota_cities}
\end{figure}

Figure~\ref{fig:derrota_cities} presents the number of cities in the lost matches. There is an interesting result here: we have observed that, for the first time, the non-independent coefficients ($b_1$) were lower than zero, i.e. when an agent loses a game, its territory decreases at the ending.

Once we finished this analysis, by the first time we were able to note the variability generated by the game result, precisely in the number of cities. The observed variability is large because in some matches, after a specific turn, the cities may continue increasing or start decreasing.

The analysis of the \textit{Land} indicator for victory and defeat was very similar to the general analysis. We were able to characterize well the expansionist period but we were not able to distinguish the coefficients between agents, not even for matches won or lost.

As in the general analysis, the second interval presents a much higher variability. In the set of won matches, the coefficient of determination of \textit{Alexander} and \textit{Hatshepsut} was, respectively, 93.65\% and 53.95\%, with all coefficients different from zero with a confidence of 99\%. This difference has already been discussed.

We were able to show in this regression that the growth rate for \textit{Alexander} is greater than \textit{Hatshepsut}'s growth rate, also with a confidence of 99\%. The results for lost matches were extremely variable and any discussion comparing them is meaningless. Despite the variability decrease in the won matches subset, we were not able to obtain additional useful information with this separation.

The last indicator related to this preference that needs to be revisited is \textit{Plots}. The division keeps the excellent characterization of the expansionist period in the victory and defeat subsets, but we were still unable to distinguish different agents. The data variability decreased after the division but no additional discussion or analysis is possible.

\begin{figure}[t]
  \centering
  \subfloat[Mansa Musa (Defeat)] {\label{fig:goldRate_derrota_MansaMusa}\includegraphics[width=0.5\columnwidth]{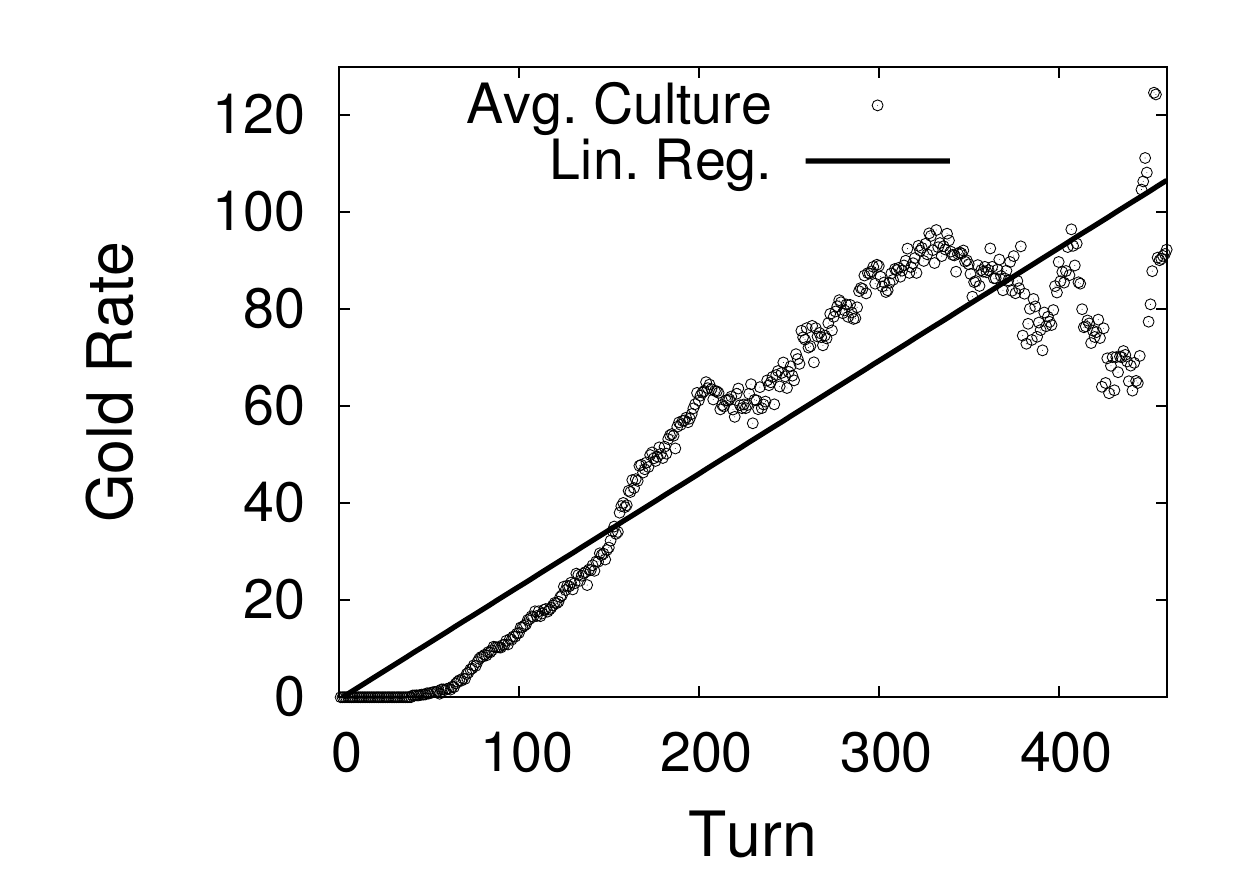}}
  \subfloat[Mansa Musa (Defeat)] {\label{fig:gold_derrota_MansaMusa}\includegraphics[width=0.5\columnwidth]{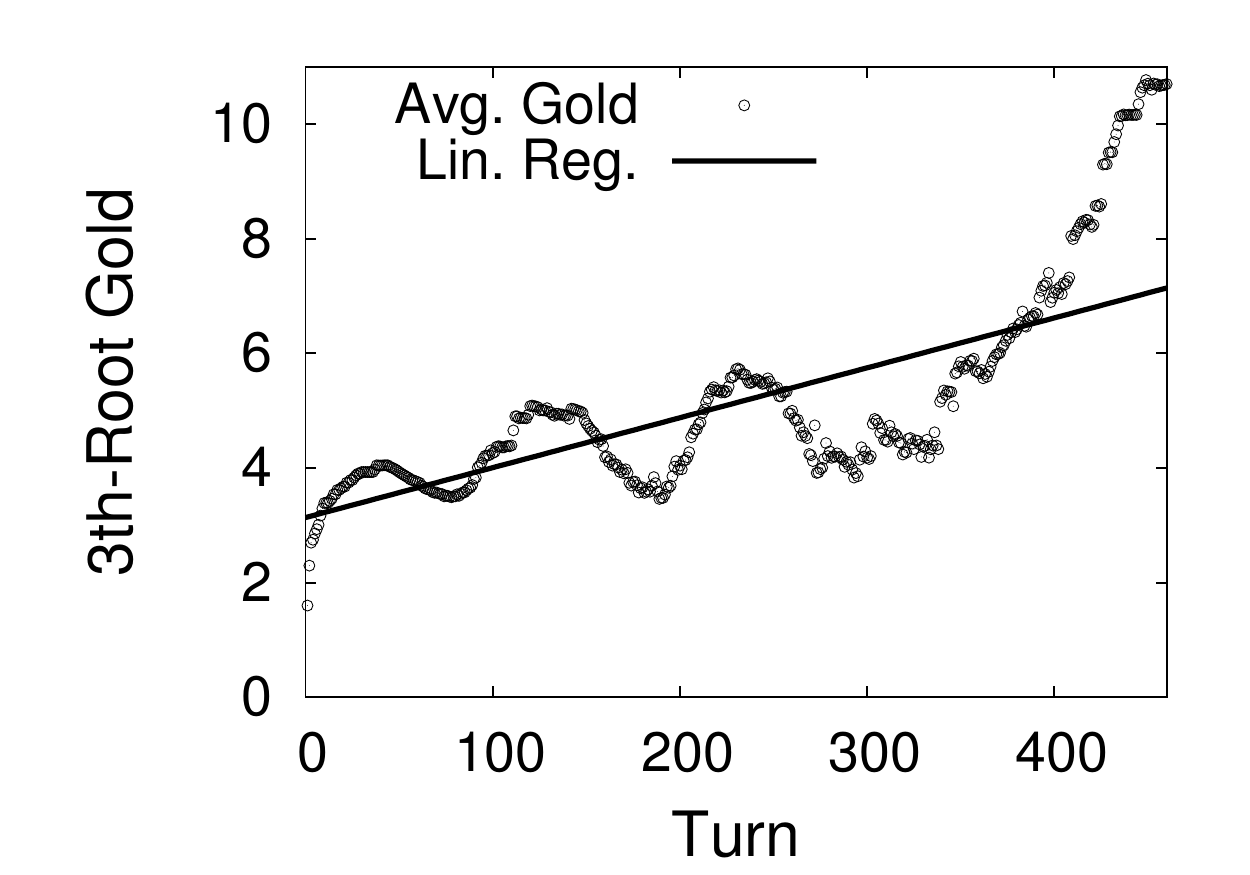}}                
  \caption{Linear Regressions for the \textit{Gold} preference in the lost matches subset.}
  \label{fig:derrota_gold}
\end{figure}

Finally, regarding \textit{Gold}, firstly analyzing the victory subset, we were able to obtain very good regressions to the \textit{GoldRate} indicator, with an $R^2$ equals to 98.48\% for \textit{Louis XIV}, with all coefficients different from zero with a confidence of 99\%. We also achieved an $R^2$ of 97.03\% for \textit{Mansa Musa}, with a confidence of 95\% that all coefficients are different from zero. Similarly to general characterization, the coefficients of these regressions overlap. The same happened in the lost matches subset. We were able to obtain coefficients different from zero with a confidence of 99\% and regressions with $R^2$ equals to 97.98\% to \textit{Alexander} and 84.60\% to \textit{Mansa Musa}. Figure~\ref{fig:goldRate_derrota_MansaMusa} presents the regression for this last case.

The analysis of the defeat situations shows us that the higher variability observed in the general data of \textit{Mansa Musa} is explained by the matches lost. Probably this occurs because he loses cities at the end of the game and this implies in a smaller gold rate.

This analysis, despite helping us to understand the higher variability in this agent's indicators, did not allow us to distinguish both, and we believe the reasons are the same previously discussed.

We modeled all gold scores as single lines. \textit{Mansa Musa}'s gold indicator in matches lost could have been characterized as two distinct lines, but we decided to keep its regression similar to the others of this indicator, as it can be seen in Figure~\ref{fig:gold_derrota_MansaMusa}.

As in all other graphs, we can observe a high variability of the indicator values and, as previously discussed, probably due to the cities feature. In the matches won subset, we observed that \textit{Louis XIV}'s non-independent coefficient ($b_1$) is greater than \textit{Mansa Musa}'s coefficient, that has preference for \textit{Gold}. This relation is inverted on the subset of matches lost. Is interesting to note that these results seem counter intuitive since we would expect a higher coefficient in all cases. It does not happen because \textit{Louis~XIV} has a high preference for \textit{Culture} and this influences its amount of gold since the \textit{Culture} generates a territorial expansion that implies in a higher number of resources.

In summary, at the end of all these analyses we achieve two different conclusions: first of all, for most of the analyzed data, the separation between matches won and lost generally implied in getting a better fit for half of the models and a worse fit for the other half. Despite being able to better characterize the separated models, we are not able to distinguish agents preferences that had not been already distinguished by the general data.

Whether the division is beneficial is not a trivial decision. It depends on the chosen ML technique, since the fusion between victory and defeat may make implicit the average performance of the agent: if it tends to lose more than win, with the time we would be able to better characterize its losing performance since natural ``weights'' would raise automatically with the results accumulation. This would be more difficult considering separated matches.

\section{Players Representation}\label{sec:experiments_representation}

Regarding the generic representation proposed in previous chapter, in this section we instantiate it to the game \textsc{Civilization IV}. As already discussed, a good model must be able to generate different behaviors by varying the representation of a virtual agent, as well as to be ``inferable'', i.e. by observing behavior one must be able to represent it. We discuss these two topics in sequence. 

In order to evaluate whether it is possible to infer agents' models, we perform an experiment in which we manually infer some	of the weights that could model virtual agents, comparing these weights to those in their predefined models. This shows how different behaviors can be explained and expressed by different models.

A second experiment to be performed would be the evaluation of whether the weights variation would generate distinct behaviors. However, this task would be extremely complicated to be performed in the game \textsc{Civilization IV}. \cite{Machado_SBGames_12} further discuss this topic using the game \textsc{Counter Strike} as test-bed. We show how small changes in the model can generate different behaviors in the game.  Here, we decided to not discuss this topic further, to keep the focus on the game \textsc{Civilization IV}.

\subsection{Civilization IV: from behaviors to models}

As we are interested int inferring models by observing players, we used the game \textsc{Civilization IV} to perform our tests, since it allows data to be easily extracted and explicitly presents agents characteristics in {\em XML} files, allowing us to convert it into our representation.

As previously stated in Chapter~\ref{background}, \textsc{Civilization IV} provides {\em XML} files with several game characteristics, such as agents preferences, buildings and units info. This interface allows us to observe and modify the game characteristics. Each agent presents several attributes in the XML defining them. This is quite useful since we are able to correctly generate relevant variables for the model. Six agents preferences are defined in the game: {\em Culture, Gold, Growth, Military, Religion} and {\em Science}. Each of these preferences serve as multipliers to agents decisions and cost of specific actions.

Since we do not intend to present the best possible model for the game, we decided to use the preferences described and valued in the {\em XML} to generate an agent representation:

$$
Pm = \langle C_u, G_o, G_r, M_i, R_e, S_c\rangle
$$
where $C_u$ is preference for culture, $G_o$ for gold, $G_r$ for growth, $M_i$ for military, $R_e$ for religion and $S_c$ for science.

Our goal is to infer different models from different behaviors, and verify if this representation is coherent with observed behaviors. Hence, just like in Section~\ref{sec:characterization}, we selected two different agents: one with a high preference for a specific feature and other without it. Our goal was to analyze game indicators that would, theoretically, be affected by this preference. We expect higher weights for the agent with a higher preference.

For example, agent \textit{Alexander} has a strong \textit{Military} preference while agent \textit{Hatshepsut} prefers \textit{Culture}. These agents may be represented by us using the information available on the \textit{XML} files. Extracting the respective weights from the configuration files, we obtain the following models ($Pm_A$ for \textit{Alexander} and $Pm_H$ for \textit{Hatshepsut}): 
$$Pm_A = \langle 0, 0, 2, 5, 0, 0\rangle$$
$$Pm_H = \langle 5, 0, 0, 0, 2, 0\rangle$$
As we can see, the set of variables different from zero of each agent is a disjoint set and it is expected to result in distinct behaviors during the game, assuming that the above representation is useful.

In fact, in Section~\ref{sec:characterization}, we already showed differences between different agents. We will discuss here some of those results applied to this evaluation. We used the same dataset used to characterize agents with linear regressions.

\begin{figure}[t]
\begin{minipage}[b]{0.45\linewidth}
\centering
\includegraphics[width=\textwidth]{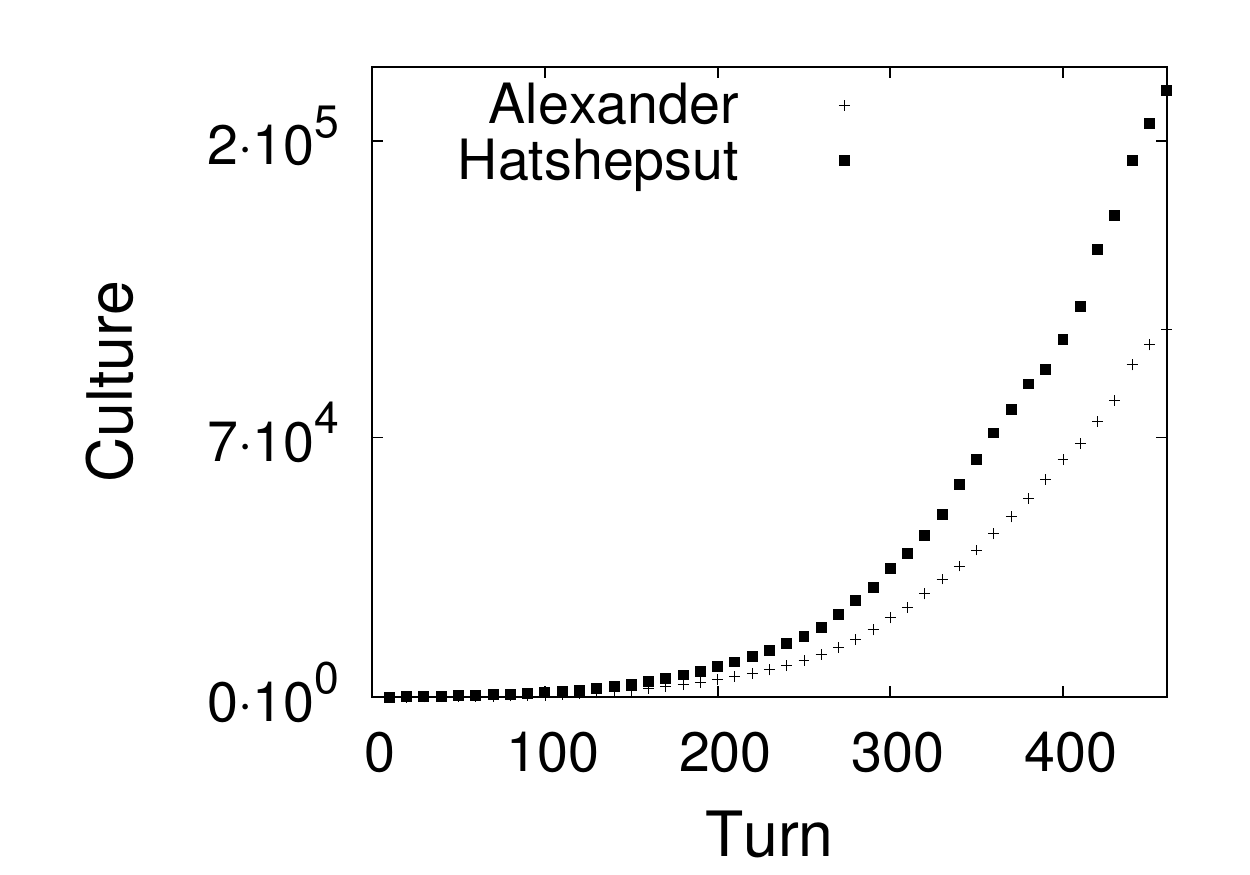}
\caption{Comparison of Culture between different agents.}
\label{fig:culture}
\end{minipage}
\hspace{0.5cm}
\begin{minipage}[b]{0.45\linewidth}
\centering
\includegraphics[width=\textwidth]{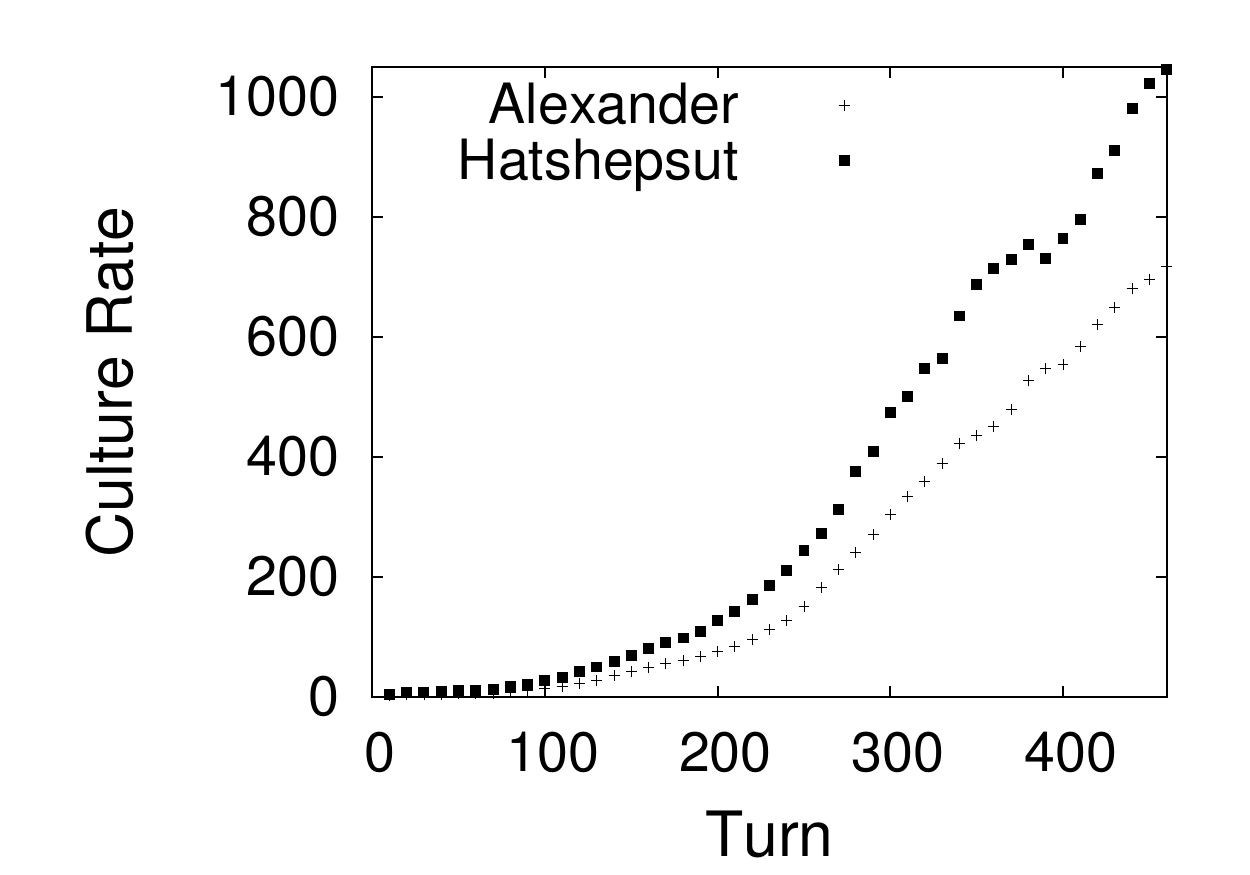}
\caption{Comparison of Culture Rate between different agents.}
\label{fig:cultureRate}
\end{minipage}
\end{figure}

The plots for \textit{Alexander} and \textit{Hatshepsut} overall culture score along all matches is presented in Figure~\ref{fig:culture}. An analogous plot, but for \textit{Culture Rate}, is presented in Figure~\ref{fig:cultureRate}. Note that differently from previous section, we did not apply any transformation to the data.

As expected, based on the agents' models, \textit{Hatshepsut}'s curves represent a superior bound for \textit{Alexander}'s curves. This is coherent with \textit{Hatshepsut}'s higher preference for \textit{Culture}.

Performing the linear regressions previously discussed, we obtained the following straight line equations for \textit{Alexander} ($y_A$) and \textit{Hatshepsut} ($y_H$), for \textit{Culture}. Between parenthesis we present the coefficient of determination.
$$y_A = 0.0183x + 1.7772 \ (99.86\%)$$ \vspace{-0.6cm}
$$y_H = 0.0194x + 2.1366 \ (99.85\%)$$ 

Using the same pattern, the line equations for the \textit{Culture Rate} indicator were:
$$y_A = 0.0096x + 1.0939 \ (99.93\%)$$ \vspace{-0.6cm}
$$y_H = 0.0101x + 1.3567 \ (99.11\%)$$

Finally, we were able to show, using \textit{paired t-tests}, that both coefficients are statistically different with a confidence of 99\%.

After this characterization we can clearly see that \textit{Alexander} indicators are bounded below those from \textit{Hatshepsut}. Then we can infer that ${C_u}_A < {C_u}_H$. Simplifying the assignment, assuming that we have only two different values, 0 and 5, we can say that $Pm_A = \langle 0, ... \rangle$ and $Pm_H = \langle 5, ... \rangle$, which corresponds to the agents original model.

Our objective here was to show that it is possible to generate models in the discussed representation from data collected during play. We performed the same operations for other two randomly selected preferences (\textit{Growth} and \textit{Gold}) in previous section, and its analyses are also applicable here. 

\section{Overview}

In this chapter, we have instantiated the three first steps of the generic approach proposed to model the problem of preference prediction. We first distinguished virtual agent's preferences using gameplay data, showing that these different behaviors can be observed along the time and manually inferred for some preferences, what assisted us defining the features to be used by the ML algorithm we define (the gameplay indicators). We also concluded that the impact of the division between matches won and lost is not so useful, assisting us to understand what are relevant examples, as defined in Chapter~\ref{generic_approach}. Then we presented an evaluation regarding the generic representation, showing its feasibility in the game \textsc{Civilization IV}. 

The next chapter discusses the use of ML to automatically define player's preferences, first discussing the last steps of the instantiation of the approach discussed in the previous chapter: the selection of the ML task, the appropriate algorithm and its best parameters configuration.

%% file: experimental_results.tex
\chapter{Experimental Results}\label{experimental_results}

\epigraphhead[240]{\scriptsize It doesn't matter how beautiful your theory is, it doesn't matter
                               how smart you are - if it doesn't agree with the experiment, it's wrong.\\
									 \vspace{-0.6cm}
									 \begin{flushright}
									 Richard Feynman
									 \end{flushright}
								   }
\vspace{1.6cm}

This chapter presents our experimental results regarding the prediction of \textsc{Civilization IV} player's preferences. Our approach, which follows a supervised learning framework, is detailed together with the used algorithms and the search method for their best parameters configuration.

Three different experiments were performed: we first evaluated the accuracy of the selected algorithms in the \textit{Traditional Dataset}, composed of known virtual agents. After that, we used the generated models to predict the preferences of virtual agents in the \textit{Alternative Dataset}, composed of virtual agents different from those used in the training phase. Finally, after evaluating the generality of the obtained models in virtual agents, these models are used to predict self-reported human players' preferences.

\section{Experimental Methodology}

Before presenting our experimental methodology, we shall define how we are going to approach the preference modeling problem, the fourth step in the methodology proposed in Chapter~\ref{generic_approach}. We decided to model it as a supervised learning task, once we know the preferences of each artificial player. Additionally, it is easy to ask human players to self-classify themselves. In this approach, the algorithm learns a model by finding relations between a set of features that describe the examples (matches) and a class (preference). For the preference learning problem, each example represents a player turn and the features are the game score indicators, as discussed in the previous chapter.

We decided to model the class as the presence or absence of a given preference (binary classification), allowing us to make a more precise inference. This decision was based on the assumption presented in Section~\ref{sec:generic_problem_modeling}, where we state that human players may face difficulties when mapping their preferences into different levels. Additionally, as we show in Table~\ref{tb:classDistribOriginalDatasetComplete}, some preferences do not have more than two values in the original dataset. This approach represents one of the main differences between our work and the one presented in \citep{denTeuling_Thesis_10, Spronck_AIIDE_10}, which tackled player modeling as a multi-class classification problem. As showed by experimental results, our approach achieves better accuracies. 

Regarding the classification algorithms used, the fifth step listed in Chapter~\ref{generic_approach}, our first idea was to use the SVM to make our results directly comparable with those reported in \citep{denTeuling_Thesis_10, Spronck_AIIDE_10}. We used a Radial Basis Function (RBF) kernel. Note that, as mentioned, the problem of choosing the best classifier for a learning problem  is an open issue in the machine learning literature, tackled by the sub-area of meta-learning \citep{Bradzil_Book_09}. 

Besides \textit{SVM}, we also evaluated three different classifiers based in different paradigms, namely \textit{NaiveBayes, AdaBoost} and \textit{JRip}, aiming to compare the performance of these different approaches since, as far as we know, no other work had done this properly in the past. Each classifier has different assumptions about the input data, and generate different models that may explore distinct data characteristics, allowing us to evaluate different approaches for tackling preference modeling.

The first experiment we performed was the application of the four classification algorithms in the \textit{Traditional Dataset}, using a 10-fold cross validation. Hence we used this first result to select the best algorithms to evaluate how the models generated by them generalize. In a second experiment, we used the whole \textit{Traditional Dataset} as training set, and generated models that classified six different artificial players that were never seen by the classifiers (\textit{Alternative Dataset}), then we used the same models to predict human players preferences.

\subsection{K-Fold Cross Validation}

A cross-validation is a method traditionally used in the ML literature to give experiments statistical confidence. It divides the dataset in $k$ different folds (in our case, k = 10), where one fold is separated for test, and the other $k-1$ are used during the training process. This approach generates $k$ different training and test sets, each test set being different from the other nine. The final results reported correspond to the average error over the $k$ folds, in order to guarantee that the results were not found by chance according to the characteristics of the learning data. When this process is stratified, the folds preserve the class distribution of the original dataset. 

\subsection{Parameters Optimization and Data Sampling}

Once we included the \textit{SVM} algorithm as one of the classifiers, we went to parameter tuning, the last step in the approach proposed in Chapter~\ref{generic_approach}. We sought for the best possible parameters only for \textit{SVM} because it is known to be extremely sensitive to parameters variations. The best parameters configuration for each data fold was obtained using a tool called \textit{easy}, present in the \textit{libSVM} implementation \citep{Chang_ACMTrans_11}.

In practice, we optimized two different parameters: \textit{cost~(c)} and \textit{gamma~(g)}. Their meanings were already presented in Chapter~\ref{background}. Algorithm~\ref{alg2} presents the grid search performed by \emph{easy} and its maximum and minimum values (defined in the tool). This algorithm is applied to each fold of the cross-validation. We are looking for $best\_c$ and $best\_g$. The step values used are also those defined in the tool \textit{easy}.

\begin{algorithm}
\caption{Grid Search}
\label{alg2}
\begin{algorithmic}

\ENSURE Best parameters $c$ and $g$

\vspace{0.1cm}

\STATE $c \leftarrow 0$, $best\_c \leftarrow 0$
\STATE $g\leftarrow 0$, $best\_g \leftarrow 0$
\STATE $best\_accuracy\leftarrow 0$

\FOR{$c=-5$ to 15 with step 2}
  \FOR{$g=3$ to -15 with step -2}
    \STATE \COMMENT{Evaluate accuracy with these parameters}
    \IF{Evaluate($2^c, 2^g$) $> best\_accuracy$}
      \STATE $best\_c \leftarrow c$
      \STATE $best\_g \leftarrow g$
    \ENDIF
  \ENDFOR
\ENDFOR

\vspace{0.1cm}

\RETURN {$\{best\_c, best\_g\}$}

\end{algorithmic}
\end{algorithm}

The \textit{Evaluate} function is responsible for running the classifier (training and testing) with the parameters passed. The accuracy is calculated as the number of instances correctly classified divided by the number of instances in the dataset.

As observed in the algorithm, this search is computationally expensive, requiring 6,600\footnote{11~from \textit{c} search $\times$ 10~from \textit{g} search $\times$ 10~from 10-fold $\times$ 6~from number of preferences.} different experiments
with \textit{SVM}. It was not possible to run these experiments in a feasible time. For this reason, a sampling method was used to reduce the 72,653 vectors available for training, speeding up the learning process. To ease the comparison between the learning algorithms, we used the same sampled dataset to evaluate all algorithms.

Since our dataset contained data from complete matches and their evolution, we sampled the data considering not its vectors,
but its matches. In other words, we added or removed complete sets of vectors, with each set representing a whole match. It is important to stress that we sampled just the training set, not the test set. In other words, we first generate a set
containing 1/10 of the dataset (due to the 10-fold cross validation). This is the test set. After this step we sample the remaining matches to create the training set. We originally had 240 matches and sampled 25\% of their original size, obtaining a test set of 24 matches and a training set with 54 matches\footnote{25\% of 216 matches (the result of
the total set minus the test set).}. 

All results reported were obtained by the best parameter configuration chosen by \emph{easy} using the sampling process described above. Note that the sampling process is done with data different from those used for testing the models, and used by all learning algorithms. 

Tables~\ref{tb:classDistribOriginalDatasetComplete} and~\ref{tb:classDistribNewDataset}
present the number of examples belonging to each class in the test set of the original dataset
(Table~\ref{tb:classDistribOriginalDatasetComplete}) and the sampled dataset (Table~\ref{tb:classDistribNewDataset}). The sampling algorithm is showed in Algorithm~\ref{alg1}. In the new dataset, we mapped the instances with class $0$ as those with class $-1$, and those with class $2$ and $5$ as $1$.
\begin{center}
\begin{table}[H]
\caption{Number of samples in each class for the test sets of the original data in the \textit{Traditional} and \textit{Alternative Dataset} as in \citep{denTeuling_Thesis_10} (problem modeled with 3 classes).}
\label{tb:classDistribOriginalDatasetComplete}
{\footnotesize
\hfill{}
\begin{tabular}{l | c c c | c c c}
    \noalign{\hrule height 1.2pt}
    \multirow{2}{*}{Preferences} &\multicolumn{3}{c|}{Known Agents (\textit{Traditional})} &\multicolumn{3}{c}{Unknown Agents (\textit{Alternative})}\\
    \cline{2-4} \cline{5-7}
     &0  &2   &5 &0  &2   &5 \\
    \hline
    Culture       & 11,828 & -     & 6,111  & 16,330     & -      & -     \\
    Gold          & 11,937 & 3,023 & 2,979  & 11,412     & 2,242  & 2,676 \\
    Growth        & 14,756 & 3,183 & -      & 4,981      & 11,349 & -     \\
    Military      & 6,337  & 5,396 & 6,206  & 5,454      & -      &10,876 \\
    Religion      & 11,602 & 6,337 & -      & 13,951     & 2,739  & -     \\
    Science       & 15,296 & -     & 2,643  & 13,552     & -      &2,778  \\
    \hline
\end{tabular}}
\hfill{}
\vspace{0.1cm}
\end{table}
\end{center}

\begin{center}
\begin{table}[H]
\caption{Number of samples in each class of the test set for the sampled \textit{Traditional} and \textit{Alternative Datasets}. Standard deviation showed between parenthesis.}
\label{tb:classDistribNewDataset}
{\footnotesize
\hfill{}
\begin{tabular}{l | c c | c c}
    \noalign{\hrule height 1.2pt}
    \multirow{2}{*}{Preferences} &\multicolumn{2}{c|}{Avg. of Known Agents} &\multicolumn{2}{c}{Unknown Agents}\\
    \cline{2-3} \cline{4-5}
     &-1  &1   &-1  &1\\
    \hline
    Culture       & 4,914.7 (158.2) & 2,420.2 (161.6)  &16,330 & -      \\
    Gold          & 5,037.7 (311.5) & 2,464.4 (201.8)  &11,412 & 4,918  \\
    Growth        & 5,821.9 (359.3) & 1,304.2 (135.0)  &4,981  & 11,349 \\
    Military      & 2,472.8 (129.5) & 4,853.2 (574.7)  &5,454  & 10,876 \\
    Religion      & 4,714.6 (241.5) & 2,576.2 (322.6)  &13,951 & 2,739  \\
    Science       & 5,788.5 (601.3) & 1,262.0 (82.6)   &13,552 & 2,778  \\
    \hline
\end{tabular}}
\hfill{}
\vspace{0.1cm}
\end{table}
\end{center}

\renewcommand{\algorithmicrequire}{\textbf{Input:}}
\renewcommand{\algorithmicensure}{\textbf{Output:}}

\begin{algorithm}
\caption{Sample Dataset}
\label{alg1}
\begin{algorithmic}
\REQUIRE  Sample percentage $perc$ \COMMENT{$0 \leq perc \leq 1$}\\
\hspace{0.63cm} Array $matches$ containing all matches \COMMENT{Each match contains all its turns}
\ENSURE Array $sampledMatches$ with sampled matches

\vspace{0.1cm}
\STATE $sampledMatches \leftarrow \emptyset$
\STATE $matchesWithPref \leftarrow \emptyset$
\STATE $matchesWithoutPref \leftarrow \emptyset$
\FOR{$i=0$ to $matches.size$}
  \STATE \COMMENT{Check if the agent of that match has the preference}
  \IF{$matches[i].preference$ = \TRUE}
    \STATE $matchesWithPref.add(matches[i])$
  \ELSE
    \STATE $matchesWithoutPref.add(matches[i])$
  \ENDIF
\ENDFOR
\vspace{0.1cm}
\STATE $shuffle(matchesWithPref)$
\STATE $shuffle(matchesWithoutPref)$
\vspace{0.1cm}
\FOR{$i=0$ to $matchesWithPref.size \times perc $}
\STATE $sampledMatches.add(matchesWithPref[i])$
\ENDFOR
\vspace{0.1cm}
\FOR{$i=0$ to $matchesWithoutPref.size \times perc $}
\STATE $sampledMatches.add(matchesWithoutPref[i])$
\ENDFOR
\vspace{0.1cm}
\RETURN $sampledMatches$

\end{algorithmic}
\end{algorithm}

\section{Classification of Virtual Agents Preferences}

The experimental phase with artificial agents (not human players) was divided in two steps. As already mentioned, we first predicted preferences of virtual agents from the \textit{Traditional Dataset}, using a standard 10-fold cross-validation procedure. Then, in a second experiment, we used the entire \textit{Traditional Dataset} to generate a model that classified the agents in the \textit{Alternative Dataset}. This second experiment was designed to evaluate the generalization capabilities of the models when predicting preferences of unknown agents, since generalization is the most important characteristic to enable the model to be used in games in real situations.

The first experiment used 130 features, including two available at the end of each match, called \textit{match result} and \textit{peace}. This allowed us to perform an \textit{off-line review} (as discussed in Chapter~\ref{related_work}). Since we do not have these two features in the \textit{Alternative Set}, with unknown agents, we retrained our dataset removing them, using 128 features to simulate an \textit{on-line tracking} in our second experiment. We divided the results presentation in these two sections.

\subsection{Off-line Review of Known Agents}

We compare the four methods using binary classification (\textit{Binary-Class SVM}, \textit{Naive Bayes}, \textit{AdaBoost} and \textit{JRip}) with the Majority Class and the results reported by \cite{Spronck_AIIDE_10}, which we named \textit{Multi-Class SMO}. As we previously discussed, this work can be considered our baseline, since the authors tackled the same problem of preference modeling in the game \textsc{Civilization IV}. The experimental results are reported in Table~\ref{tb:smoResultMinus100}.

The \textit{Majority Class} corresponds to the percentage of the most frequent class of the dataset. For example, for the \textit{Culture} preference, 67.0\% of the turns were played by agents with no preference for \textit{Culture}. This means that if the classifier learned nothing and generated a model classifying every turn as ``without preference'', it would obtain the reported accuracy. The column \textit{Multi-Class SMO} presents the results reported in~\citep{denTeuling_Thesis_10, Spronck_AIIDE_10}. They modeled the problem as a multi-class classification, {\em i.e.}, instead of modeling preferences as existent or not, they modeled three levels of preference. This result is shown just for a high-level comparison, since the baseline did not performed cross-validation on its experiments, making a more detailed comparison meaningless. 

We tried to reproduce the results reported in \citep{denTeuling_Thesis_10, Spronck_AIIDE_10} but we obtained a different number than those reported, hence we were not able to run a cross-validation in the baseline's approach to compare theirs with our approach, in this first experiment. Thus, in Table~\ref{tb:smoImprovementResultMinus100}, we use the \textit{Majority Class} as baseline, only presenting the improvement of our approach over it. We understand that it is not an ideal baseline and, we only use it to evaluate whether the applied algorithms are learning ``something''.

First analyzing the results generated by the four algorithms we ran using the binary classification approach, we can see that the unique algorithm that performed better than a possible \textit{Majority Class} approach for all preferences was \textit{JRip}. Surprisingly, \textit{SVM} did not present a good performance when compared to \textit{AdaBoost} and \textit{JRip}, even demanding considerably more time to be trained with good parameters.

It is interesting to observe that the accuracies corroborate most of the discussion presented in Chapter~\ref{player_modeling_civ4}, since the worse improvements occurred for the \textit{Gold} and \textit{Growth} preferences. The only improvement obtained for these preferences that was greater than zero was 0.7\%, using \textit{JRip}. These results may be explained by the importance of gold  in the game (hence all players  somehow prioritize it) and the absence of the first 100 turns in the dataset. As we have shown in the last chapter, the first turns are very important to characterize \textit{Growth}. We kept the dataset without the first 100 turns to be coherent with \citep{denTeuling_Thesis_10, Spronck_AIIDE_10}. A future study should further evaluate the impact of removing these first 100 turns, using the complete dataset in experiments similar to those we performed.

Let us now analyze \textit{Military} and \textit{Religion}, which we were able to correctly classify all instances (accuracy of 100\% using \textit{SVM, AdaBoost} and \textit{JRip}, generating an improvement greater than 50\%). These happened because two features available in the end of the game, \textit{match result} and \textit{peace}, are able to differ \textit{Military} and \textit{Religion} preferences by themselves. They are only available when performing off-line review. Since we do not have this information in the second dataset, we retrained our model without them to classify the unknown agents. These results will be presented in Section~\ref{sec:online_agents}.


Finally, we compared the accuracies obtained by different classifiers. We performed paired t-tests with 95\% of confidence to present them. The unique preference in which \textit{SVM} presents a higher accuracy, when compared to the other methods, is \textit{Culture} (78.9\% while the second higher is 69.2\%). For this preference, we may obtain the following ordering: \textit{Naive Bayes} $<$ \textit{AdaBoost} $<$ \textit{JRip} $<$ \textit{SVM}, with 95\% of confidence. 

For all other preferences, \textit{JRip} and \textit{AdaBoost} presented the higher accuracies. \textit{SVM} presented accuracies similar to the higher ones for some preferences, while similar to \textit{Naive Bayes} in others. We decided not to use \textit{SVM} in the next experiment, since its results vary a lot from one preference to another, its running time is costly and, most importantly, its results are statistically inferior to those obtained by \textit{JRip} and \textit{AdaBoost} for most preferences.

\begin{landscape}

\begin{center}
\begin{table*}[th]
\centering 
\caption{Accuracy of our methods (Binary-SMO, Naive Bayes, AdaBoost and JRip) contrasted with the most frequent class (Majority Class) and with \cite{Spronck_AIIDE_10}'s approach (Multi-Class SMO). The results are the average accuracy of 10-folds. The Root Mean Squared Error (RMSE) is shown in parenthesis.}
\label{tb:smoResultMinus100}

\hfill{}
\begin{tabular}{ l >{\centering\arraybackslash}m{2.5cm}  >{\centering\arraybackslash}m{2.5cm}  >{\centering\arraybackslash}m{2.5cm}  >{\centering\arraybackslash}m{2.5cm} >{\centering\arraybackslash}m{2.5cm} >{\centering\arraybackslash}m{2.5cm}}
    \noalign{\hrule height 1.2pt}
    Preference &Majority Class &Multi-Class SMO &Binary-Class SVM &Naive Bayes &AdaBoost &JRip\\
    \hline
    Culture       &67.0\%    & 78.9\% (0.46)    & 73.1\% (3.08)    & 67.1\% (2.99)     &66.5\%  (5.20)     &69.2\%  (4.66)\\
    Gold          &67.2\%    & 74.6\% (0.38)    & 63.9\% (7.41)    & 62.6\% (3.04)     &64.0\%  (5.51)     &67.6\%  (5.14)\\
    Growth        &81.7\%    & 83.5\% (0.41)    & 78.0\% (4.27)    & 76.1\% (3.55)     &83.1\%  (3.52)     &82.3\%  (2.79)\\
    Military      &66.1\%    & 61.0\% (0.43)    & 100.0\% (0.00)    & 84.4\% (2.58)     &100.0\% (0.00)     &100.0\% (0.00)\\
    Religion      &64.8\%    & 79.0\% (0.46)    & 100.0\% (0.00)    & 84.9\% (2.93)     &100.0\% (0.00)     &100.0\% (0.00)\\
    Science       &82.0\%    & 88.4\% (0.34)    & 81.0\% (7.24)    & 79.1\% (1.14)     &83.4\%  (3.07)     &84.8\%  (4.02)\\
    \hline
\end{tabular}
\hfill{}
\vspace{0.1cm}
\end{table*}
\end{center}

\begin{center}
\begin{table*}[th]
\vspace{-0.2cm}
\centering 
\caption{Improvement of each approach over Majority Class. Since our approach has a different  number of classes, it is not fair to evaluate our improvement over \cite{Spronck_AIIDE_10}'s approach. The improvement is computed as the difference between the accuracy and the baseline divided by the baseline.}

\label{tb:smoImprovementResultMinus100}

\hfill{}
\begin{tabular}{ l >{\centering\arraybackslash}m{2.5cm}  >{\centering\arraybackslash}m{2.5cm}  >{\centering\arraybackslash}m{2.5cm} >{\centering\arraybackslash}m{2.5cm}}
    \noalign{\hrule height 1.2pt}
    Preference &Binary-Class SMO &Naive Bayes &AdaBoost &JRip\\
    \hline
    Culture    & 9.0\% \ \ \textcolor[rgb]{00,0.45,0.10}{$\blacktriangle$}    & 0.1\% \ \  \textcolor[rgb]{00,0.45,0.10}{$\blacktriangle$}      & -0.7\% \ \textcolor[rgb]{0.7,00,00}{$\blacktriangledown$}    &3.3\% \ \ \textcolor[rgb]{00,0.45,0.10}{$\blacktriangle$} \\
    Gold       & -4.9\% \ \textcolor[rgb]{0.7,00,00}{$\blacktriangledown$}    & -6.8\% \  \textcolor[rgb]{0.7,00,00}{$\blacktriangledown$}     & -4.8\% \ \textcolor[rgb]{0.7,00,00}{$\blacktriangledown$}     &0.7\% \ \ \textcolor[rgb]{00,0.45,0.10}{$\blacktriangle$} \\
    Growth     & -4.5\% \ \textcolor[rgb]{0.7,00,00}{$\blacktriangledown$}    & -6.9\% \ \textcolor[rgb]{0.7,00,00}{$\blacktriangledown$}     & 1.7\% \ \ \textcolor[rgb]{00,0.45,0.10}{$\blacktriangle$}    &0.7\% \ \ \textcolor[rgb]{00,0.45,0.10}{$\blacktriangle$}\\
    Military   & 51.3\% \textcolor[rgb]{00,0.45,0.10}{$\blacktriangle$}    & 27.7\% \textcolor[rgb]{00,0.45,0.10}{$\blacktriangle$}      & 51.3\% \textcolor[rgb]{00,0.45,0.10}{$\blacktriangle$}     &51.3\%  \textcolor[rgb]{00,0.45,0.10}{$\blacktriangle$}\\
    Religion   & 54.3\% \textcolor[rgb]{00,0.45,0.10}{$\blacktriangle$}    & 31.0\% \textcolor[rgb]{00,0.45,0.10}{$\blacktriangle$}     & 54.3\% \textcolor[rgb]{00,0.45,0.10}{$\blacktriangle$}    &54.3\% \textcolor[rgb]{00,0.45,0.10}{$\blacktriangle$}\\
    Science    & -1.2\% \ \textcolor[rgb]{0.7,00,00}{$\blacktriangledown$}    & -3.5\% \  \textcolor[rgb]{0.7,00,00}{$\blacktriangledown$}     & 1.7\% \ \ \textcolor[rgb]{00,0.45,0.10}{$\blacktriangle$}    &3.4\% \ \ \textcolor[rgb]{00,0.45,0.10}{$\blacktriangle$}\\
    \hline
\end{tabular}
\hfill{}
\vspace{0.1cm}
\end{table*}
\end{center}

\end{landscape}

To conclude this first discussion, it is interesting to stress some observed behaviors. A first conclusion that can be drawn is that \textit{SVM} seems to not be the best approach for preference modeling, despite the preference given to it by researchers in the area, such as \citep{Spronck_AIIDE_10, Machado_CIG_12}. Regarding \textit{Naive Bayes}, we were able to see that in three preferences it is as good as \textit{SVM}, despite running instantly while \textit{SVM} may take days to classify agents. Additionally, \textit{Naive Bayes} presented the lowest variation between classifications (no RMSE higher than 4.0), an indicative that it may be a more stable method to this problem. \textit{JRip} may seem an interesting approach, because it does not only present the best performance overall, with all accuracies higher than 65\%, but it also gives us rules that allow game designers to understand and ``debug'' a specific classification. These rules may be even intuitive, such as a learned rule that says that: ``If you have preference for \textit{Gold} then you will not be at war''.

\subsection{Online Tracking of Unknown Agents}\label{sec:online_agents}

To conclude our experiments regarding virtual agents, we evaluated how general the learned models are. In this case, we used all instances of the first dataset as a training set, and the \textit{Alternative Dataset} as test set, composed of agents that were not used to generate the matches in the training set. Recall the class distribution of all preferences is presented in Table~\ref{tb:classDistribOriginalDatasetComplete}. As previously explained, due to the low \textit{SVM} performance and its high computational cost, we did not use it in this second experiment. The results are presented in Table~\ref{tb:smoAltResult} and the improvement over the baseline in Table~\ref{tb:improvementAltResult}.

\begin{center}
\begin{table*}[th]
\caption{Accuracy of the evaluated approaches (Naive Bayes, AdaBoost and JRip) contrasted with the most frequent class (Majority Class) and with \cite{Spronck_AIIDE_10}'s approach (Multi-Class SMO) for unknown agents, not seen in the training process. No error is shown since we just executed this classification once.}

\hfill{}
\begin{tabular}{ l >{\centering\arraybackslash}m{1.8cm}  >{\centering\arraybackslash}m{2.5cm}  >{\centering\arraybackslash}m{2.5cm}  >{\centering\arraybackslash}m{2cm} >{\centering\arraybackslash}m{1cm}}
    \noalign{\hrule height 1.2pt}
    Preference &Majority Class &Multi-Class SMO &Naive Bayes &AdaBoost &JRip\\
    \hline
    Culture      &100.0\%     & 88.2\% (0.34)      &74.8\%      &97.7\%      &96.5\%\\
    Gold         &69.9\%      & 38.6\% (0.50)      &50.4\%      &68.1\%      &62.5\%\\
    Growth       &69.5\%      & 30.8\% (0.83)      &35.1\%      &30.5\%      &33.1\%\\
    Military     &66.6\%      & 34.6\% (0.56)      &51.7\%      &66.6\%      &59.3\%\\
    Religion     &83.2\%      & 59.0\% (0.64)      &61.3\%      &83.2\%      &73.0\%\\
    Science      &83.0\%      & 71.0\% (0.54)      &67.7\%      &83.0\%      &81.5\%\\
    \hline
\end{tabular}
\hfill{}
\vspace{0.1cm}
\label{tb:smoAltResult}
\end{table*}
\end{center}

\begin{center}
\begin{table*}[th]
\centering 
\caption{Improvement of each approach over the baseline \citep{denTeuling_Thesis_10, Spronck_AIIDE_10} when predicting unknown virtual agents. The improvement is computed as the difference between the accuracy and the baseline divided by the baseline.}

\label{tb:improvementAltResult}

\hfill{}
\begin{tabular}{ l  >{\centering\arraybackslash}m{2.5cm}  >{\centering\arraybackslash}m{2.5cm} >{\centering\arraybackslash}m{2.5cm}}
    \noalign{\hrule height 1.2pt}
    Preference &Naive Bayes &AdaBoost &JRip\\
    \hline
    Culture    & -15.2\%  \textcolor[rgb]{0.7,00,00}{$\blacktriangledown$}      & 10.8\% \ \textcolor[rgb]{00,0.45,0.10}{$\blacktriangle$}    &9.4\% \ \ \textcolor[rgb]{00,0.45,0.10}{$\blacktriangle$} \\
    Gold       & 30.6\% \ \textcolor[rgb]{00,0.45,0.10}{$\blacktriangle$}     & 76.4\% \ \textcolor[rgb]{00,0.45,0.10}{$\blacktriangle$}     &61.9\% \textcolor[rgb]{00,0.45,0.10}{$\blacktriangle$} \\
    Growth     & 14.0\% \ \textcolor[rgb]{00,0.45,0.10}{$\blacktriangle$}     & -1.0\% \ \ \textcolor[rgb]{0.7,00,00}{$\blacktriangledown$}    &7.5\% \ \ \textcolor[rgb]{00,0.45,0.10}{$\blacktriangle$}\\
    Military   & 49.4\% \ \textcolor[rgb]{00,0.45,0.10}{$\blacktriangle$}      & 92.5\% \ \textcolor[rgb]{00,0.45,0.10}{$\blacktriangle$}     &71.4\% \textcolor[rgb]{00,0.45,0.10}{$\blacktriangle$}\\
    Religion   & 3.9\% \ \ \ \textcolor[rgb]{00,0.45,0.10}{$\blacktriangle$}     & 41.0\% \ \textcolor[rgb]{00,0.45,0.10}{$\blacktriangle$}    &23.7\% \textcolor[rgb]{00,0.45,0.10}{$\blacktriangle$}\\
    Science    & -4.6\% \ \ \textcolor[rgb]{0.7,00,00}{$\blacktriangledown$}     & 16.9\% \ \textcolor[rgb]{00,0.45,0.10}{$\blacktriangle$}    &14.8\% \textcolor[rgb]{00,0.45,0.10}{$\blacktriangle$}\\
    \hline
\end{tabular}
\hfill{}
\vspace{0.1cm}
\end{table*}
\end{center}

A first interesting result to observe is that the used algorithms, with the binary classification approach, surpass \cite{denTeuling_Thesis_10}'s (\textit{SVM}) in accuracy, corroborating our assumption that \textit{SVM}, despite being considered the state-of-the-art for several classification problems, may not be adequate for modeling player preferences.

\textit{AdaBoost} and \textit{JRip} performances were remarkable, mainly when compared to our baseline. For most of the preferences, these methods were able to obtain accuracies above 60\%, such as 68.1\% (\textit{AdaBoost}) and 62.5\% (\textit{JRip}) against 38.6\% of accuracy obtained by the baseline when predicting the \textit{Gold} preference; and 66.6\% (\textit{AdaBoost}) and 59.3\% (\textit{JRip}) against 34.6\% of the baseline predicting the \textit{Military} preference.

\textit{AdaBoost} seems to have learned to classify every instance as being of the most frequent class and this approach will be further evaluated in the next section, when we use the same models to classify human players preferences. \textit{JRip} presented slightly worse results when compared to \textit{AdaBoost}, but it still presented very good accuracies, such as 96.5\% for \textit{Culture}, 73.0\% for \textit{Religion} and \textit{81.5\%} for \textit{Science}. Additionally, as already discussed, this method has the advantage of generating comprehensive rules, which can be verified by a game designer or AI programmer.

After removing the \textit{Peace} and \textit{Victory Type} features, \textit{JRip} started to generate longer rules that, as we can observe, were able to generalize well. As an example, some rules that were able to correctly classify some preferences are (correctly/incorrectly classified instances between parenthesis): 

\begin{itemize}

\item \textbf{Culture}: \textit{CitiesDiff = -1 $\wedge$ CitiesTrend = 5 $\wedge$ CumulativeWar = 20 $\wedge$ War = 0 $\rightarrow$ Culture Preference}~(722/0);
\item \textbf{Military}: \textit{CumulativeDeclaredWar = 0 $\wedge$ StateReligionDiff = 0 $\wedge$ CumulativeWar = 12 $\wedge$ War = 0 $\rightarrow$ No Military Preference}~(309/0);
\item \textbf{Religion}: \textit{CumulativeDeclaredWar = 0 $\wedge$ CumulativeWar = 35 $\wedge$ War = 0 $\rightarrow$ Religion Preference}~(570/0);
\item \textbf{Science}: \textit{CitiesDiff = 3 $\wedge$ CumulativeWar = 77 $\wedge$ Cities = 8 $\rightarrow$ Science Preference}~(182/0);

\end{itemize}

These rules already give us a glimpse about virtual agents behaviors and preferences, such as the importance of wars and number of cities to define a player preference. In fact, we can even observe very sound rules, such as a lower number of war declarations for virtual agents with preference for \textit{Culture} and \textit{Religion}. The listed rule that detects no \textit{Military} preference is also very intuitive, corroborating our discussion about \textit{JRip} benefits.

\textit{Naive Bayes} presented the worst performance among the evaluated algorithms, despite being better than the baseline in four different preferences. This poor performance may be explained by \textit{Naive Bayes} assumption of independence between features. This independence assumption does not hold in our problem, since the turn number is a feature and the score features are directly related to the game turn.

The unique preference to which our methods performed poorly was \textit{Growth}, with no accuracy greater than 35\%. We believe this is mainly due to the removal of the first 100 turns of each player. As we already showed in Chapter~\ref{player_modeling_civ4}, these turns are extremely relevant to differ artificial players preference for \textit{Growth}, while the later turns may be misleading.

\section{Player Modeling}\label{player_modeling}

After presenting an approach to classify virtual agents preferences that was able to surpass previous approaches in the literature, it is important to test our method with ``real'' data. In this last section we discuss how we gathered data from human players and the performance of the algorithms, already mentioned, when applied to these data.

\subsection{Players' Data}

Aiming at evaluating our approach in data generated by human players, we recruited volunteers to play the game \textsc{Civilization IV}. They were required to add an script in the game directory in order to sniffer their scores along a match. This script is called \textit{AIAutoPlay} and it was modified and used by \cite{denTeuling_Thesis_10} to generate the datasets with virtual agents. We also used it in this thesis, as previously discussed. Its installation only consists in replacing some \textit{dlls} of the original game. It is important to stress that, at the end, we had the same 128 features used in the \textit{online tracking} experiment.

We did not make any restrictions to players regarding their experience or any other characteristic. We required them to play an 1x1 match and to log the game. Before playing, we requested them to sign a consent form and to fill a pre-test questionnaire about their experience in TBS games and, more specifically, games of the \textsc{Civilization} series. After filling this questionnaire, they played the game and then filled a post-test questionnaire informing their self-labeled preference in the played game and evaluating their confidence in this self-labeling. These questionnaires were written in Portuguese and are available in Appendix~B.

Seven players participated in the tests and sent us their data. We judge this number satisfactory due to the difficult to obtain players. A \textit{Civilization IV} game may take longer than four hours and few people accepted to participate in such test.

For usability tests, for example, it is said that three to five users are enough to perform an experiment \citep{Nielsen_CHI_93}.

\begin{figure}
\begin{tabular}{cc}
\subfloat[Have you ever played Turn Based Strategy games?]{\includegraphics[width=8cm]{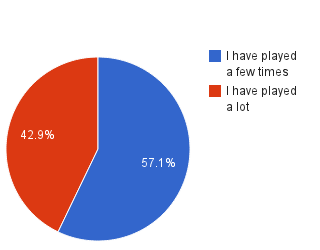}} 
   & \subfloat[When did you play a turn-based strategy game for the last time?]{\includegraphics[width=8cm]{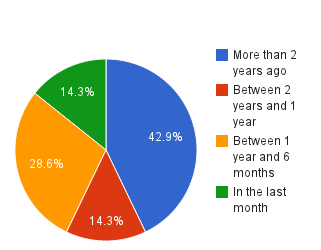}}\\
\subfloat[More specifically, have you ever played any game of the Civilization IV series?]{\includegraphics[width=8cm]{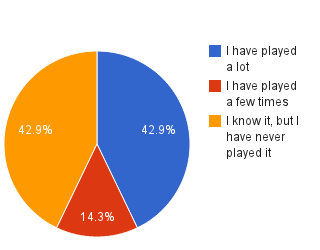}} 
   & \subfloat[How do you classify your ability as a Civilization player?]{\includegraphics[width=8cm]{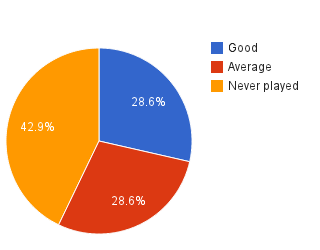}}\\
\end{tabular}

\caption{Players' experience in TBS and \textsc{Civilization} games.}\label{fig:players_characteristics}
\end{figure}

Table~\ref{tb:classDistribHuman} contains the classes distribution in our dataset, while Figure~\ref{fig:players_characteristics} shows the players' self-reported information about the frequency they play(ed) TBS games and, more specifically, games of the series \textsc{Civilization}. 

\begin{center}
\begin{table}[H]
\caption{Number of samples in each class for the dataset generated from matches played by human beings. As all other experiments, we removed the first 100 turns of each match.}
\label{tb:classDistribHuman}
{\footnotesize
\hfill{}
\begin{tabular}{l | c c }
    \noalign{\hrule height 1.2pt}
    Preferences &-1  &1 \\
    \hline
    Culture       & 1,725        & \ \ \ 100 \\
    Gold          & 1,466        & \ \ \ 359 \\
    Growth        & \ \ \ 647    & 1,178     \\
    Military      & 1,436        & \ \ \ 389 \\
    Religion      & \ \ \ 848    & \ \ \ 977 \\
    Science       & \ \ \ \ \ 29 & 1,796     \\
    \hline
\end{tabular}}
\hfill{}
\vspace{0.1cm}
\end{table}
\end{center}

\vspace{-0.8cm}

We can see that there are players who consider themselves ``good'' and those who have never played the game. We believe to have an interesting range to perform our experiments, mainly because \cite{denTeuling_Thesis_10} suggested that experienced players are easier to be classified than those without experience.

\subsection{Classification of Players' Preference}

To classify human players using supervised learning, a large dataset containing labeled samples is required. This is unfeasible in practice because no player can provide so much data in a short period of time, thus another approach must be used. We tackle this problem using data generated from AI $\times$ AI matches to train each classifier. This one of the reasons to perform the \textit{online tracking} experiment in last section, to evaluate the feasibility of this approach.

We use the same training set used in the \textit{online tracking} experiment, containing all matches of the \textit{Traditional Dataset}, to train the models that classify players' matches. In fact, we used those already generated. The results of the classification are presented in Table~\ref{tb:smoPlayersResult}. We decided not to report \cite{denTeuling_Thesis_10} results regarding classification of human players because they performed experiments with only two players, using a different modeling and algorithm.

Observing the results of all seven players, we notice that, among all classifiers, at least one of them obtained an accuracy above 50\% for each preference. \textit{Naive Bayes} obtained accuracies over 50\% for \textit{Growth} (62.7\%), \textit{Military} (61.7\%) and \textit{Science} (89.7\%) while \textit{AdaBoost} and \textit{JRip} obtained accuracies over 50\% for \textit{Culture} (91.5\%, 94.5\%), \textit{Gold} (74.5\%, 74.5\%) and \textit{Religion} (60.7\%, 54.5\%).

\textit{JRip} seems to have generated rules that classify most turns in the most frequent class for the preference \textit{Culture}, and to the most infrequent class for the preferences \textit{Growth} and \textit{Science}. As we already discussed, the \textit{Growth} preference has a particularity that we removed its most discriminative turns: the first 100.
This may have impaired the learning algorithm when dealing to the \textit{Growth} preference, what was constistent with all other obtained results. 

Despite presenting a bad performance for some preferences, different from other classifiers, \textit{JRip} allows improvement with assistance of human designers, since its generated model is comprehensible. Maybe the reported results would have been improved if an expert, such as the game designer, analyzed the generated rules and decided which ones to use, or if a specific rule is classifying a behavior in a wrong class.

\begin{center}
\begin{table*}[th]
\caption{Accuracy of our approaches (Naive Bayes, AdaBoost and JRip) for classifying human players' preferences. No error is shown since we just executed this classification once.}

\hfill{}
\begin{tabular}{ l >{\centering\arraybackslash}m{2cm} >{\centering\arraybackslash}m{2.5cm}  >{\centering\arraybackslash}m{2cm} >{\centering\arraybackslash}m{2cm}}
    \noalign{\hrule height 1.2pt}
    Preference &Majority Class &Naive Bayes &AdaBoost &JRip\\
    \hline
    Culture    &94.5\%		&12.8\%      &91.5\%      &94.5\%\\
    Gold       &80.3\%		&35.9\%      &74.5\%      &74.5\%\\
    Growth     &64.6\%		&62.7\%      &35.5\%      &36.3\%\\
    Military   &78.7\%		&61.7\%      &21.3\%      &25.7\%\\
    Religion   &53.5\%		&25.3\%      &60.7\%      &54.5\%\\
    Science    &98.4\%		&89.7\%      &\ \ 1.6\%   &\ \ 1.6\%\\
    \hline
\end{tabular}
\hfill{}
\vspace{0.1cm}
\label{tb:smoPlayersResult}
\end{table*}
\end{center}

\vspace{-0.5cm}

\textit{Naive Bayes} presented the worse results in average. However, it is interesting to point out that it succeeded where more complex algorithms did not. This may be an indicative that some preferences may be biased, causing overfitting in more complex algorithms. Maybe \textit{Naive Bayes} was able to avoid this overfitting due to its simpler paradigm. It was the unique algorithm that presented good results for the preferences \textit{Growth, Military} and \textit{Science}.

As previously mentioned, \textit{AdaBoost} learned to classify all instances in a specific class. It happened in the previous experiment (\textit{online tracking}) and also here. Its results were good for half of the preferences because it correctly selected the most frequent class for them, while it did not to the other half.

Interestingly, \textit{Naive Bayes} succeeded in three different preferences (\textit{Growth, Military} and \textit{Science}) while \textit{AdaBoost} and \textit{JRip} presented an accuracy over 50\% in the other three (\textit{Culture, Gold} and \textit{Religion}). Further studies must be performed in order to assure that one specific classifier is better and always obtains better results for a specific preference. If it is not true, a hybrid approach considering different classifiers for different preferences may lead to a higher overall accuracy .
Nevertheless, the first thing to evaluate is whether these results may be due to players wrongly self-reporting their preferences. We will present this discussion in sequence.

\subsubsection{Evaluating Players' Preference Considering their Expertise}

The post-test questionnaire asked users to list their preferences in the played match, and their confidence on the provided list. Based on the pretest questionnaire and in the question about player's confidence, we clustered players in two different groups:

\begin{itemize}
	\item Experienced: Three players are part of this group, those who had already played a game of the \textsc{Civilization} series ``a lot'', and that stated to have a high confidence in their self-labeling.
	\item Beginners: This group contains 4 players, those who played, in the best case, few times a game of the \textsc{Civilization} series.
\end{itemize}

This experiment was designed to evaluate whether the results were impaired by inexperienced players who did not label themselves ``properly'', i.e. as experienced players, with a great knowledge of the game, would do. In other words, we evaluate the assumption whether the generated model is correctly identifying preferences, but human players do not have a common understanding about these preferences definition. It is based on the assumption that we have sufficient data to represent both types of players in our training set.

We classified both groups separately, and its accuracies are presented in Tables~\ref{tb:smoExperiencedPlayersResult} and~\ref{tb:smoBeginnerPlayersResult}. These results confirm our assumptions about \textit{AdaBoost}. The algorithm was not able to learn a model different from classifying all instances in a class, and for most of the results, in Tables~\ref{tb:smoExperiencedPlayersResult} and~\ref{tb:smoBeginnerPlayersResult}, we can see that it either obtained an accuracy close to the majority class or it obtained an accuracy close to 100\% minus the majority class (when it chose the wrong class). Due to that, the \textit{AdaBoost} performance in the two groups (experienced and beginners) should not be considered, since it is much more a matter of luck than any other topic.

Just as the first experiment, we can observe that \textit{JRip} presents a good accuracy for \textit{Culture, Gold} and \textit{Religion}, while \textit{NaiveBayes} obtained good accuracies classifying \textit{Growth, Military} and \textit{Science}.

\begin{center}
\begin{table*}[th]
\caption{Accuracy of the evaluated approaches (Naive Bayes, AdaBoost and JRip) for classifying the preferences of experienced human players. No error is shown since we just executed this classification once.}

\hfill{}
\begin{tabular}{ l >{\centering\arraybackslash}m{2cm} >{\centering\arraybackslash}m{2.5cm}  >{\centering\arraybackslash}m{2cm} >{\centering\arraybackslash}m{2cm}}
    \noalign{\hrule height 1.2pt}
    Preference &Majority Class &Naive Bayes &AdaBoost &JRip\\
    \hline
    Culture    &\ \ 86.1\%		&15.5\%      &\ \ 86.1\%      &85.9\%\\
    Gold       &100.0   \%		&\ \ 8.5\%   &\ \ 99.6\%      &80.9\%\\
    Growth     &\ \ 63.9\%		&65.6\%      &\ \ 36.1\%      &40.5\%\\
    Military   &100.0   \%		&96.2\%      &\ \ \ \ 0.0\%      &11.4\%\\
    Religion   &\ \ 63.9\%		&\ \ 3.8\%   &100.0\%      &83.7\%\\
    Science    &100.0   \%		&97.5\%      &\ \ \ \ 0.0\%      &\ \ 0.0\%\\
    \hline
\end{tabular}
\hfill{}
\vspace{0.1cm}
\label{tb:smoExperiencedPlayersResult}
\end{table*}
\end{center}

\begin{center}
\begin{table*}[th]
\caption{Accuracy of the evaluated approaches (Naive Bayes, AdaBoost and JRip) for classifying the preference of beginner human players. No error is shown since we just executed this classification once.}

\hfill{}
\begin{tabular}{ l >{\centering\arraybackslash}m{2cm} >{\centering\arraybackslash}m{2.5cm}  >{\centering\arraybackslash}m{2cm} >{\centering\arraybackslash}m{2cm}}
    \noalign{\hrule height 1.2pt}
    Preference &Majority Class &Naive Bayes &AdaBoost &JRip\\
    \hline
    Culture    &100.0   \%		&11.0\%      &94.9\%      &100.0\%\\
    Gold       &\ \ 67.6\%		&53.7\%      &58.2\%      &\ \ 70.3\%\\
    Growth     &\ \ 65.0\%		&60.8\%      &35.0\%      &\ \ 33.6\%\\
    Military   &\ \ 64.9\%		&39.3\%      &35.1\%      &\ \ 35.0\%\\
    Religion   &\ \ 64.9\%		&39.3\%      &35.1\%      &\ \ 35.5\%\\
    Science    &\ \ 97.4\%		&84.6\%      &\ \ 2.6\%   		&\ \ \ \ 2.6\%\\
    \hline
\end{tabular}
\hfill{}
\vspace{0.1cm}
\label{tb:smoBeginnerPlayersResult}
\end{table*}
\end{center}

\vspace{-0.8cm}

Regarding the division between player's trustfulness, we observe that there is not a consistent pattern when classifying the two groups. Differently from \cite{denTeuling_Thesis_10}, we did not obtained higher accuracies for experienced players. \textit{JRip}, for example, had half of its results improved in the experienced group (\textit{Gold, Growth} and \textit{Religion}) and the other half reduced (\textit{Culture, Military} and \textit{Science}).

Observing the results of \textit{NaiveBayes}, we notice that it leads to higher accuracy variation. The algorithm obtained an accuracy, when classifying beginners' \textit{Gold} preference, equals to 53.7\%, while it obtained only 8.5\% of accuracy 	for experienced players. A similar variation occurred for \textit{Religion} (39.3\% against 3.8\%) and \textit{Military} (39.3\% against 96.2\%). 

Due to these results, we can conclude that, in this experiment, the separation between player's experience, or confidence in their self-labeled preferences, does not consistently improves classifiers' accuracy. These results contradict the assumption presented by \cite{denTeuling_Thesis_10} that experienced player are ``easier'' to be classified. Table~\ref{tb:specif_player_accuracy}, at Appendix~\ref{accuracy_humans}, presents the accuracies obtained classifying each individual player, and its experience level (classified by us using their answers in the questionnaires). We can see that there are experienced players who are not correctly classified while there are beginners who are. Thus, further investigation is necessary.

\section{Overview about Modeling Players using ML}

After we executed experiments classifying virtual agents and human players, we are able to perform a higher level discussion about the machine learning approach applied to the \textit{player modeling} problem. As an overview, we can say that we were able to generate models, from virtual agents, that are generic enough to classify other virtual agents with a satisfactory accuracy. However, the same generated models were not always effective when classifying human players, what may be justified by virtual agents behaviors not generalizing to human players.

Despite our efforts to validate our assumption that matches played between virtual agents are capable to generate useful examples for the classifier, we must consider the possibility that the generated data may not be sufficient to determine the classifier, \textit{overfitting} it. This may be the reason for our consistent poor performance when classifying the \textit{Growth} preference, as well as our results when classifying human players' \textit{Military} and \textit{Science} preferences using \textit{JRip}.

To better understand this topic, we quote \cite{Domingos_CommACM_12}:

\begin{Exemplo}
``One way to understand overfitting is by decomposing generalization error into \textit{bias} and \textit{variance} \citep{Domingos_ICML_00}. Bias is a learner's tendency to consistently learn the same wrong thing. Variance is the tendency to learn random things irrespective of the real signal.	''
\end{Exemplo}

When decomposing overfitting in these two topics, we observe the need to further evaluate both. Regarding bias, it may be possible that the classifiers we used did not assume the correct frontier shape of the \textsc{Civilization IV} \textit{player modeling} problem. This may cause a poor performance of the classifier when predicting preferences that are not structured as it assumes. This may also justify the better performance of the \textit{NaiveBayes} classifier when modeling \textit{Gold, Growth} and \textit{Science}, since it is known that simpler algorithms may yield better solutions in complex spaces, if more sophisticated algorithms are not properly tunned \citep{Domingos_CommACM_12}. It is also important to stress that \textit{NaiveBayes} assumes independence between features, different from \textit{JRip}, for example. This assumption may also have an impact when classifying different preferences.

This problem is a complex topic that must be further studied. Not surprisingly, we were able to show that, as it seems, the algorithm that was considered the state of the art for several classification tasks, and more specifically to this problem \citep{Spronck_AIIDE_10}, was surpassed by those we used in our methodology.

Related to variance, ``After overfitting, the biggest problem in machine learning is the \textit{curse of dimensionality} (...) Generalizing correctly becomes exponentially harder as the dimensionality (number of features) of the examples grows (...)'' \citep{Domingos_CommACM_12}. This may be the second problem with ML approaches. In this thesis, we have shown that a subset of our features was useful to be used by a classifier. However, we added many other features by intuition. This is a common flaw in \textit{player modeling} papers that may lead to worse results. \cite{Domingos_CommACM_12} also discusses it:

\begin{Exemplo}
``Naively, one might think that gathering more features never hurts, since at worst they provide no new information about the class. But in fact their benefits may be outweighed by the curse of dimensionality.''
\end{Exemplo}

It is clear that the features used are also extremely important when performing an ML task, specifically to our problem, when \textit{modeling players}. As far as we know, this is an unexplored topic in the field, mainly because games present several features and its combination may provide useful information. This justifies why automate feature engineering does not solve the problem, since selecting features by information gain, for example, may be useless once it will ignore relations between features. This is why we performed the extensive analysis in Chapter~\ref{player_modeling_civ4}, relating turns with game score indicators, and analyzing its meaning in the game. 

%% file: conclusion.tex
\chapter{Conclusion}\label{conclusion}

\epigraphhead[270]{\scriptsize ``All right,'' said Deep Thought. ``The Answer to the Great Question...'' \\
									 ``Yes..!''\\
									 ``Of Life, the Universe and Everything...'' said Deep Thought. \\
									 ``Yes...!'' \\
									 ``Is...'' said Deep Thought, and paused.\\
									  (...) \\
									 ``Forty-two,'' said Deep Thought, with infinite majesty and calm. \\
									 \vspace{-0.6cm}
									 \begin{flushright}
									 Douglas Adams, The Hitchhiker's Guide to the Galaxy
									 \end{flushright}
								   }
\vspace{3.8cm}

This chapter presents the conclusions of this thesis. In the next section, we summarize our main contributions, present some considerations regarding our results and discuss some limitations of our work. Section~\ref{future_work} lists some possible extensions of this research.

\section{Contributions and Discussion}\label{main_considerations}

In this work we have presented an extensive discussion about player modeling. Among our main contributions, and its correspondent publications, we can list:

\begin{itemize}
\item The organization of a taxonomy to organize the field \citep{Machado_CGames_11};
\item The proposal of a generic approach to tackle player modeling as a machine learning problem \citep{Machado_CIG_12};
\item The evaluation of a generic representation that may be used to model players in different games \citep{Machado_SBGames_12};
\item The proposition of an approach based on linear regressions to evaluate features' ability to distinguish different classes \citep{Machado_SBGames_11}; and
\item The evaluation of different classifiers, regarding its accuracy and generalization capabilities, when applied to virtual agents and human players.
\end{itemize}

\cite{Domingos_CommACM_12} states that ``the fundamental goal of machine learning is to generalize beyond the examples in the training set'', and we were able to do this for most of the preferences, what indicates the effectiveness of the approach proposed here. When comparing all the evaluated classifiers, we can conclude that there are two main choices that may be done for future applications regarding player modeling: one may choose \textit{NaiveBayes} as a simpler and faster approach to the problem, but which may lead to worse performances; or \textit{JRip}, which is a method that obtains higher accuracies and generates rules that may be evaluated by game designers, despite being a more computational expensive method.

In spite of being able to surpass, sometimes by far, the state of the art of player's preferences classification \citep{denTeuling_Thesis_10, Spronck_AIIDE_10}, it became evident that player modeling is a very hard task. We were not able to obtain, with a single classifier, good accuracies for all preferences modeled, and this leads us to the challenges of the field. A first obvious challenge is the difficult to label human preferences. In spite of not further discussing this topic in the last chapter, sometimes it is extremely hard to know, even analyzing by hand, whether a player has or has not a preference for a given game feature. It is even harder to ask him to label himself, since different comparison basis may exist between those who design and those who play the game. It is also important to stress that it may be difficult to a player distinguish between what are the preferences he would like to have and what are the preferences he expresses in the game.

Another challenge in this field is the difficulty to obtain gameplay data from human players. In this thesis, we used matches played by AI agents in order to generate a training set. This approach may be deceptive since the assumption that the training data is similar to the data that will be classified is weaker than in other applications of ML.

As we discussed in last chapter, the selection of the appropriate classifier to a specific task is a hard topic. We have shown that a method that is considered the state of the art for several classification problems (\textit{SVM}) did not perform well in this problem, and we cannot guarantee that another, different from those tested, will not perform better. We selected methods based on different assumptions to evaluate the impact of their differences on this type of problem. 

At this point it is important to discuss the main limitations of our work. We address some of them as future work, in the next section. A specific characteristic that we did not explore extensively in our approach was the fact that each feature we used was dependent of its previous values, and that they represented, in fact, the story of a match. We assumed each turn as independent, and classified them separately, sometimes labeling different turns of a same match with different preferences. We are aware that a player may change its preference along a match, but we believe the reported results are more likely to indicate that the set of turns should be taken as an evidence. We tried to do this adding temporal features such as derivatives, but it is not clear whether it is enough.

Finally, despite performing evaluations about the representativeness of some features, we assumed that all the others available would also be representative and we used all of them in our classifiers. This contradicts what \cite{Domingos_CommACM_12} says, that ``one might think that gathering more features never hurts, since at worst they provide no new information about the class. But in fact their benefits may be outweighed by the curse of dimensionality''. So a different approach should be evaluated in the future.
	
\section{Future Work}\label{future_work}

There are several paths for future work. Regarding the proposed taxonomy, a more quantitative work can be developed to analyze the current status of the field, classifying more papers with the proposed taxonomy and looking for trends. Also, a complete new analysis can be done in terms of learning algorithms, answering questions as ``what can we learn in games?''. We could, for instance, correlate the main learning techniques to the main problems of the field.

Related to the generic representation discussed in Chapter~\ref{generic_approach}, a possible future study is the impact of its application in the game development process, evaluating its benefits for level designers and programmers. We could also analyze the application of hierarchical representations. This topic was proposed by \cite{Houlette_AIWisdom2_03}, which suggests the generation of weights that represent very low level actions, such as \textit{throw grenade} or \textit{use rifle} and a hierarchical organization that extracts higher level information as the combination of its leaf nodes. Higher level representations could be \textit{aggression} or \textit{intelligence}, for example. A promising evaluation research on this topic is the evaluation of the impact of each preference on game indicators, as well as the impact of the interaction between preferences. Additionally, we could also correlate the victory types and players' preferences, aiming an \textit{off-line review}; this could also assist us in our discussing regarding the impact of match results in the classification.

Finally, related to the main topic of this work, the automatic classification of player's preferences, several tasks can be performed. Regarding the evaluation of features, we want to perform different data analyses, which could assist us in the task of selecting appropriate features to the classifiers. Some interesting approaches are: to evaluate the impact of using the first 100 turns in each preference classification and to apply some feature selection methods to the dataset, maybe trying to consider indicators' semantics. As we already discussed, we believe that the absence of the first 100 turns was determinant in the poor accuracy obtained when classifying \textit{Growth} preference. The feature selection problem is a hard topic since, as discussed, a feature by itself may be meaningless (such as turn), but when associated with others, it may be very useful. In fact, there are several possibilities to further evaluate this topic and, as far as we know, few works have presented concerns in automatically selecting the smaller and most representative set of features for modeling players.

Additionally, our results may be improved by the application of over/undersampling techniques in our dataset, what could deal with unbalanced classes. Another topic that we did not discuss, but may also worth investigation due to its applicability in the whole field, is the evaluation of the correctness of players' self labeled preferences.

A last possible future work, and most ambitious, is the use (or proposal) of a technique that considers each turn as an intermediate state of the player. Instead of classifying each player turn, the algorithm would understand each turn as a stream and would classify a player after a given number of turns. These turns would be responsible for increasing the algorithm confidence that a player has a specific preference. A work that used this approach is \cite{Bard_AAAI_07}, who used particle filtering to model \textsc{Poker} players. However, their approach was much simpler than the one that would be required here.

In summary, we have made several contributions to the field, such as the taxonomy, the evaluation of a generic representation and the study of appropriate features to model players, as well as the evaluation of several different classifiers, discussing its possible use in industry. In fact, we believe that this thesis highlighted the specific need of better evaluating ML decisions that are generally made without much care, such the selection of features and classifiers. We consider that player modeling is still an open topic that deserves lots of investigation.

%% file: appendix.tex
\begin{appendices}

\chapter{\textsc{Civilization IV} Dataset Features}\label{appendix_features}

\begin{small}
\begin{center}
\begin{longtable}{ l l l }
\caption{List of all features used in the training process. \cite{denTeuling_Thesis_10} also presents this list, including features' range. Recall that features 129 and 130 were used only in the first classification experiment called \textit{off-line review}, in Chapter~\ref{experimental_results}. The operators \textit{Derivate, Trend, TrendDerivate, Diff, DiffDerivate, DiffTrend} and \textit{DiffTrendDerivate} are in Table~\ref{tb:modifications}.}\\

1: \ \ Turn                     &45: PlotsTrend                &89: \ \ AgricultureDiffTrend         \\
2: \ \ War                      &46: PlotsTrendDerivate        &90: \ \ AgricultureDiffTrendDerivate \\
3: \ \ Cities                   &47: PlotsDiff                 &91: \ \ Power                        \\ 
4: \ \ CitiesDerivate           &48: PlotsDiffDerivate         &92: \ \ PowerDerivate                \\
5: \ \ CitiesTrend              &49: PlotsDiffTrend            &93: \ \ PowerTrend                   \\
6: \ \ CitiesTrendDerivate      &50: PlotsDiffTrendDerivate    &94: \ \ PowerTrendDerivate           \\
7: \ \ CitiesDiff               &51: Techs                     &95: \ \ PowerDiff                    \\ 
8: \ \ CitiesDiffDerivate       &52: TechsDerivate             &96: \ \ PowerDiffDerivate            \\
9: \ \ CitiesDiffTrend          &53: TechsTrend                &97: \ \ PowerDiffTrend               \\ 
10: CitiesDiffTrendDerivate     &54: TechsTrendDerivate        &98: \ \ PowerDiffTrendDerivate       \\
11: Units                       &55: TechsDiff                 &99: \ \ Culture                      \\
12: UnitsDerivate               &56: TechsDiffDerivate         &100: CultureDerivate                 \\
13: UnitsTrend                  &57: TechsDiffTrend            &101: CultureTrend                    \\ 
14: UnitsTrendDerivate          &58: TechsDiffTrendDerivate    &102: CultureTrendDerivate            \\
15: UnitsDiff                   &59: Score                     &103: CultureDiff                     \\
16: UnitsDiffDerivate           &60: ScoreDerivate             &104: CultureDiffDerivate             \\
17: UnitsDiffTrend              &61: ScoreTrend                &105: CultureDiffTrend                \\ 
18: UnitsDiffTrendDerivate      &62: ScoreTrendDerivate        &106: CultureDiffTrendDerivate        \\ 
19: Population                  &63: ScoreDiff                 &107: Maintenance                     \\ 
20: PopulationDerivate          &64: ScoreDiffDerivate         &108: MaintenanceDerivate             \\ 
21: PopulationTrend             &65: ScoreDiffTrend            &109: MaintenanceTrend                \\ 
22: PopulationTrendDerivate     &66: ScoreDiffTrendDerivate    &110: MaintenanceTrendDerivate        \\
23: PopulationDiff              &67: Economy                   &111: GoldRate                        \\
24: PopulationDiffDerivate      &68: EconomyDerivate           &112: GoldRateDerivate                \\
25: PopulationDiffTrend         &69: EconomyTrend              &113: GoldRateTrend                   \\
26: PopulationDiffTrendDerivate &70: EconomyTrendDerivate      &114: GoldRateTrendDerivate           \\
27: Gold                        &71: EconomyDiff               &115: ResearchRate                    \\
28: GoldDerivate                &72: EconomyDiffDerivate       &116: ResearchRateDerivate            \\
29: GoldTrend                   &73: EconomyDiffTrend          &117: ResearchRateTrend               \\
30: GoldTrendDerivate           &74: EconomyDiffTrendDerivate  &118: ResearchRateTrendDerivate       \\
31: GoldDiff                    &75: Industry                  &119: CultureRate                     \\
32: GoldDiffDerivate            &76: IndustryDerivate          &120: CultureRateDerivate             \\ 
33: GoldDiffTrend               &77: IndustryTrend             &121: CultureRateTrend                \\
34: GoldDiffTrendDerivate       &78: IndustryTrendDerivate     &122: CultureRateTrendDerivate        \\
35: Land                        &79: IndustryDiff              &123: StateReligionDiff               \\
36: LandDerivate                &80: IndustryDiffDerivate      &124: DeclaredWar                     \\
37: LandTrend                   &81: IndustryDiffTrend         &125: CumulativeDeclaredWar           \\
38: LandTrendDerivate           &82: IndustryDiffTrendDerivate &126: AverageDeclaredWar              \\ 
39: LandDiff                    &83: Agriculture               &127: CumulativeWar                   \\
40: LandDiffDerivate            &84: AgricultureDerivate       &128: AverageWar                      \\
41: LandDiffTrend               &85: AgricultureTrend          &129: VictoryType                     \\
42: LandDiffTrendDerivate       &86: AgricultureTrendDerivate  &130: Peace                           \\
43: Plots                       &87: AgricultureDiff           & \\
44: PlotsDerivate               &88: AgricultureDiffDerivate   & \\

\vspace{0.1cm}
\label{tb:listOfFeatures}
\end{longtable}
\end{center}

\end{small}

\chapter{Summary of Indicators' Linear Regressions}\label{app:regression}

\begin{tiny}
  \begin{longtable}{ | c | c | c | c | c | c | c | c |}
     \caption{Summary table of the linear regressions discussed in Chapter~\ref{player_modeling_civ4}, where the adopted model was $y = b_0 + b_1x$. The column meanings are, respectively: the data collected in the game, the agent who generated the data, the interval (in turns) the data represents, the game result evaluated (general (victory + defeat), only victories or only defeats), the coefficient of determination (how much of the data is explained by the regression), both coefficients and its confidence intervals and, finally, the confidence used to generate these intervals.}\\
    \hline
    Indicator &Agent &Interval &Result &$R^{2}$ &$b_0$ &$b_1$ &Confidence\\ \hline \hline \hline
    \textit{GoldRate} & Louis XIV &[1:460] &General &98.72\% &$-19.7615 (\pm8.1688)$ &$0.3853 (\pm0.0307)$ &99\%\\\hline
    \textit{GoldRate} & Mansa Musa &[1:460] &General &96.14\% &$-11.2732 (\pm20.0013)$ &$0.3419 (\pm0.0752)$ &99\%\\\hline
    \textit{Gold} & Louis XIV &[1:300] &General &62.39\% &$44.0798 (\pm76.3993)$ &$0.2969 (\pm0.4400)$ &90\%\\\hline
    \textit{Gold} & Mansa Musa &[1:340] &General &31.08\% &$47.5944 (\pm295.0593)$ &$0.2771 (\pm1.4998)$ &90\%\\\hline
    \textit{Gold} & Louis XIV &[301:460] &General &75.33\% &$-948.8215 (\pm11127.5812)$ &$3.5891 (\pm28.9012)$ &90\%\\\hline
    \textit{Gold} & Mansa Musa &[341:460] &General &94.67\% &$-2059.4734 (\pm4691.6026)$ &$6.2651 (\pm11.6708)$ &90\%\\\hline 
    \hline
    $\sqrt[4]{CultureRate}$ & Alexander &[1:460] &General &98.93\% &$1.0939(\pm0.0035)$ &$0.0096(\pm1 \times 10^{-5})$ &99\%\\\hline
    $\sqrt[4]{CultureRate}$ & Hatshepsut &[1:460] &General &99.11\% &$1.3567 (\pm0.0047)$ &$ 0.0101(\pm2 \times 10^{-5})$ &99\%\\\hline
    $\sqrt[5]{Culture}$ & Alexander &[1:460] &General &99.86\% &$1.7772 (\pm0.0019)$ &$0.0183 (\pm7 \times 10^{-6})$ &99\%\\\hline
    $\sqrt[5]{Culture}$ & Hatshepsut &[1:460] &General &99.85\% &$2.1366 (\pm0.0023)$ &$0.0194 (\pm9 \times 10^{-6})$ &99\%\\\hline 
    \textit{Cities} & Alexander &[1:220] &General &97.17\% &$0.49439(\pm0.0408)$ &$0.03143(\pm0.0003)$ &99\%\\\hline
    \textit{Cities} & Hatshepsut &[1:220] &General &96.80\% &$0.5561(\pm0.0411)$ &$0.0296(\pm0.0003)$ &99\%\\\hline
    \textit{Cities} & Alexander &[221:460] &General &71.39\% &$6.2654(\pm0.0072)$ &$0.0021(\pm2 \times 10^{-5})$ &99\%\\\hline
    \textit{Cities} & Hatshepsut &[221:460] &General &56.02\% &$5.9320(\pm0.0099)$ &$0.0018(\pm2 \times 10^{-5})$ &99\%\\\hline
    \textit{Land} & Alexander &[1:200] &General &97.90\% &$7.7862(\pm4.0299)$ &$0.4899(\pm0.0347)$ &90\%\\\hline
    \textit{Land} & Hatshepsut &[1:200] &General &93.76\% &$16.6760(\pm13.2991)$ &$0.5043(\pm0.1147)$ &90\%\\\hline
    \textit{Land} & Alexander &[201:460] &General &23.90\% &$94.9505(\pm0.8058)$ &$0.0078(\pm0.0015)$ &90\%\\\hline
    \textit{Land} & Hatshepsut &[201:460] &General &50.04\% &$99.2728(\pm0.7268)$ &$0.0166(\pm0.0021)$ &90\%\\\hline
    \textit{Plots} & Alexander &[1:200] &General &99.15\% &$3.6923(\pm7.0220)$ &$0.8176(\pm0.0606)$ &99\%\\\hline
    \textit{Plots} & Hatshepsut &[1:200] &General &98.05\% &$12.9458(\pm19.3722)$ &$0.8900(\pm0.1671)$ &99\%\\\hline
    \textit{Plots} & Alexander &[201:460] &General &78.73\% &$149.6914(\pm13.6018)$ &$0.1109(\pm0.0401)$ &99\%\\\hline
    \textit{Plots} & Hatshepsut &[201:460] &General &88.22\% &$163.7941(\pm6.0591)$ &$0.1053(\pm0.0178)$ &99\%\\\hline
    \hline
    \textit{GoldRate} & Louis XIV &[1:460] &Victory &98.48\% &$-25.3125(\pm9.2060)$ &$0.4694(\pm0.0346)$ &90\%\\\hline
    \textit{GoldRate} & Mansa Musa &[1:460] &Victory &97.03\% &$-24.2833(\pm19.5158)$ &$0.4842(\pm0.0733)$ &90\%\\\hline
    \textit{GoldRate} & Louis XIV &[1:460] &Defeat &97.98\% &$-11.8778(\pm4.6118)$ &$0.2867(\pm0.0173)$ &90\%\\\hline
    \textit{GoldRate} & Mansa Musa &[1:460] &Defeat &84.60\% &$-0.4737(\pm26.7967)$ &$0.2326(\pm0.1007)$ &90\%\\\hline
    $\sqrt[3]{Gold}$ & Louis XIV &[1:460] &Victory &79.94\% &$2.6224(\pm0.1775)$ &$0.0128(\pm0.0007)$ &99\%\\\hline
    $\sqrt[3]{Gold}$ & Mansa Musa &[1:460] &Victory &72.60\% &$3.0379(\pm0.1408)$ &$0.0093(\pm0.0005)$ &99\%\\\hline
    $\sqrt[3]{Gold}$ & Louis XIV &[1:460] &Defeat &82.35\% &$3.1341(\pm0.0692)$ &$0.0087(\pm0.0003)$ &99\%\\\hline
    $\sqrt[3]{Gold}$ & Mansa Musa &[1:460] &Defeat &60.15\% &$2.7212(\pm0.3289)$ &$0.0108(\pm0.0012)$ &99\%\\\hline
    \hline
    $\sqrt[4]{CultureRate}$ & Alexander &[1:460] &Victory &99.33\% &$1.0614(\pm0.0029)$ &$0.0102(\pm1 \times 10^{-5})$ &99\%\\\hline
    $\sqrt[4]{CultureRate}$ & Hatshepsut &[1:460] &Victory &98.78\% &$1.3165(\pm0.0065)$ &$0.0111(\pm2 \times 10^{-5})$ &99\%\\\hline
    $\sqrt[4]{CultureRate}$ & Alexander &[1:460] &Defeat &98.95\% &$1.0867(\pm0.0034)$ &$0.0086(\pm1 \times 10^{-5})$ &99\%\\\hline
    $\sqrt[4]{CultureRate}$ & Hatshepsut &[1:460] &Defeat &98.20\% &$1.4589(\pm0.0057)$ &$0.0085(\pm1 \times 10^{-5})$ &99\%\\\hline
    $\sqrt[5]{Culture}$ & Alexander &[1:460] &Victory &99.86\% &$1.7152(\pm0.0022)$ &$0.0195(\pm8 \times 10^{-6})$ &99\%\\\hline
    $\sqrt[5]{Culture}$ & Hatshepsut &[1:460] &Victory &99.87\% &$2.0610(\pm0.0024)$ &$0.0210(\pm8 \times 10^{-6})$ &99\%\\\hline
    $\sqrt[5]{Culture}$ & Alexander &[1:460] &Defeat &99.84\% &$1.8082(\pm0.0019)$ &$0.0171(\pm7 \times 10^{-6})$ &99\%\\\hline
    $\sqrt[5]{Culture}$ & Hatshepsut &[1:460] &Defeat &99.65\% &$2.2963(\pm0.0046)$ &$0.0176(\pm2 \times 10^{-5})$ &99\%\\\hline
    \hline
    \textit{Cities} & Alexander &[1:200] &Victory &98.01\% &$0.3077(\pm0.0188)$ &$0.0343(\pm0.0002)$ &90\%\\\hline
    \textit{Cities} & Hatshepsut &[1:200] &Victory &97.91\% &$0.3045(\pm0.0205)$ &$0.0350(\pm0.0002)$ &90\%\\\hline
    \textit{Cities} & Alexander &[201:460] &Victory &95.11\% &$4.9651(\pm0.0143)$ &$0.0082(\pm4 \times 10^{-5})$ &99\%\\\hline
    \textit{Cities} & Hatshepsut &[201:460] &Victory &81.79\% &$6.1602(\pm0.0202)$ &$0.0047(\pm5 \times 10^{-5})$ &99\%\\\hline
    \textit{Cities} & Alexander &[1:200] &Defeat &96.98\% &$0.4894(\pm0.0278)$ &$0.0309(\pm0.0002)$ &95\%\\\hline
    \textit{Cities} & Hatshepsut &[1:200] &Defeat &96.58\% &$0.5466(\pm0.0262)$ &$0.0281(\pm0.0002)$ &95\%\\\hline
    \textit{Cities} & Alexander &[201:460] &Defeat &79.89\% &$6.9878(\pm0.0072)$ &$-0.003(\pm2 \times 10^{-5})$ &95\%\\\hline
    \textit{Cities} & Hatshepsut &[201:460] &Defeat &39.59\% &$6.0004(\pm0.0199)$ &$-0.0020(\pm7 \times 10^{-5})$ &95\%\\\hline
    \textit{Land} & Alexander &[1:180] &Victory &98.45\% &$5.7677(\pm3.8054)$ &$0.5498(\pm0.0364)$ &95\%\\\hline
    \textit{Land} & Hatshepsut &[1:180] &Victory &96.59\% &$11.5814(\pm10.8481)$ &$0.6202(\pm0.1039)$ &95\%\\\hline
    \textit{Land} & Alexander &[181:460] &Victory &93.65\% &$89.7060(\pm0.7846)$ &$0.0606(\pm0.0023)$ &95\%\\\hline
    \textit{Land} & Hatshepsut &[181:460] &Victory &53.95\% &$111.5217(\pm1.3284)$ &$0.0222(\pm0.0040)$ &95\%\\\hline
    \textit{Land} & Alexander &[1:180] &Defeat &97.96\% &$6.7485(\pm3.3395)$ &$0.4475(\pm0.0320)$ &95\%\\\hline
    \textit{Land} & Hatshepsut &[1:180] &Defeat &93.27\% &$15.6509(\pm13.7272)$ &$0.4877(\pm0.1315)$ &95\%\\\hline
    \textit{Land} & Alexander &[181:460] &Defeat &74.29\% &$85.2093(\pm0.2312)$ &$-0.0145(\pm0.0009)$ &95\%\\\hline
    \textit{Land} & Hatshepsut &[181:460] &Defeat &2.74\% &$91.6748(\pm2.5132)$ &$-0.0047(\pm0.0076)$ &95\%\\\hline
    \textit{Plots} & Alexander &[1:250] &Victory &98.43\% &$7.9857(\pm17.0135)$ &$0.7904(\pm0.1175)$ &99\%\\\hline
    \textit{Plots} & Hatshepsut &[1:250] &Victory &97.26\% &$17.4191(\pm38.4130)$ &$0.8929(\pm0.2653)$ &99\%\\\hline
    \textit{Plots} & Alexander &[251:460] &Victory &96.77\% &$0.1694(\pm0.0104)$ &$151.8041(\pm3.7521)$ &99\%\\\hline
    \textit{Plots} & Hatshepsut &[251:460] &Victory &73.78\% &$203.3449(\pm10.4762)$ &$0.0866(\pm0.0290)$ &99\%\\\hline
    \textit{Plots} & Alexander &[1:210] &Defeat &99.06\% &$7.0925(\pm6.3405)$ &$0.7110(\pm0.0521)$ &99\%\\\hline
    \textit{Plots} & Hatshepsut &[1:180] &Defeat &97.21\% &$14.0890(\pm20.5100)$ &$0.8248(\pm0.1965)$ &99\%\\\hline
    \textit{Plots} & Alexander &[211:460] &Defeat &50.76\% &$145.9592(\pm7.0910)$ &$0.0424(\pm0.0207)$ &99\%\\\hline
    \textit{Plots} & Hatshepsut &[181:460] &Defeat &42.52\% &$152.8303(\pm11.8983)$ &$0.0461(\pm0.0359)$ &99\%\\\hline
  \end{longtable}
 \end{tiny}

\chapter{Questionnaires applied to Human Players}\label{questionnaires}

This appendix presents both questionnaires applied to the human players. After signing a consent form, they were asked to fill a pretest questionnaire. Its objective was to obtain the player's profile. After filling it, they played a match of the game \textsc{Civilization IV} and were asked to fill the post-test questionnaire. All questionnaires were presented to the players in Portuguese.

\section{Pretest Questionnaire}

The pretest questionnaire was the larger questionnaire players were asked to answer. Not all players have answered all the questions, since the questionnaire was dynamic, i.e. it does not ask the player how frequently he/she has played a game if, in the previous question, the player answered that he/she does not know turn-based strategy games.

All the available questions on the questionnaire are presented in Figures \ref{fig:basic_pre-test}, \ref{fig:tbs_pre-test}, \ref{fig:civ_pre-test}, \ref{fig:tbs_xp_pre-test} and \ref{fig:ack}. Regarding its flow, it is presented below.

All players started answering the questionnaire~\ref{fig:basic_pre-test}. At the last question, if the player selects one of the first two bullets, he/she is forwarded to questionnaire~\ref{fig:ack}, otherwise to questionnaire~\ref{fig:tbs_pre-test}. In questionnaire~\ref{fig:tbs_pre-test}, the next questionnaire to be answered is also defined by the last question. If the player choses one of the first two bullets, it is forwarded to questionnaire~\ref{fig:tbs_xp_pre-test}, otherwise to questionnaire~\ref{fig:civ_pre-test}. Finally, both questionnaires, when answered, forward the player to a final acknowledgment message (\ref{fig:ack}), with further instructions to play the game.

\begin{figure} 
\renewcommand{\labelitemi}{\Large$\circ$\normalsize}
\centering
\begin{boxedminipage}{\linewidth}
\footnotesize

\paragraph{\footnotesize Dados Pessoais\vspace{0.2cm}}

\subparagraph{\footnotesize Qual o seu nome?}

\subparagraph{\tiny O nome ser\'a utilizado apenas para agrupamento das respostas deste question\'ario com as do question\'ario a ser respondido ap\'os os testes. N\~ao ser\'a divulgado ou apresentado de forma alguma, de acordo com o termo de consentimento.\vspace{0.5cm}}

\subparagraph{\footnotesize Qual a sua faixa et\'aria?}

\begin{itemize}\itemsep=0ex
  \item Abaixo de 18 anos
  \item 18 a 20 anos
  \item 21 a 25 anos
  \item 26 a 30 anos
  \item Acima de 30 anos
\end{itemize}

\subparagraph{\footnotesize Você j\'a jogou jogos de estrat\'egia baseados em turnos (TBS - Turn Based Strategy)?\vspace{-0.6cm}}
\subparagraph{\tiny Alguns exemplos de jogos TBS: "Civilization", "Heroes of Might and Magic", "Panzer General", "Galactic Civilizations", "Age of Wonders", "Colonization" e "Call to Power".}

\begin{itemize}\itemsep=0ex
  \item N\~ao conhe\c{c}o
  \item Conhe\c{c}o, mas nunca joguei
  \item J\'a joguei um pouco
  \item J\'a joguei bastante
\end{itemize}

\normalsize
\end{boxedminipage}

\caption{Pretest questionnaire: Questions about player's personal information.}

\label{fig:basic_pre-test}
\renewcommand{\labelitemi}{$\bullet$}
\end{figure} 

\begin{figure} 
\renewcommand{\labelitemi}{\Large$\circ$\normalsize}
\centering
\begin{boxedminipage}{\linewidth}
\footnotesize
\paragraph{\footnotesize Experi\^encia com jogos de estrat\'egia baseados em turnos\vspace{0.2cm}}

\subparagraph{\footnotesize Qual foi a \'ultima vez que jogou uma partida de um jogo de estrat\'egia baseado em turnos?}

\begin{itemize}\itemsep=0ex
  \item H\'a mais de 2 anos
  \item Entre 2 anos e 1 ano
  \item Entre 1 ano e 6 meses
  \item Entre 6 meses e 1 m\^es
  \item No \'ultimo m\^es
\end{itemize}

\subparagraph{\footnotesize Mais especificamente, você j\'a jogou algum jogo da s\'erie \textsc{Civilization}?}

\begin{itemize}\itemsep=0ex
  \item N\~ao conhe\c{c}o
  \item Conhe\c{c}o, mas nunca joguei
  \item J\'a joguei um pouco
  \item J\'a joguei bastante
\end{itemize}

\normalsize
\end{boxedminipage}

\caption{Pretest questionnaire: Questions about player's experience in turn-based strategy games.}

\label{fig:tbs_pre-test}
\renewcommand{\labelitemi}{$\bullet$}
\end{figure} 

\begin{figure} 
\renewcommand{\labelitemi}{\Large$\circ$\normalsize}
\centering
\begin{boxedminipage}{\linewidth}
\footnotesize
\paragraph{\footnotesize Experi\^encia com jogos da s\'erie \textsc{Civilization IV} \vspace{0.2cm}}

\subparagraph{\footnotesize Qual jogo da s\'erie \textsc{Civilization} voc\^e j\'a jogou?}

\begin{itemize}\itemsep=0ex
  \item Civilization
  \item Civilization II
  \item Civilization III
  \item Civilization IV
  \item Civilization V
\end{itemize}

\subparagraph{\footnotesize Qual foi a \'ultima vez que jogou uma partida de um jogo da s\'erie \textsc{Civilization}?}

\begin{itemize}\itemsep=0ex
  \item H\'a mais de 2 anos
  \item Entre 2 anos e 1 ano
  \item Entre 1 ano e 6 meses
  \item Entre 6 meses e 1 m\^es
  \item No \'ultimo m\^es
\end{itemize}

\subparagraph{\footnotesize Na \'epoca em que mais jogava algum jogo da s\'erie \textsc{Civilization}, com qual frequ\^encia jogava?}

\begin{itemize}\itemsep=0ex
	\item Menos de 1 vez por semana
	\item 1 vez por semana
	\item 2 vezes por semana
	\item 3 vezes por semana
	\item 4 vezes por semana
	\item 5 vezes por semana
	\item Mais de 5 vezes por semana
\end{itemize}

\subparagraph{\footnotesize Como voc\^e classifica seu n\'ivel de habilidade como jogador dos jogos da s\'erie \textsc{Civilization}?}

\begin{itemize}\itemsep=0ex
	\item Fraco
	\item Razo\'avel
	\item Bom
	\item Excelente
\end{itemize}

\normalsize
\end{boxedminipage}

\caption{Pretest questionnaire: Questions about player's experience in games of the \textsc{Civilization} series.}

\label{fig:civ_pre-test}
\renewcommand{\labelitemi}{$\bullet$}
\end{figure} 

\begin{figure} 
\renewcommand{\labelitemi}{\Large$\circ$\normalsize}
\centering
\begin{boxedminipage}{\linewidth}
\footnotesize

\paragraph{\footnotesize Experi\^encia com jogos de estrat\'egia baseados em turnos\vspace{0.2cm}}

\subparagraph{\footnotesize Como você classifica seu nível de habilidade como jogador de jogos de estratégia baseados em turnos?}

\begin{itemize}\itemsep=0ex
  \item Fraco
  \item Razo\'avel
  \item Bom
  \item Excelente
\end{itemize}

\normalsize
\end{boxedminipage}

\caption{Pretest questionnaire: Questions about player's experience in turn-based strategy games.}

\label{fig:tbs_xp_pre-test}
\renewcommand{\labelitemi}{$\bullet$}
\end{figure} 

\begin{figure} 
\renewcommand{\labelitemi}{\Large$\circ$\normalsize}
\centering
\begin{boxedminipage}{\linewidth}
\footnotesize

\paragraph{\footnotesize Conclus\~ao\vspace{0.2cm}}

\subparagraph{\footnotesize Obrigado por participar do question\'ario. Por favor, agora siga as instru\c{c}\~oes presentes no endere\c{c}o abaixo \url{http://www.dcc.ufmg.br/~marlos/civ4.html} para continuar o teste.}

\subparagraph{\footnotesize Obrigado por participar do question\'ario. Por favor, agora siga as instru\c{c}\~oes presentes no endere\c{c}o abaixo \url{http://www.dcc.ufmg.br/~marlos/civ4.html} para continuar o teste.}

\normalsize
\end{boxedminipage}

\caption{Acknowledgment of the pretest questionnaire.}

\label{fig:ack}
\renewcommand{\labelitemi}{$\bullet$}
\end{figure} 

\clearpage

\section{Post-test Questionnaire}

After playing the game, all players were required to fill the post-test questionnaire. Its objective was to obtain the players' preferences and their confidence on the self-labeled preferences. This questionnaire is presented in Figure~\ref{fig:post_test}.

\begin{figure}[h]
\renewcommand{\labelitemi}{\Large$\circ$\normalsize}

\centering
\begin{boxedminipage}{\linewidth}
\footnotesize

\paragraph{\footnotesize Question\'ario P\'os-Teste\vspace{0.2cm}}

\subparagraph{\footnotesize Qual o seu nome?}

\subparagraph{\tiny O nome ser\'a utilizado apenas para agrupamento das respostas deste question\'ario com as do question\'ario a ser respondido ap\'os os testes. N\~ao ser\'a divulgado ou apresentado de forma alguma, de acordo com o termo de consentimento.\vspace{0.5cm}}
	
\subparagraph{\footnotesize Qual o nome do seu agente, no jogo?}

\subparagraph{\tiny Exemplos de agentes s\~ao: \textit{Alexander, Cyrus, Mansa Musa}. No arquivo de log que me enviar\'a, existir\'a um nome em cada arquivo, um nome \'e o seu, o outro, do seu rival.\vspace{0.5cm}}

\subparagraph{\footnotesize Como voc\^e classificaria suas prefer\^encias durante a partida jogada de \textsc{Civilization IV?}}

\subparagraph{\tiny Marque apenas os itens que voc\^e julga ter agido como se tivesse prefer\^encia por ele. Essa \'e uma avalia\c{c}\~ao bin\'aria, ou seja, ou voc\^e apresentou ou voc\^e n\~ao apresentou um comportamento condizente com cada uma das prefer\^encias abaixo.\vspace{0.5cm}}

\begin{itemize}
\item Ci\^encia
\item Crescimento
\item Cultura
\item Militar
\item Ouro
\item Religi\~ao
\end{itemize}

\subparagraph{\footnotesize Qu\~ao confiante voc\^e se sente com rela\c{c}\~ao \`a sua avalia\c{c}\~ao anterior?\vspace{-0.5cm}}

\subparagraph{\tiny D\^e uma nota de 0 a 5, onde 0 significa sem confian\c{c}a alguma e 5 total confian\c{c}a.}

\begin{itemize}
\item 0
\item 1
\item 2
\item 3
\item 4
\item 5
\end{itemize}

\normalsize
\end{boxedminipage}

\caption{Post-test questionnaire: Questions about the match played and the player's preference.}

\label{fig:post_test}
\renewcommand{\labelitemi}{$\bullet$}
\end{figure} 

\chapter{Accuracy Classifying each Human Player}\label{accuracy_humans}

\vspace{-0.8cm}

\begin{center}
\begin{table*}[th]
\centering 
\caption{Accuracy of the three evaluated methods (\textit{Naive Bayes, AdaBoost} and \textit{JRip}) when classifying each individual human player (\%). They are identified by numbers to hide their identity. Players' numbers with an asterisk represent experienced players, while those without asterisk are beginners.}
\label{tb:specif_player_accuracy}
\hfill{}

\begin{tabular}{cc c c c c c c}
\cline{1-8}
&\textbf{Player \#} & \textbf{Culture} & \textbf{Gold} & \textbf{Growth} & \textbf{Military} &\textbf{Religion} &\textbf{Science} \\[0.15cm] \cline{1-8}
\multicolumn{1}{ c }{\multirow{7}{*}{\begin{sideways}\textbf{NaiveBayes}\end{sideways}} } &
\multicolumn{1}{ |c| }{\ \ 1*} & \ \ 93.0 	& \ \ \ \ 1.0 		& \ \ 95.0 		&100.0 		& \ \ \ \ 0.0 		& 100.0   \\ 
\multicolumn{1}{ c  }{}                        &
\multicolumn{1}{ |c| }{2} & \ \ \ \ 2.8 		& \ \ 84.1 		& \ \ 91.1 			& \ \ 85.5 		& \ \ 85.5 	& \ \ 91.6    \\ 
\multicolumn{1}{ c  }{}                        &
\multicolumn{1}{ |c| }{3} & 100.0 	& 100.0 	& \ \ \ \ 3.4 			&100.0 		&100.0 	& \ \ \ \ 0.0    	\\ 
\multicolumn{1}{ c  }{}                        &
\multicolumn{1}{ |c| }{\ \ 4*} & \ \ \ \ 1.9		& \ \ 11.2			& \ \ \ \ 7.7			& \ \ 95.4			& \ \ \ \ 4.6		& \ \ 97.7    	\\ 
\multicolumn{1}{ c  }{}                        &
\multicolumn{1}{ |c| }{5} & \ \ \ \ 6.1			& \ \ \ \ 2.2			& \ \ 95.0			& \ \ \ \ 5.8			& \ \ \ \ 5.8		& \ \ 99.7 		\\ 
\multicolumn{1}{ c  }{}                        &
\multicolumn{1}{ |c| }{6} & \ \ 17.0		& \ \ 71.0			& \ \ \ \ 0.8			& \ \ 21.7			& \ \ 21.7		& \ \ 69.4			\\ 
\multicolumn{1}{ c  }{}                        &
\multicolumn{1}{ |c| }{\ \ 7*} & \ \ \ \ 3.6			& \ \ \ \ 8.6			& \ \ 99.2			& \ \ 95.8			& \ \ \ \ 4.2		& \ \ 96.7    	\\ \cline{1-8} 
\multicolumn{1}{ c  }{\multirow{7}{*}{\begin{sideways}\textbf{AdaBoost}\end{sideways}} } &
\multicolumn{1}{ |c| }{\ \ 1*} & \ \ \ \ 0.0			& \ \ 97.0			& \ \ \ \ 0.0			& \ \ \ \ 0.0			&100.0	&\ \ \ \ 0.0	    \\
\multicolumn{1}{ c  }{}                        &
\multicolumn{1}{ |c| }{2} &100.0		& \ \ \ \ 0.0			& \ \ \ \ 0.0			& \ \ \ \ 0.0			& \ \ \ \ 0.0		& \ \ \ \ 0.0 	   	\\
\multicolumn{1}{ c  }{}                        &
\multicolumn{1}{ |c| }{3} &100.0		&100.0		&100.0		&100.0		&100.0	&100.0		\\
\multicolumn{1}{ c  }{}                        &
\multicolumn{1}{ |c| }{\ \ 4*} &100.0		&100.0		&100.0		& \ \ \ \ 0.0			&100.0	& \ \ \ \ 0.0		 	\\
\multicolumn{1}{ c  }{}                        &
\multicolumn{1}{ |c| }{5} &100.0		& \ \ 71.1			& \ \ \ \ 0.0			&100.0		&100.0	& \ \ \ \ 0.0 			\\
\multicolumn{1}{ c  }{}                        &
\multicolumn{1}{ |c| }{6} & \ \ 84.4		&100.0		&100.0		& \ \ \ \ 0.0			& \ \ \ \ 0.0		& \ \ \ \ 0.0	    \\
\multicolumn{1}{ c  }{}                        &
\multicolumn{1}{ |c| }{\ \ 7*} &100.0		&100.0		& \ \ \ \ 0.0			& \ \ \ \ 0.0			&100.0	& \ \ \ \ 0.0 			\\ \cline{1-8}
\multicolumn{1}{ c  }{\multirow{7}{*}{\begin{sideways}\textbf{JRip}\end{sideways}} } &
\multicolumn{1}{ |c| }{\ \ 1*} & \ \ \ \ 0.0			& \ \ 74.0			& \ \ \ \ 0.0			& \ \ 29.0			& \ \ 59.0		& \ \ \ \ 0.0 			\\
\multicolumn{1}{ c  }{}                        &
\multicolumn{1}{ |c| }{2} &100.0		& \ \ 15.6			& \ \ \ \ 0.0			& \ \ \ \ 7.5			& \ \ \ \ 6.1		& \ \ \ \ 0.0			\\
\multicolumn{1}{ c  }{}                        &
\multicolumn{1}{ |c| }{3} &100.0		&100.0		& \ \ 41.4			&100.0		&100.0	&100.0\\
\multicolumn{1}{ c  }{}                        &
\multicolumn{1}{ |c| }{\ \ 4*} &100.0		& \ \ 71.8			&100.0		& \ \ 20.1			& \ \ 77.6		& \ \ \ \ 0.0\\
\multicolumn{1}{ c  }{}                        &
\multicolumn{1}{ |c| }{5} &100.0		& \ \ 92.8			& \ \ \ \ 0.3			& \ \ 91.9			& \ \ 95.0		& \ \ \ \ 0.0\\
\multicolumn{1}{ c  }{}                        &
\multicolumn{1}{ |c| }{6} &100.0		&100.0		&100.0		& \ \ \ \ 0.0			& \ \ \ \ 0.0		& \ \ \ \ 0.0\\
\multicolumn{1}{ c  }{}                        &
\multicolumn{1}{ |c| }{\ \ 7*} &\ \ 99.7		& \ \ 89.4			& \ \ \ \ 8.9			& \ \ \ \ 0.3			& \ \ 95.0		& \ \ \ \ 0.0\\ \cline{1-8}
\end{tabular}

\hfill{}
\vspace{0.1cm}
\end{table*}
\end{center}

\end{appendices}